\documentclass{article} 
\usepackage{iclr2027_conference,times}

\usepackage{amsmath,amsfonts,bm}

\def\eqref#1{equation~\ref{#1}}

\def\1{\bm{1}}

\DeclareMathAlphabet{\mathsfit}{\encodingdefault}{\sfdefault}{m}{sl}
\SetMathAlphabet{\mathsfit}{bold}{\encodingdefault}{\sfdefault}{bx}{n}

\usepackage[hidelinks]{hyperref}
\usepackage{url}
\usepackage{booktabs}
\usepackage{xcolor}
\usepackage{listings}
\usepackage{longtable}
\usepackage{booktabs}
\usepackage{caption}
\usepackage{adjustbox}
\usepackage{algorithm}
\usepackage{algorithmic}
\definecolor{boost}{HTML}{548235}

\usepackage{multirow}

\usepackage{xcolor}
\usepackage{booktabs}
\usepackage{array}
\usepackage{amsmath,amssymb}
\usepackage{graphicx}
\usepackage{subcaption}
\usepackage{listings}
\usepackage{needspace}
\usepackage{hyperref}

\definecolor{codekw}{RGB}{0,0,160}
\definecolor{codecomment}{RGB}{110,110,110}
\definecolor{codestr}{RGB}{160,40,40}
\definecolor{codeskill}{RGB}{190,85,0}
\lstdefinestyle{skillcode}{%
  language=Python,
  basicstyle=\ttfamily\scriptsize,
  keywordstyle=\color{codekw},
  commentstyle=\color{codecomment}\itshape,
  stringstyle=\color{codestr},
  identifierstyle=\color{black},
  emph={fold_edge_over,alternating_smoothing_pats,flip_object_over,adjust_in_place,
        spread_taut_and_release,open_closure,pour_into_target,stir_contents,
        wipe_surface_back_and_forth,fold_over_and_crease,wrap_around_target,
        brace_and_take_tool,scoop_and_carry},
  emphstyle=\bfseries,
  showstringspaces=false,
  breaklines=true,
  breakatwhitespace=false,
  columns=fullflexible,
  keepspaces=true,
  frame=single,
  framerule=0.4pt,
  rulecolor=\color{black!25},
  backgroundcolor=\color{black!4},
  xleftmargin=0.6em,
  xrightmargin=0.2em,
  framexleftmargin=0.6em,
  aboveskip=4pt,
  belowskip=4pt,
}
\usepackage[most]{tcolorbox}
\usepackage{enumitem}

\newtcolorbox{agentprompt}[1]{
  enhanced, breakable,
  colback=black!2,
  colframe=black!30,
  colbacktitle=black!7,
  coltitle=black,
  title={#1},
  fonttitle=\bfseries\small,
  fontupper=\small,
  boxrule=0.4pt,
  arc=1pt,
  left=7pt, right=7pt,
  top=6pt, bottom=6pt,
  before skip=8pt,
  after skip=8pt,
  before upper={
    \raggedright
    \setlength{\parindent}{0pt}
  }
}

\newenvironment{promptrules}{
  \begin{description}[
    font=\normalfont\bfseries,
    style=nextline,
    leftmargin=0pt,
    labelindent=0pt,
    itemsep=4pt,
    parsep=0pt,
    topsep=0pt
  ]
}{
  \end{description}
}
\lstdefinestyle{prompt}{basicstyle=\ttfamily\footnotesize,breaklines=true,
  breakindent=0pt,frame=single,columns=fullflexible,keepspaces=true}

\newcommand{\Ours}{AW-0}

\title{Agentic World Action Models}

\author{%
\begin{minipage}[t]{\dimexpr\textwidth-2\tabcolsep\relax}
\raggedright\normalfont\bfseries
Zhuoyang Liu\textsuperscript{1*}, Dantong Niu\textsuperscript{1,2*\textdagger}, Botong Zhang\textsuperscript{3*}, Boning Shao\textsuperscript{1*}, \\[3pt]
Jing Wang\textsuperscript{2}, Ruijie Zheng\textsuperscript{2}, Yuvan Sharma\textsuperscript{2}, Runyu Lu\textsuperscript{2},
Jiaming Liu\textsuperscript{2}, \\[3pt]
Renrui Zhang\textsuperscript{3}, Roei Herzig\textsuperscript{1}, Yutong Bai\textsuperscript{1},
Ken Goldberg\textsuperscript{1}, Danfei Xu\textsuperscript{2},  \\[3pt]
Yuke Zhu\textsuperscript{2\textdagger}, Guanzhi Wang\textsuperscript{2\textdagger},
Linxi “Jim” Fan\textsuperscript{2\textdagger}, Trevor Darrell\textsuperscript{1\textdagger}\normalfont
\\[6pt]
\textsuperscript{1} UC Berkeley \quad \textsuperscript{2} NVIDIA \quad \textsuperscript{3} University of Pennsylvania \\
{\small \textsuperscript{*}Equal Contribution \quad \textsuperscript{\textdagger}Equal Advising }\\[3pt]
{\small \textbf{Project Page:} \url{https://agentic-wam.github.io/}}
\end{minipage}%
}

\iclrfinalcopy 
\begin{document} \addtocontents{toc}{\protect\setcounter{tocdepth}{-1}}
\raggedbottom

\maketitle
\lhead{}

\begin{center}
\begin{minipage}{\linewidth}
    \centering
    \includegraphics[width=\linewidth]{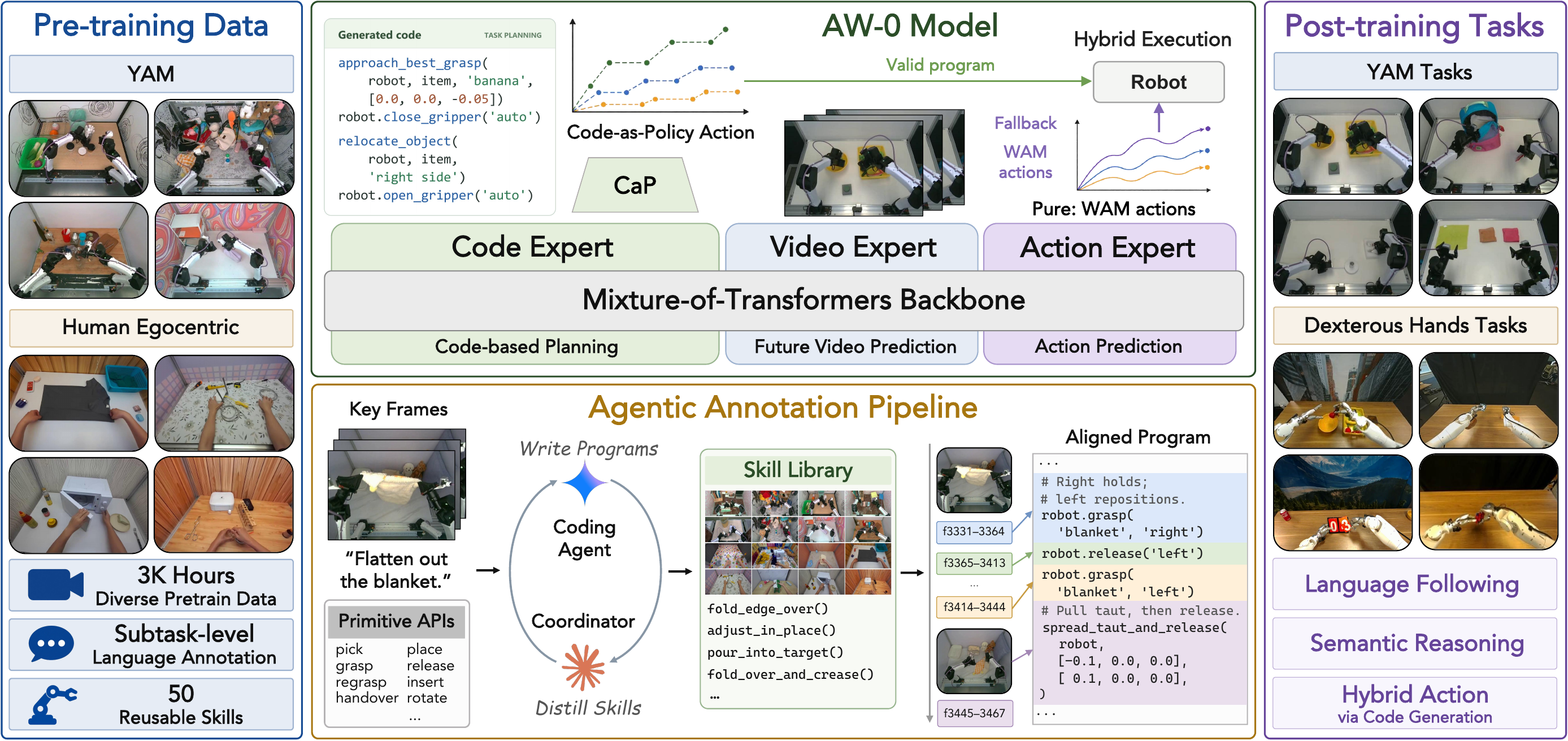}
    \captionof{figure}{
        \textbf{Overview of \Ours{}.} \Ours{} is an agentic world action model that unifies code-based planning with world action modeling by jointly denoising skill programs, future video, and control trajectories. An agentic pipeline discovers reusable skills and annotates robot and human demonstrations with programs for both pretraining and downstream fine-tuning. The resulting policy achieves strong performance on both parallel-gripper and dexterous-hand tasks.
    }
    \label{fig:teaser}
\end{minipage}
\end{center}

\begin{abstract}
Recent world action models provide visual foresight and future prediction for robot manipulation, but lack explicit mechanisms for compositional skill reuse. Agentic robotics combines code-based planning and reasoning with reusable skills, but often lacks predictive grounding in physical dynamics. 
We introduce \textbf{\Ours{}},
an agentic world action model framework that integrates code-based planning with joint video and action prediction. We jointly model code, future video, and robot actions as three coupled generative streams within a unified mixture-of-transformers architecture. 
Within this architecture a code expert generates parameterized skill routines and control logic while video and action experts generate future visual states and low-level motor commands. 
Code serves two complementary roles: it structures multi-step prediction and control within the world model and represents skills as reusable programs for subsequent agentic execution. 
We further introduce an automated curation pipeline that labels robot and human video trajectories with aligned programs and consolidates recurring routines into reusable skill libraries. 
Evaluations on RoboTwin 2.0 and physical bimanual manipulation tasks show improved task performance over VLA and WAM baselines, alongside gains in language following, semantic reasoning and robustness,
demonstrating that agentic code enables connection between planning, prediction and execution within WAMs.
\end{abstract}

\section{Introduction}
\label{sec:introduction}

Learning-based robotic manipulation has progressed from small task-specific visuomotor policies over a decade ago 
to broadly supervised foundation models that connect perception, language, and action~\citep{levine2016endtoendtrainingdeepvisuomotor,zitkovich2023rt,kim2024openvla,niu2024llarvavisionactioninstructiontuning,black2024pi0, bjorck2025gr00t, niu2025arm4r,liu2025hybridvlacollaborativediffusionautoregression, black2025pi05, liu2026last0latentspatiotemporalchainofthought}. 
In parallel, self-supervised video prediction has long been valued as a  learning signal to model the world: image inpainting extended to video prediction  led to action-conditioned video predictors that can imagine the future before committing to an action \citep{pathak2016contextencodersfeaturelearning,he2021maskedautoencodersscalablevision,tong2022videomaemaskedautoencodersdataefficient,finn2016unsupervisedlearningphysicalinteraction, assran2025vjepa2selfsupervisedvideo, ye2026dreamzero}.

World Action Models (WAMs) couple visual prediction with visuomotor policies, grounding control in representations of how the physical world evolves~\citep{ye2026dreamzero, kim2026cosmospolicyfinetuningvideomodels, yuan2026fastwamworldactionmodelsneed}.
This predictive grounding is particularly valuable for complex manipulation, where successful execution depends on anticipating object motion, contact, and the consequences of low-level actions. Yet predicted video and action chunks do not by themselves specify how a long-horizon task should be decomposed: which reusable skills to invoke, how to parameterize them, or how to compose them across subtasks.
Predicting the world and/or actions in general can also be expensive, especially as it can be difficult to establish limits on how many degrees of freedom need to be “world modeled” (or teleoperated, in the supervised setting).

Agentic robot systems provide complementary structure to WAMs: a programmatic harness that selects reusable skills and tools, binds them to the current scene, tracks progress, and replans as observations change~\citep{liang2023code,fu2026capx,zhang2026playfulagenticrobotlearning, xiao2026enpireagenticrobotpolicy,lu2026aspire}. This programmatic representation makes the structure of multi-step tasks explicit, specifying which skills and tools to invoke, with what arguments, and in what sequence.
Yet agentic robotics often lacks predictive grounding in real-world physical dynamics and does not leverage a learned model of how the world will change as the robot acts. 

We advocate for an architecture that combines the foresight and action execution of WAMs with the cognitive efficiency, reusable skills, and potential for future self-improvement of agentic robotics.
We propose \textbf{\Ours}, an agentic WAM framework that incorporates code-based planning and generation directly into world action modeling via a grounded code expert that harnesses broad programming priors to synthesize modular skills featuring explicit parameters and control logic. Code guides synchronized video and control modules within a mixture-of-transformers architecture, simultaneously denoising discrete code structures, predicted visual observations, and low-level motor commands. 

We first show that such agents can be realized by adding a code expert to a WAM, leveraging multisensor action and code generation in a planning stage.  
We develop a modular generation framework:  planning via code generation significantly improves WAM performance, by appropriately conditioning the long-horizon generations required for complex robotic tasks. An automated curation pipeline extracts systematic routines into reusable program libraries, mapping code segments directly onto robotic and human video trajectories. This approach bridges high-level semantic reasoning with predictive physical dynamics, establishing a unified language for compositionality across distinct environments and tasks. 
We further demonstrate  agentic skill reuse via a hybrid deployment strategy that selects between raw action chunk execution and programmatic skill invocation. Empirical evaluations across RoboTwin 2.0 benchmarks and physical bimanual setups show substantial performance gains compared to standard VLA and WAM baselines, highlighting programmatic code as an effective structural backbone for predictive planning, skill discovery, and embodied robotic control.

Our primary contribution is the conceptualization and realization of agentic robotic skills within a world action model framework, showing that code generation improves action prediction in the world model, and that the code generated from the world model can be re-used in an agentic framework.  Specific novel contribution elements include: a  mixture-of-transformers framework integrating a code expert into world action models via simultaneous denoising of code, video, and control; an automated agentic pipeline designed for program annotation and skill discovery; a dataset of such structured skill programs; and a hybrid execution policy that dynamically toggles between programmatic skill invocation and direct deployment of action chunks from the WAM action expert.
\section{Agentic World Action Model}
\label{sec:method}

\label{sec:preliminaries}
We formulate \Ours{} as a unified conditional generative model over code, video, and actions, which respectively describe structured skill plans, future scene dynamics, and continuous robot motion. We represent a skill program as a discrete token sequence $C_0=(c_1,\ldots,c_{N_c})$, future observations as continuous video latents $Z_0$ encoded by a pretrained variational autoencoder, and robot motion as an action chunk $A_0=(a_t,\ldots,a_{t+H-1})$ over a horizon of $H$ steps.

The model is conditioned on the current multi-view observations $\mathcal{O}_t$, a language instruction $L$, the current robot state $S_t$, and a compact memory $M_t$ of the preceding program slices. Together, these inputs define the conditioning context for our joint generative model:
\begin{equation}
  \mathcal{C}_t=\{\mathcal{O}_t,L,S_t,M_t\},
  \qquad p_\theta(C_0,Z_0,A_0\mid\mathcal{C}_t).
  \label{eq:joint_conditional_model}
\end{equation}

The unified conditional generation proceeds by jointly denoising the three modalities. Given masked code $\widetilde C_r$ and noisy continuous states $Z_{\sigma_v}$ and $A_{\sigma_a}$, the unified model predicts
\begin{equation}
  (\ell_\theta^c,v_\theta^v,v_\theta^a)
  =F_\theta(\widetilde C_r,Z_{\sigma_v},A_{\sigma_a},\sigma_v,\sigma_a;\mathcal{C}_t),
  \label{eq:joint_mot_generation}
\end{equation}
where $r$ is the code-mask ratio, $\sigma_v$ and $\sigma_a$ are the video and action noise levels, $\ell_\theta^c$ are code-token logits, and $v_\theta^v,v_\theta^a$ are the video and action velocity predictions. The formulation unifies structured skill programming, future dynamics prediction, and continuous control in one generative process.

\subsection{Mixture of Transformer Architecture}
\begin{figure}[t]
  \centering
  \includegraphics[width=1.0\linewidth]{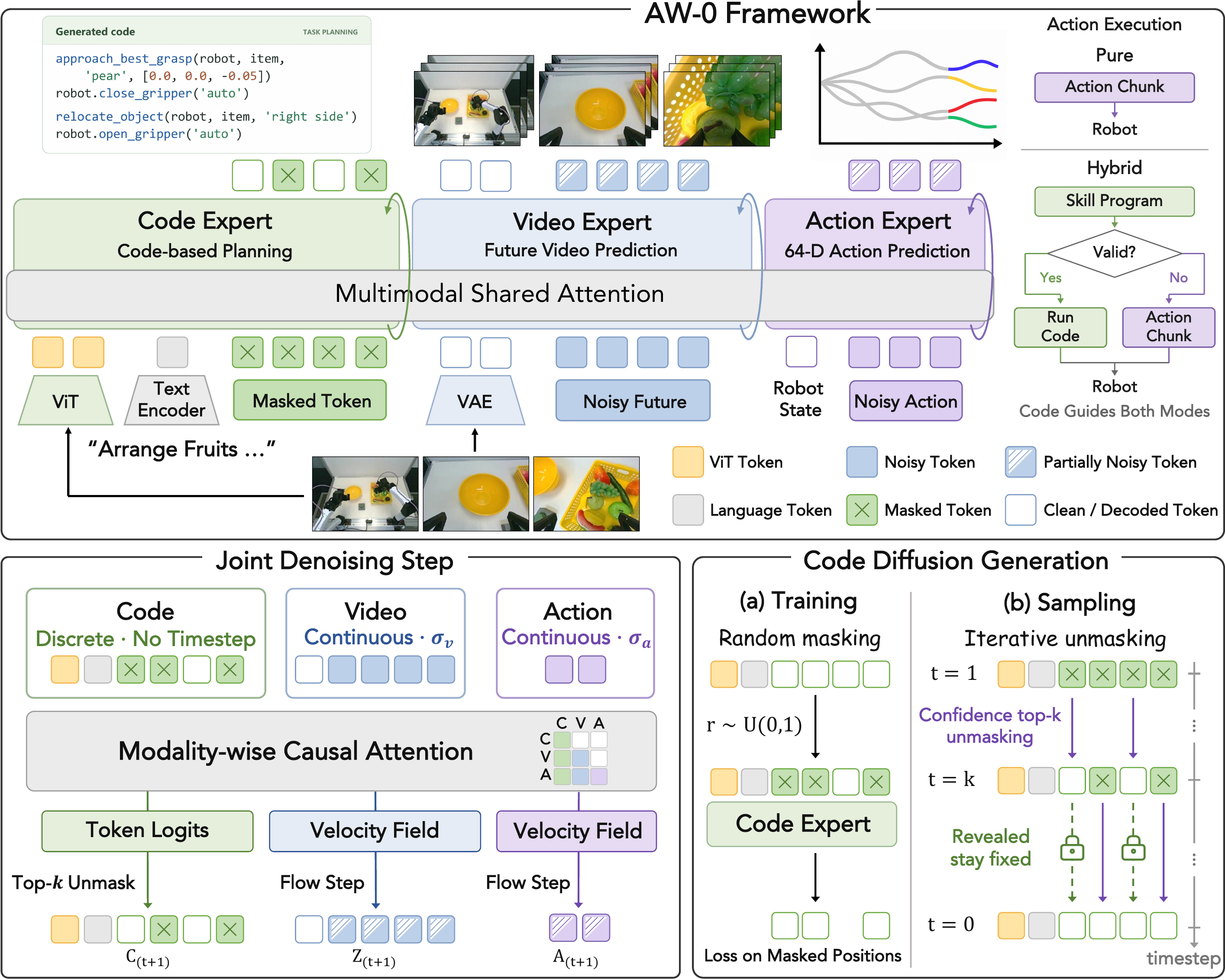}
  \caption{\textbf{\Ours{} Framework.} \Ours{} includes a Mixture-of-Transformers backbone with three experts generating code, video and action, respectively, which are denoised jointly, with video and action under continuous flow matching and the code under discrete masked diffusion.}
  \label{fig:model_architecture}
\end{figure}

As shown in Figure~\ref{fig:model_architecture}, \Ours{} adopts a Mixture-of-Transformers (MoT)~\citep{liang2025mixtureoftransformerssparsescalablearchitecture} architecture with three experts for skill planning, future visual prediction, and continuous control. Each expert keeps its own attention projections ($W_Q,W_K,W_V,W_O$), layer normalizations, and FFNs, communicating through masked joint attention at every transformer layer. The attention mask is bidirectional within each expert and blockwise causal across experts, following the order of code, video and action. We provide additional model and training details in Appendix~\ref{app:model_training_details}.

\textbf{Code Expert.}
The code expert augments a pretrained language backbone with a frozen Qwen3-VL~\citep{Qwen3-VL} ViT and adopts LLaDA-style masked-token diffusion~\citep{nie2025largelanguagediffusionmodels}. Current multi-view observations are encoded into visual tokens and prepended to the instruction and program tokens to ground skill planning in the scene.

\textbf{Video and Action Experts.}
The video expert uses a video DiT initialized from Cosmos-Predict2.5~\citep{ali2025worldsimulationvideofoundationmodels} to model VAE-encoded future frames conditioned on observed frames, while a separate action DiT generates continuous action chunks. Both experts receive language embeddings through cross-attention, with additional robot-state conditioning for the action expert. 

\subsection{Joint Denoising Generation}
\label{sec:joint_denoising}

At each denoising step, the MoT jointly predicts code logits and continuous velocities as defined in Eq.~\ref{eq:joint_mot_generation}. The code stream supplies planning context to the video and action experts, while the action expert also reads video representations. These connections operate throughout generation, allowing all three modalities to evolve within the same iterative process.

\textbf{Modality-Specific Corruption.}
Video and action generation use continuous flow matching. For $m\in\{v,a\}$, with $X_0^v=Z_0$ and $X_0^a=A_0$, we construct noisy inputs as
\begin{equation}
  X_{\sigma_m}^m=(1-\sigma_m)X_0^m+\sigma_m\epsilon_m,
  \qquad \epsilon_m\sim\mathcal{N}(0,I),
  \qquad \sigma_m=\phi_{\gamma_m}(u_m)
  =\frac{\gamma_m u_m}{1+(\gamma_m-1)u_m}.
  \label{eq:continuous_modality_corruption}
\end{equation}
The original video sampling distribution $p_{\mathrm{pre}}$ is induced by $u_v=s b$, where $b\sim\mathrm{Beta}(7,1)$ and $s=0.999$; actions use $u_a\sim U(0,1)$. Both shifts are $\gamma_m=5$. Code instead uses discrete absorbing-mask corruption. We sample a mask ratio $r\sim U(0,1)$ and corrupt each token independently:
\begin{equation}
  \widetilde c_i=
  \begin{cases}
    [\mathrm{MASK}], & b_i=1,\\
    c_i, & b_i=0,
  \end{cases}
  \qquad b_i\sim\mathrm{Bernoulli}(r).
  \label{eq:code_mask_corruption}
\end{equation}
Observed frames, visual prefixes, and language conditions remain uncorrupted. Training samples the modality-specific noise levels independently. The continuous experts regress the velocity target $\epsilon_m-X_0^m$, while the code expert minimizes cross-entropy over masked program positions.

\textbf{Noise Resampling for Predictive Guidance.}
To provide more informative video features for action prediction, we broaden the high-noise pretrained distribution with a uniform component:
\begin{equation}
  u_v\sim q(u)=\alpha p_{\mathrm{pre}}(u)+(1-\alpha)p_{\mathrm{ref}}(u),
  \qquad p_{\mathrm{ref}}=U(0,1).
  \label{eq:video_noise_resampling}
\end{equation}
Objective-specific importance weights balance video learning and predictive guidance:
\begin{equation}
  w_v(u)=\frac{p_{\mathrm{pre}}(u)}{q(u)},
  \qquad w_g(u)=\frac{p_{\mathrm{ref}}(u)}{q(u)},
  \qquad
  \mathcal{L}=\mathbb{E}_{u_v\sim q}\!\left[
    \lambda_v\bar w_v\ell_v+
    \bar w_g(\lambda_a\ell_a+\lambda_c\ell_c)
  \right].
  \label{eq:reweighted_joint_objective}
\end{equation}
Here, $\ell_v,\ell_a$ are flow-matching losses, and $\ell_c$ is masked-token cross-entropy; bars denote clipped weights normalized to unit batch mean. We use $\alpha=0.5$ and leave code-mask sampling unchanged.

\subsection{Agentic Coding in WAM}
\label{sec:skill_planning_code}

\begin{figure}[t]
  \centering
  \includegraphics[width=1.0\linewidth]{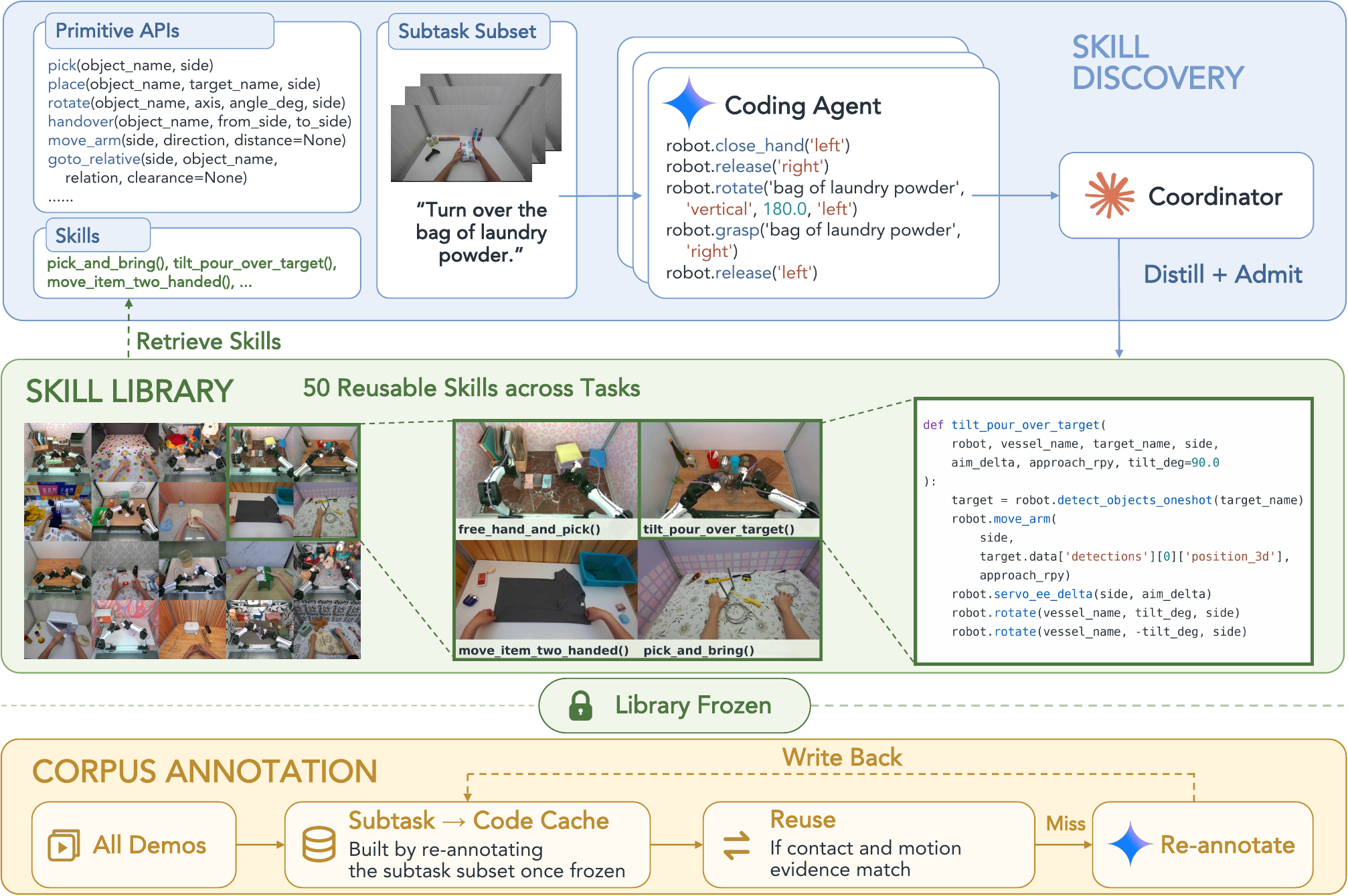}
  \caption{\textbf{Automatic Annotation Pipeline.} Coding agents generate subtask-level programs using predefined primitive APIs and available skills, while a coordinator distills recurring patterns into a library of reusable skills across tasks. With the library frozen, we annotate the corpus by reusing cached code when contact and motion evidence match and generating new labels otherwise.}
  \label{fig:pretain_labeling}
\end{figure}


\textbf{Automatic Annotation Pipeline.}
We propose an automatic annotation pipeline (Figure~\ref{fig:pretain_labeling}) that annotates the training data with skill planning code in three stages. We first sample one episode per task and use Gemini 3.1 Pro~\citep{google2026gemini31pro} to write subtask-level skill programs against predefined declaration-only APIs, while Claude Code running Claude Opus 4.8~\citep{anthropic2026claudeopus48} acts as a coordinator that distills recurring, task-general skills.
We then freeze the skill library and re-annotate the sampled episodes with the finalized vocabulary, forming a cache of reusable subtask-to-program mappings. To annotate the remaining corpus, we reuse a cached program when the subtask's contact and motion pattern matches an existing entry; otherwise, the coding agent generates a new annotation and adds it to the cache. See Appendix~\ref{app:agentic_annotation_details} for further pipeline details.


\textbf{Code Usage in Model.}
Annotated programs define the interface through which the code expert plans. 
During training, the code stream concatenates a conditioning prompt and the active slice, a short block of primitive or skill calls. Only the slice is corrupted by the absorbing-mask process of Eq.~\ref{eq:code_mask_corruption} and supervised by the masked-token loss. 
At each replanning point, the policy rebuilds the prompt and appends a canvas of $G$ mask tokens. The canvas is decoded jointly with the video and action states over the $T$ denoising steps, committing the $G/T$ most confident masked positions per step.
By default, the decoded slice is not executed. It serves as a subtask plan that conditions the action expert through joint denoising, while the predicted action chunk drives the robot.

\textbf{Memory Cache.}
Since a single observation is often insufficient to reveal task progress, the code expert receives a compact cache of the slices generated so far. The history slices are compressed into $K$ learnable memory slots and read back as prefix tokens before the prompt in the code stream. During training, the buffer holds the preceding annotated slices, replaced with probability $p$ by the model's own predictions. At inference, it is reset per episode and grows with each decoded slice.

\textbf{Hybrid Actions.}
We additionally explore an inference mode in which the decoded code slice is executed in the manner of Code-as-Policies (CaP) against a library of perception and control primitives, and the executed actions are integrated with the direct actions of the action expert. At deployment, the two channels are arbitrated at every replanning point: the decoded slice takes over the current stage only if it passes a set of validity and progress checks; otherwise the action chunk from the same forward pass drives the robot until the next replan. See Appendix~\ref{app:hybrid_action_algorithm} for the full algorithm.

\subsection{Training Recipe}
\label{sec:training_recipe}

We first pretrain \Ours{} on approximately 2{,}000 hours of YAM robot data and approximately 1{,}000 hours of egocentric human manipulation data to learn a general unified generative prior. We apply the automatic annotation pipeline in Section~\ref{sec:skill_planning_code} to extract two embodiment-specific skill libraries containing 50 skill entries in total (23 for YAM and 27 for egocentric human demonstrations) and annotate demonstrations with temporally aligned program slices. The annotated data support joint pretraining of code, video, and action generation, connecting structured skill plans with visual dynamics and continuous control across embodiments and scenarios. We further include the details of dataset statistics, preprocessing, and state-action representations in Appendix~\ref{app:analysis_pretrain_dataset}. 
For each downstream task, we reuse the same annotation pipeline under the frozen skill library to label task-specific demonstrations and jointly fine-tune \Ours{} on the resulting code, video, and action targets using the objective in Eq.~\ref{eq:reweighted_joint_objective}.

\section{Experiments}
\label{sec:experiment}

We evaluate \Ours{} on RoboTwin~2.0~\citep{chen2025robotwin20scalabledata} and 8 real-world tasks spanning parallel-gripper and dexterous-hand embodiments, assessing language following, semantic reasoning, and skill execution, as well as the robustness benefits of hybrid execution in variant scenarios.

\textbf{Baselines.}
We compare four baselines: (1) $\pi_{0.5}$~\citep{black2025pi05}, a state-of-the-art VLA with strong semantic reasoning; (2) EgoScale~\citep{zheng2026egoscale}, a VLA pretrained on large-scale egocentric human data; (3) DreamZero~\citep{ye2026dreamzero}, a WAM jointly generating future video and actions; and (4) WLA~\citep{yang2026worldlanguageactionmodelunifiedworld}, an autoregressive model jointly predicting textual subtasks, subgoal images, and actions. 
All baselines and the default \Ours{} configuration use the same task-specific demonstrations for post-training and follow the same evaluation protocol. The simulation hybrid configuration uses separately collected CaP-generated demonstrations, as described below. Baseline implementation details are provided in Appendix~\ref{app:baseline_details}.


\subsection{Simulation Experiment}
\label{sec:exp_sim}



\textbf{Benchmarks.}
We evaluate \Ours{} on RoboTwin~2.0.
Following the training-data requirements specified on the official RoboTwin~2.0 leaderboard\footnote{\url{https://robotwin-platform.github.io/leaderboard}}, we train on 50 clean demonstrations per task (2,500 episodes in total) and evaluate under the benchmark's two settings: \textit{clean}, with a clean tabletop, plain background, fixed lighting, and no distractors; and \textit{random}, with domain randomization over background and tabletop textures, lighting, tabletop height, and distractor objects. We report the mean success rate over 50 rollouts per task.

\textbf{Hybrid Evaluation.}
We additionally evaluate \Ours{} in the hybrid mode, in which the training demonstrations are collected by executing CaP programs written against a set of CaP-X-style~\citep{fu2026capx} primitive APIs that we develop for RoboTwin~2.0. The programs are written automatically by Claude Opus 4.8, following the agentic framework of ASPIRE~\citep{lu2026aspire}.

\begin{figure*}[t]
\centering
\begin{minipage}[c]{0.50\textwidth}
\centering
\captionof{table}{\textbf{RoboTwin~2.0 Results under \textit{Clean} and \textit{Random} Settings.}
Success rate (\%) averaged over 50 tasks with 50 rollouts each. Per-task numbers are reported in Table~\ref{tab:robotwin_pertask}.}
\label{tab:robotwin_main}
\footnotesize
\setlength{\tabcolsep}{5pt}
\adjustbox{max width=\linewidth}{%
\begin{tabular}{l|cc|c}
\toprule
Method & Clean & Random & Average \\
\midrule
$\pi_{0.5}$ & 54.9 & 27.7 & 41.3 \\
DreamZero   & 33.6 & 17.6 & 25.6 \\
EgoScale    & 26.2 &  2.9 & 14.5 \\
WLA         & 64.8 &  3.6 & 34.2 \\
\midrule
\textbf{\Ours{} (ours)}   & \textbf{80.8} & 20.6 & \textbf{50.7} \\
\textbf{\Ours{} (hybrid)} & 63.1 & \textbf{29.2} & 46.2 \\
\bottomrule
\end{tabular}}
\end{minipage}%
\hfill
\begin{minipage}[c]{0.48\textwidth}
\centering
\includegraphics[width=\linewidth]{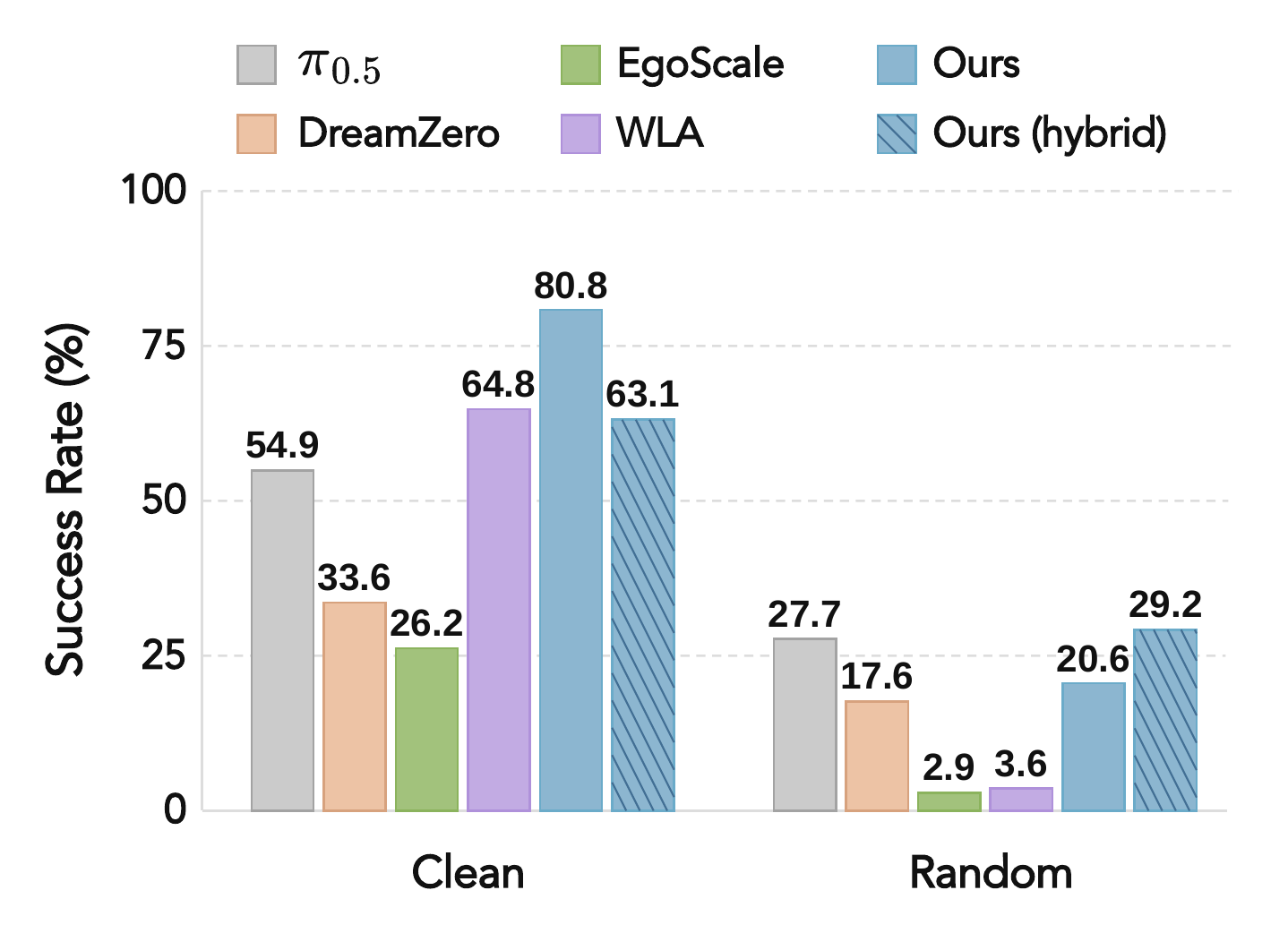}
\vspace{-0.6cm}
\captionof{figure}{\textbf{Average Scores from Table~\ref{tab:robotwin_main}.}}
\label{fig:robotwin_bars}
\end{minipage}
\end{figure*}

\textbf{Main Results.}
Table~\ref{tab:robotwin_main} and Figure~\ref{fig:robotwin_bars} report the results on RoboTwin 2.0. The two inference modes exhibit complementary strengths.
Overall, \Ours{} achieves the highest average success rate of 50.7\% across both settings, outperforming all baselines including models designed for broad generalization. 
The advantage comes mostly from the clean setting, where \Ours{} reaches 80.8\%, exceeding the strongest baseline WLA (64.8\%) by 16.0 points. Under visual randomization, performance drops to 20.6\%, consistent with the challenges of pixel-level O.O.D prediction discussed in prior work~\citep{yan2026robustwambridginggenerativepretraining}. It falls below the VLA baseline $\pi_{0.5}$ (27.7\%) but still outperforms the strongest WAM baseline DreamZero (17.6\%) and remains far ahead of EgoScale (2.9\%) and WLA (3.6\%).

We further observe that hybrid execution improves robustness under distribution shift. In the random setting, it raises the success rate from 20.6\% to 29.2\%, outperforming all baselines including the strong VLA baseline $\pi_{0.5}$ (27.7\%). Relative to its clean-setting performance, the hybrid mode retains $46\%$ of its success rate under randomization, compared with 25\% for the default mode. This robustness comes at a cost in the clean setting, where the hybrid mode reaches 63.1\%, on par with the strongest baseline, but 17.7 points below the default mode. For further analysis, see Appendix~\ref{app:additional_sim_experiment}.

\vspace{-0.2cm}

\subsection{Real-world Experiment}
\label{sec:exp_real}

\textbf{Robot Platforms and Action Space.}
We conduct real-world experiments on two platforms: the YAM station, a fixed-base bimanual setup with two 6-DoF arms and parallel grippers, and the Tianji--Sharpa platform, a bimanual robot with two 22-DoF Sharpa dexterous hands.
Both platforms observe three RGB views and act in the unified 64-D state--action space defined in Appendix~\ref{app:model_training_details}. Further details on the robot setups are provided in Appendix~\ref{app:real_world_setup}.

\textbf{Evaluation Protocol and Metrics.}
We use 100 demonstrations per task for post-training and evaluate on the 8 real-world tasks in Table~\ref{tab:real_main}, 4 on each platform, with $3\times16$ rollouts per task (3 training seeds, 16 evaluation rollouts per seed) under randomized initial object poses. Task definitions, evaluation settings, and per-task scoring rubrics are provided in Appendix~\ref{apdx:task_rubrics}.

\newcommand{\hdr}[1]{\begin{tabular}[b]{@{}c@{}}#1\end{tabular}}
\newcommand{\hdrcenter}[2][0]{%
  \begingroup
  \setbox0=\hbox{\hdr{Arrange\\Fruits}}%
  \setbox2=\hbox{M}%
  \dimen0=\ht0 \ifdim\dimen0<\ht\strutbox \dimen0=\ht\strutbox \fi
  \ifnum#1=1 \advance\dimen0 by \dimexpr\ht\strutbox+\dp\strutbox\relax \fi
  \smash{\raisebox{\dimexpr(\dimen0-\dp\strutbox+\belowrulesep-\aboverulesep-\ht2)/2\relax}{#2}}%
  \endgroup}
\newlength{\datacolw}

\begin{table}[t]
\centering
\caption{\textbf{Real-world Results on 8 Tasks across 2 Embodiments.} Average rubric score (\%) over $3\times16$ rollouts/task (Appendix~\ref{apdx:task_rubrics}).}
\label{tab:real_main}
\vspace{-3mm}
\small
\setlength{\tabcolsep}{1.5pt}
\settowidth{\datacolw}{Backpack}
\resizebox{\linewidth}{!}{%
\begin{tabular}{l|*{4}{w{c}{\datacolw}}|*{4}{w{c}{\datacolw}}|w{c}{\datacolw}}
\toprule
& \multicolumn{4}{c|}{\textit{YAM (parallel grippers)}}
& \multicolumn{4}{c|}{\textit{Tianji--Sharpa (dexterous hands)}}
& \\
\cmidrule(lr){2-5}\cmidrule(lr){6-9}
\hdrcenter[1]{Method}
& \hdr{Arrange\\Fruits}
& \hdr{Pack\\Backpack}
& \hdr{Replace\\Lightbulb}
& \hdr{Fold\\Towels}
& \hdr{Fruits\\w/ Tong}
& \hdr{Deal\\Poker}
& \hdr{Flip\\Scoreboard}
& \hdr{Solve\\Cube}
& \hdrcenter[1]{Avg.} \\
\midrule
$\pi_{0.5}$             & 70.8 & 49.7 & 63.8 & 81.7 & 7.3 & 13.1 & 26.3 & 36.1 & 43.6 \\
DreamZero               & 53.3 & 36.7 & 67.4 & 79.8 & 22.6 & 37.1 & 28.4 & 42.5 & 46.0 \\
EgoScale                & 37.5 & 27.3 & 60.2 & 53.3 & 17.6 & 25.6 & 40.8 & 24.9 & 35.9 \\
WLA                     & 39.6 & 29.6 & 56.0 & 74.6 & 14.2 & 22.9 & 29.7 & 52.3 & 39.9 \\
\midrule
\textbf{\Ours{} (ours)} & \textbf{91.7} & \textbf{64.6} & \textbf{75.0} & \textbf{84.4} & \textbf{54.2} & \textbf{58.3} & \textbf{68.8} & \textbf{72.1} & \textbf{71.1} \\ \bottomrule
\end{tabular}}
\end{table}

\vspace{-0.1cm}

\textbf{Quantitative and Qualitative Analysis.}
As shown in Table~\ref{tab:real_main}, \Ours{} achieves the highest average rubric score across both embodiments, outperforming the strongest baseline by 25.1 points. Two observations emerge.
\textbf{Language following and semantic reasoning:}
\Ours{} outperforms the baselines strongly on \textit{Arrange Fruits} and \textit{Flip Scoreboard}, which require instruction-specific object selection and counting, respectively. The VLA baselines, particularly $\pi_{0.5}$, also show relatively strong semantic understanding, while the WAM baselines are less reliable on these properties. These results suggest that code-based planning strengthens the language-following and semantic reasoning of world action models.
\textbf{Robust skill execution:}
On contact-rich and dexterous tasks, DreamZero outperforms $\pi_{0.5}$ on \textit{Replace Lightbulb} and \textit{Deal Poker}, where predicting future states provides stable physical-dynamics features for action prediction. \Ours{} further outperforms these baselines by combining guidance from generated skill programs with predictive representations of physical dynamics, aligning skill-level planning with continuous control. See Appendix~\ref{app:additional_real_results} for further analysis.

\textbf{Hybrid Execution in the Real World.}
We further evaluate hybrid execution on \textit{Fold and Stack Towels} with the YAM platform based on Algorithm~\ref{app:hybrid_action_algorithm}, comparing it against action-chunk execution under the same post-trained model and evaluation conditions. Under the same task rubric, hybrid execution achieves \textbf{90.6\%}, compared with \textbf{84.4\%} for action-chunk execution, while the direct execution steps of code further enhance the model's reasoning capabilities in distinguishing stacking order and action stability. Further visualizations are provided in Appendix~\ref{app:hybrid_action_eval}.


\begin{figure}[t]
\centering
\includegraphics[width=\linewidth]{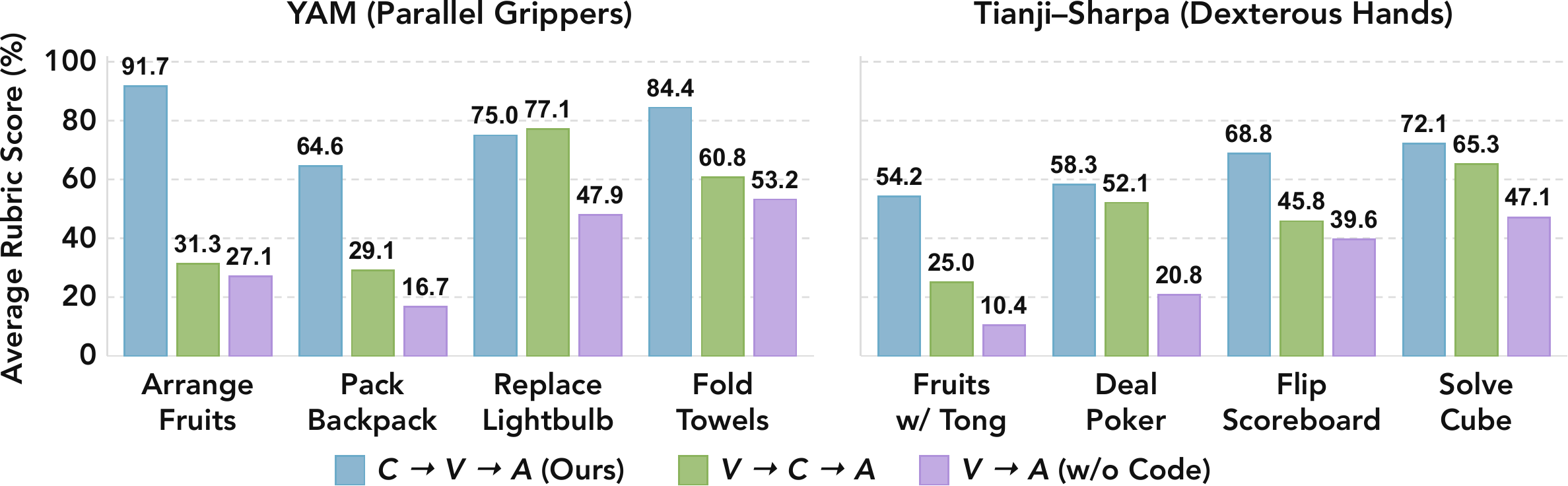}
\caption{\textbf{Impact of Code Expert and Attention Structure.} Avg. rubric score (\%) over $3\times16$ rollouts/task. $C$, $V$, and $A$ denote the code, video, and action streams.}
\label{fig:abl_arch}
\end{figure}

\subsection{Ablation Study}
\label{sec:exp_ablation}

We organize our ablation study mainly around three research questions: 1) how to integrate a visually grounded code expert into a WAM framework, 2) which components of the planning representation contribute to task performance, and 3) how joint denoising affects task performance and inference efficiency. All the variants are post-trained and evaluated under the protocol of Section~\ref{sec:exp_real}.

\suppressfloats[t]\textbf{Q1: How should the code expert be integrated into a WAM?}
We first compare three blockwise causal attention orders during joint denoising: $V\to A$ (w/o code), $V\to C\to A$, and $C\to V\to A$ (Figure~\ref{fig:abl_arch}). The two variants use the same pretraining data, training budget, and optimization settings as \Ours{}. Adding code through $V\to C\to A$ improves performance by grounding the predicted dynamics to action, but restricts its role to reasoning over video predictions without guiding or correcting them. With $C\to V\to A$, both video and action generation attend to code, allowing semantic task plans to shape future prediction throughout denoising. The performance gain of this ordering supports introducing semantic reasoning directly into video generation, which improves language following and provides more reliable guidance for stable action generation.

\vspace{-0.1cm}
\begin{table}[H]
\centering
\caption{\textbf{Visual Conditioning for Code Expert.} Avg. rubric score (\%) over $3\times16$ rollouts/task.}
\label{tab:abl_visual}
\vspace{-3mm}
\small
\setlength{\tabcolsep}{1.5pt}
\settowidth{\datacolw}{Backpack}
\begin{tabular}{l|cccc|cccc|w{c}{\datacolw}}
\toprule
\hdrcenter{Variant}
& \hdr{Arrange\\Fruits}
& \hdr{Pack\\Backpack}
& \hdr{Replace\\Lightbulb}
& \hdr{Fold\\Towels}
& \hdr{Fruits\\w/ Tong}
& \hdr{Deal\\Poker}
& \hdr{Flip\\Scoreboard}
& \hdr{Solve\\Cube}
& \hdrcenter{Avg.} \\
\midrule
\textbf{ViT (ours)} & \textbf{91.7} & \textbf{64.6} & \textbf{75.0} & \textbf{84.4} & \textbf{54.2} & \textbf{58.3} & \textbf{68.8} & \textbf{72.1} & \textbf{71.1} \\
VAE                 & 83.3 & 39.6 & 66.7 & 61.3 & 41.7 & 47.9 & 50.0 & 53.9 & 55.6 \\
\bottomrule
\end{tabular}
\end{table}
\vspace{-0.1cm}

We then compare a dedicated ViT against reused Wan2.2 VAE features for conditioning the code expert (Table~\ref{tab:abl_visual}). ViT conditioning improves performance on all 8 tasks, with substantial gains on \textit{Pack Backpack} (64.6\% vs.\ 39.6\%) and \textit{Solve Rubik's Cube} (72.1\% vs.\ 53.9\%), which require fine-grained semantic reasoning about object attributes and task states. VAE features are optimized for pixel-level reconstruction~\citep{yan2026robustwambridginggenerativepretraining} and thus poorly support the semantic distinctions needed to select and compose skills, whereas ViT features enable the code expert to translate visual observations into more accurate skill plans.

\begin{figure*}[t]
\centering
\begin{minipage}[c]{0.53\textwidth}
\centering
\captionof{table}{\textbf{Impact of Code Generation Paradigm.} Avg. rubric score (\%) over $3\times16$ rollouts/task.}
\label{tab:abl_gen}
\footnotesize
\vspace{-2mm}
\setlength{\tabcolsep}{5pt}
\adjustbox{max width=\linewidth}{%
\begin{tabular}{l|cc}
\toprule
Task & \textbf{Joint denoise (ours)} & AR \\
\midrule
Arrange Fruits    & \textbf{91.7} & 81.3 \\
Pack Backpack     & \textbf{64.6} & 62.5 \\
Replace Lightbulb & 75.0 & \textbf{79.2} \\
Fold Towels       & \textbf{84.4} & 53.2 \\
\midrule
Fruits w/ Tong    & \textbf{54.2} & 20.8 \\
Deal Poker        & \textbf{58.3} & 44.1 \\
Flip Scoreboard   & \textbf{68.8} & 39.3 \\
Solve Cube        & \textbf{72.1} & 55.8 \\
\midrule
Avg.              & \textbf{71.1} & 54.5 \\
\bottomrule
\end{tabular}}
\end{minipage}%
\hfill
\begin{minipage}[c]{0.45\textwidth}
\centering
\includegraphics[width=0.9\linewidth]{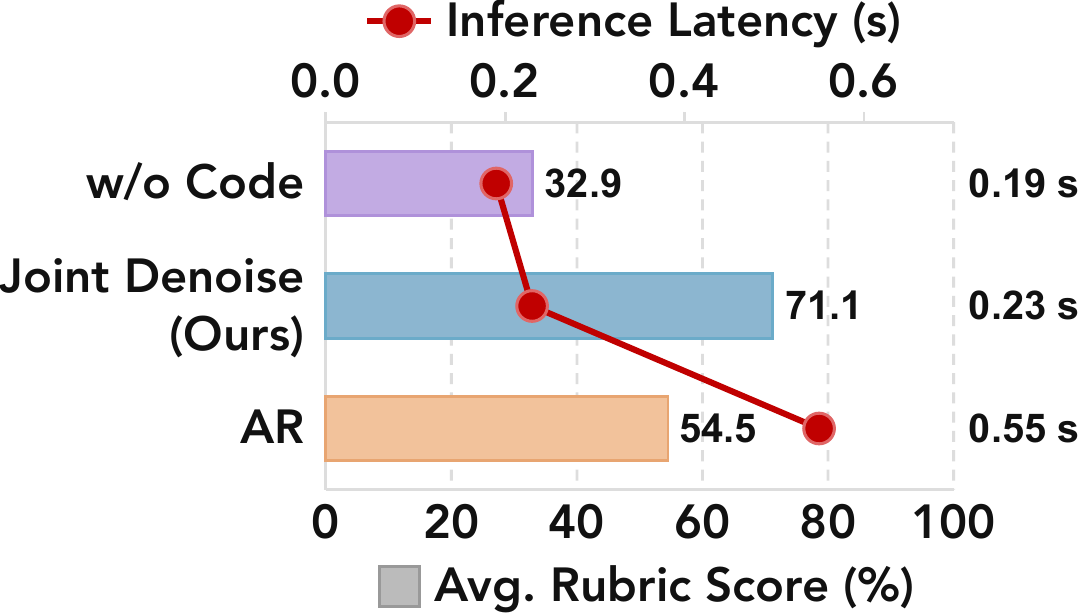}
\captionof{figure}{\textbf{Performance and Inference Latency.} Average rubric score over the 8 real-world tasks (bars) and inference latency on an NVIDIA RTX 5090 (red line).}
\label{fig:abl_gen_speed}
\end{minipage}
\end{figure*}

\begin{figure*}[t]
\centering
\includegraphics[width=\textwidth]{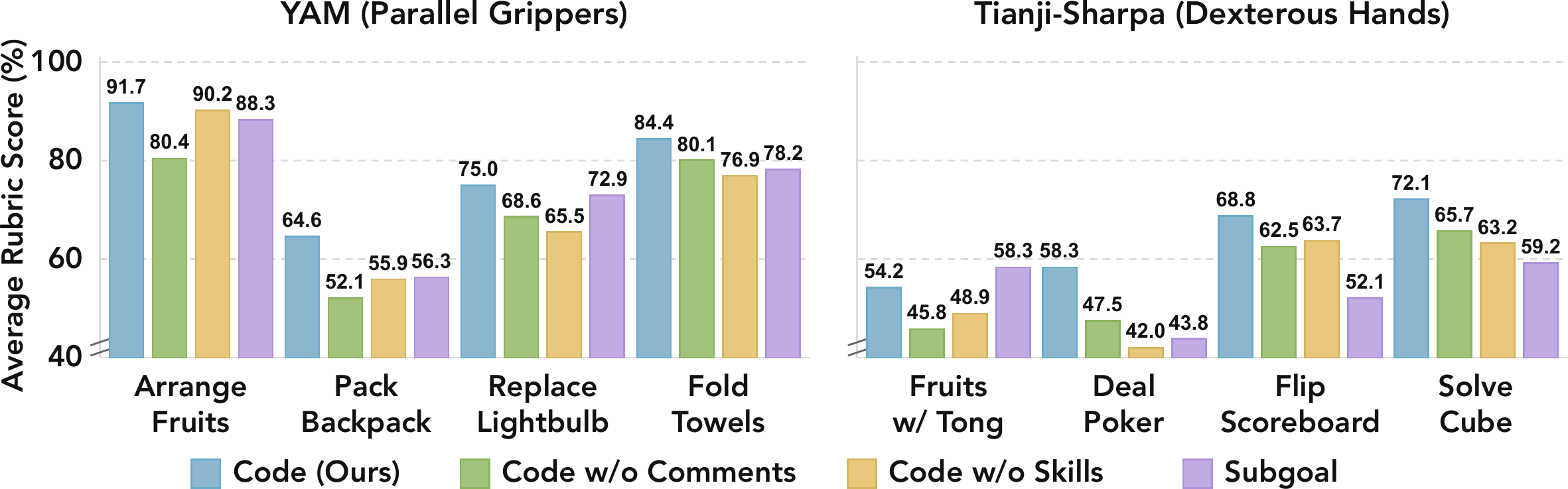}
\caption{\textbf{Impact of Code Label Content.} Average rubric score (\%) over $3\times16$ rollouts per task.}
\label{fig:abl_label}\vspace{-10pt}\end{figure*}

\textbf{Q2: What makes code an effective planning representation?}
To assess the planning content produced by our annotation pipeline, we compare \textit{(1) full annotated programs}, \textit{(2) code without comments}, \textit{(3) code without skills}, and \textit{(4) natural-language subgoals} in Figure~\ref{fig:abl_label}. 
Full programs outperform comment-free code on all 8 shared tasks, suggesting that explanatory language helps connect program structure with task intent. Removing skill calls also reduces performance on all 8 evaluated tasks, especially on skill-required tasks such as \textit{Deal Poker} (58.3\% vs.\ 42.0\%). Since this variant still uses structured code, the drop supports the value of reusable skill abstractions beyond primitive-level API composition. Full programs also beat language-only subgoals on 7 of 8 shared tasks, showing that structured semantic guidance can provide more action-centric representation as condition. Together, these comparisons support combining explanatory language, reusable skills, and explicit program structure to provide effective planning guidance.


\textbf{Q3: How does joint denoising affect performance and efficiency?}
As shown in Table~\ref{tab:abl_gen}, joint denoising outperforms AR on 7 of 8 tasks. For AR models trained with teacher forcing, the mismatch between ground-truth training prefixes and generated inference prefixes can cause errors to accumulate during inference and affect downstream video prediction. Joint denoising reduces token-by-token serial dependence by reconstructing code alongside video and actions, providing semantic guidance throughout generation.
We measure end-to-end inference latency on an NVIDIA RTX 5090, with settings detailed in Appendix~\ref{app:model_training_details}. Joint denoising takes 0.23~s versus 0.55~s for AR, a $2.39\times$ speedup (Figure~\ref{fig:abl_gen_speed}). Compared with the \Ours{} w/o code (0.19~s), it adds 0.04~s (21.1\%) while improving performance on all 8 shared tasks (Figure~\ref{fig:abl_arch}). 

\section{Related Work}
\label{sec:related_work}

\textbf{World Action Models (WAMs).}
World models predict future observations and have recently emerged as a powerful paradigm for robot manipulation, serving as learned simulators and data generators for policy learning~\citep{agarwal2025cosmosworldfoundationmodelplatform,ali2025worldsimulationvideofoundationmodels,gao2026dreamdojo,yang2023learninginteractiverealworldsimulators,jang2025dreamgenunlockinggeneralizationrobot}. Building on this progress, World Action Models (WAMs) unify future prediction and action generation, through joint video-action diffusion~\citep{ye2026dreamzero,zhu2025unifiedworldmodelscoupling}, expert-based architectures~\citep{bi2025motusunifiedlatentaction,cai2026internvlaa1unifyingunderstandinggeneration}, or direct adaptation of pretrained video models into policies~\citep{kim2026cosmospolicyfinetuningvideomodels}. Recent WAMs further couple autoregressive sub-task prediction with future-state modeling and action synthesis~\citep{yang2026worldlanguageactionmodelunifiedworld} or dynamically update subtask instructions to condition video and action generation~\citep{tang2026dreamingnecessaryadvancingworld}.
However, existing WAMs lack structured semantic guidance and explicit skill invocation, limiting their ability to perform complex semantic planning and structure predicted dynamics into actionable behavioral abstractions.

\textbf{Agentic Robotics.} Early work composed perception and control APIs into executable closed-loop behaviors without policy training~\citep{ahn2022saycan,liang2023code,singh2023progprompt}. Later systems generate full Python programs over perception APIs~\citep{huang2023instruct2act,mu2024robocodex,goldberg2025bloxnet}, and executable code has proven a stronger action space for LLM agents than text~\citep{wang2024codeact}. With the rise of coding agents, program synthesis has become explicitly agentic, with recent systems iteratively writing, executing, and repairing control programs~\citep{lu2026aspire,fu2026capx}. Frontier models now also act as controllers themselves, reasoning over instructions and feedback and driving robots directly through native tool calls and learned primitives~\citep{shi2025hirobot,li2026roboclaw,geminirobotics2025,geminiroboticser2025}. In all of these systems, the agent controls the robot from outside the policy and is bounded by the primitives it can call. We instead bring code inside an end-to-end model as a high-level planning interface and further explore how the generated code can be executed to complement the model's own actions at test time.


\section{Conclusion}
\label{sec:conclusion}

We present \Ours{}, an agentic world action model that integrates code-based skill planning with predictive dynamics and continuous control. Through joint denoising, the code expert provides structured semantic guidance for video and action generation. An automated annotation pipeline enables learning this planning interface from large-scale robot and human demonstrations. Experiments on RoboTwin~2.0 and real-world manipulation with YAM and dexterous hands demonstrate improvements in language following, semantic reasoning, and skill execution, with hybrid execution further enhancing robustness. These results establish a unified foundation for future work on skill reuse and iterative interaction between programmatic planning, world prediction, and execution, providing a potential agentic system foundation that continually interacts with and learns from environments.

\section*{Acknowledgments}

Zhuoyang Liu thanks Prof. Shanghang Zhang at Peking University for her mentorship and support.

\section*{AI use statement}

We used Gemini 3.1 Pro and Claude Code with Claude Opus 4.8 in our automatic annotation pipeline to generate demonstration-aligned program labels and distill reusable skill libraries (Section~\ref{sec:skill_planning_code}). Claude Opus 4.8 was also used to generate executable programs for collecting simulation demonstrations (Section~\ref{sec:exp_sim}). Additionally, we used GPT-6 Astra for language polishing and stylistic refinement, as well as part of figure preparation. 

\section*{Ethics statement}

This work studies code-guided world action models for robotic manipulation using robot and egocentric human demonstrations. All pretraining datasets are used in accordance with their respective licenses and data-use terms. All post-training demonstrations are collected by our team for the tasks evaluated in this work.

\section*{Reproducibility statement}

We provide detailed information in Appendix~\ref{app:model_training_details} regarding the model architecture design, parameter settings, and computational resources. Regarding training data, Appendix~\ref{app:analysis_pretrain_dataset} details the composition of the pre-training dataset. Section~\ref{sec:skill_planning_code} introduces the complete data annotation pipeline. While prompts for the agentic framework and performance analysis of the pre-trained model are provided in Appendix~\ref{app:agent-prompts} and Appendix~\ref{app:pretrained_code_quality}. For downstream tasks, we also present standardized data generation criteria and methods to facilitate extension to any further tasks.


\bibliography{iclr2027_conference}
\bibliographystyle{iclr2027_conference}

\newpage
\appendix
\addtocontents{toc}{\protect\setcounter{tocdepth}{2}} \begingroup \renewcommand{\contentsname}{Appendix Contents} \tableofcontents \endgroup \clearpage


\section{Model and Training Details}
\label{app:model_training_details}

We implement the \Ours{} architecture using three modality-specific experts coupled by masked joint attention. All experts have 28 transformer layers and project their queries, keys, and values into a common 2{,}048-dimensional attention space. The code expert receives current visual observations and language instructions, the video expert attends to the code stream, and the action expert attends to both. Tables~\ref{tab:model_arch} and~\ref{tab:training_param} summarize the architecture and the pretraining configuration.

\textbf{State and Action Representation.}
Both YAM robot data and egocentric human data are mapped into a shared 64-dimensional state--action space. Each side contains a 9-dimensional end-effector pose (3D of $x,y,z$ position and a 6D rotation representation), 22 dexterous-hand joint angles, and one parallel gripper width, giving $2\times(9+22+1)=64$ dimensions. States are all absolute; while action chunks use end-effector pose deltas relative to the first frame of the chunk, with absolute gripper targets for YAM and absolute hand-joint targets for human demonstrations/dexterous hand. Dimensions unused by an embodiment---the hand-joint blocks for YAM and the gripper dimensions for human demonstrations---are set to zero and left unnormalized. All remaining dimensions are normalized using per-dimension 1st and 99th percentiles.

\textbf{Initialization and Trainable Components.}
We initialize the video DiT from Cosmos-Predict2.5-2B~\citep{ali2025worldsimulationvideofoundationmodels} and the action DiT from scratch before joint pretraining. The code expert initializes its token embeddings, layer normalizations, FFNs, and vocabulary head from SDAR-1.7B~\citep{cheng2025sdarsynergisticdiffusionautoregressionparadigm} and freezes them, while training its MoT attention projections and modulation parameters. The Qwen3-VL~\citep{Qwen3-VL} vision tower is frozen, with a trainable merger mapping visual features to code embeddings. We replace the original VAE with a frozen Wan2.2~\citep{wan2025wanopenadvancedlargescale} VAE with 48 latent channels and reinitialize the video DiT's input and output projections. The video DiT uses $1\times2\times2$ latent patches and multi-view 3D RoPE. Video and action experts use precomputed Cosmos-Reason1 language embeddings. Downstream adaptation transfers shape-compatible parameters from joint pretraining and reinitializes action and state projections when embodiment dimensions change.

\noindent\begin{minipage}{\linewidth}
\begin{center}
\captionof{table}{\textbf{Expert Architectures and Shared Visual Inputs.}}
\label{tab:model_arch}
\begingroup

\normalsize
\setlength{\tabcolsep}{4pt}
\renewcommand{\arraystretch}{1.14}
\begin{tabular}{@{}>{\raggedright\arraybackslash}p{0.28\linewidth} >{\centering\arraybackslash}p{0.19\linewidth} >{\centering\arraybackslash}p{0.28\linewidth} >{\centering\arraybackslash}p{0.19\linewidth}@{}}
\toprule
\textbf{Configuration} & \textbf{Code} & \textbf{Video} & \textbf{Action} \\
\midrule
Backbone & SDAR-1.7B & Cosmos-Predict2.5-2B & Action DiT \\
Residual width & 2,048 & 2,048 & 1,024 \\
FFN width & 6,144 & 8,192 & 4,096 \\
Attention & Bidirectional & Bidirectional & Bidirectional \\
\midrule
Transformer layers & \multicolumn{3}{c}{28} \\
Query / KV heads & \multicolumn{3}{c}{16 / 16} \\
Head / attention width & \multicolumn{3}{c}{128 / 2,048} \\
\midrule
ViT encoder & \multicolumn{3}{c}{Qwen3-VL-2B tower} \\
Vision depth / width & \multicolumn{3}{c}{24 / 1,024} \\
Spatial patch / merge & \multicolumn{3}{c}{$16\times16$ / $2\times2$ } \\
Visual tokens / width & \multicolumn{3}{c}{210 / 2,048} \\
Views / resolution & \multicolumn{3}{c}{3 / $224\times320$} \\
Video frames / stride & \multicolumn{3}{c}{9 / 4} \\
State / action width & \multicolumn{3}{c}{64 / 64 (Real-world); 14 / 14 (RoboTwin)} \\
Action horizon & \multicolumn{3}{c}{32} \\
\bottomrule
\end{tabular}
\endgroup
\end{center}

\end{minipage}\par\textbf{Observation and Program Representations.}
Images are sampled at timestep offsets $0,4,\ldots,32$, giving nine frames per view; only the first frame is used as observation conditioning. The code expert's visual encoder also receives only this first frame from each view. At $224\times320$ resolution, spatial merging produces 70 visual tokens per view. The code expert uses the Qwen3-family tokenizer of the SDAR checkpoint. Its 512-token training budget applies to the instruction and program together, with the 210-token visual prefix added separately. At inference, the default program canvas contains 128 tokens. Visual-prefix, instruction, and padding positions are excluded from the standard masked-program loss.

\begin{center}
\captionof{table}{\textbf{Generation and Pretraining Settings.} Downstream overrides are given in the text.}
\label{tab:training_param}
\begingroup

\normalsize
\setlength{\tabcolsep}{4pt}
\renewcommand{\arraystretch}{1.14}
\begin{tabular}{@{}>{\raggedright\arraybackslash}p{0.25\linewidth} >{\centering\arraybackslash}p{0.22\linewidth} >{\raggedright\arraybackslash}p{0.25\linewidth} >{\centering\arraybackslash}p{0.22\linewidth}@{}}
\toprule
\multicolumn{2}{c}{\textbf{Generation}} & \multicolumn{2}{c}{\textbf{Optimization}} \\
\cmidrule(lr){1-2}\cmidrule(lr){3-4}
Video / Action loss & Flow matching & Optimizer & AdamW \\
Code loss & Masked CE & Adam coefficients & $(0.9,0.95)$ \\
Video mixture $\alpha$ & 0.5 & Peak learning rate & $10^{-4}$ \\
Sigma shifts $(v,a)$ & $(5,5)$ & LR schedule & Cosine \\
Action $u_a$ / code $r$ & $U(0,1)$ & Linear warmup & 5\% \\
Loss weights $(v,c,a)$ & $(1,0.5,1)$ & Weight decay & 0 \\
Importance clip & 4.0 & Gradient clip & 1.0 \\
Weight normalization & Unit batch mean & Precision / ZeRO & bf16 / Stage 1 \\
Mixture sampling & Stratified & Checkpointing & Enabled \\
Code / video dropout & $0.15$ / $0$ & Seed & 42 \\
Video / action solver & UniPC / Euler & Nodes / GPUs & 6 / 48 \\
Code order & Confidence & Batch per GPU & 6 \\
Reveal $\rho$ / warmup & 1 / 0 & Accumulation & 9 \\
Timestep scale & 1,000 & Effective batch & 2,592 \\
Code canvas & 128 tokens & Training budget & 1 epoch \\
Joint denoising steps & 4 & GPU model & NVIDIA H100 \\
\bottomrule
\end{tabular}
\endgroup
\end{center}

\textbf{Joint Training and Decoding.}
Video noise resampling follows Section~\ref{sec:joint_denoising}. The pretrained variable is $u_{\mathrm{pre}}=0.999b$, where $b\sim\mathrm{Beta}(7,1)$; training samples are drawn from $q=0.5p_{\mathrm{pre}}+0.5U(0,1)$. The video objective uses the pretrained-density weight, while action and code losses use the uniform-reference weight. Action noise levels and code-mask ratios are sampled independently of the video variable. At least one program token is masked in every nonempty training target. The pretraining recipe additionally replaces the instruction and program inputs with mask tokens for 15\% of training batches, retaining the visual prefix. The default decoder progressively commits high-confidence code predictions across the full joint trajectory. Early completion is controlled by $\rho<1$, a separate decoding-time choice that does not change the training mask distribution. Latency is measured end to end on an NVIDIA RTX 5090, including visual encoding, joint denoising, video decoding, and action postprocessing.

\textbf{Downstream Optimization.}
Downstream runs use the same optimizer, warmup schedule, and mixed-precision settings as pretraining, with task-specific batch sizes and training budgets. For both YAM and Tianji--Sharpa, we fine-tune \Ours{} for 20 epochs using 100 demonstrations per task. The YAM configurations use a per-GPU batch size of four, one gradient accumulation step, a learning rate of $10^{-4}$, and weight decay of $10^{-2}$. The RoboTwin configuration uses two 8-GPU nodes with a per-GPU batch size of $6$ and one accumulation step, giving an effective batch size of $96$, a learning rate of $2\times10^{-4}$, weight decay of $10^{-2}$, and a budget of 100,000 optimizer steps. These settings are distinct from the larger pretraining batch in Table~\ref{tab:training_param}.

\section{Analysis on Pretraining Dataset}
\label{app:analysis_pretrain_dataset}

\subsection{Pretraining Dataset Statistics}
\label{apdx:pretrain_stats}

We pretrain \Ours{} on the union of two real-world manipulation datasets: YAM and Egocentric Human data. The YAM corpus contains 73{,}101 episodes across 279 tasks, totaling 1{,}919.3 hours of bimanual robot teleoperation. Demonstrations are recorded at 30\,Hz with left-wrist, right-wrist, and top-view RGB cameras. The egocentric human corpus contains 175{,}541 episodes across 327 tasks, totaling 993.9 hours, with a single head-mounted RGB camera and tracked hand motion retargeted to 22-DoF dexterous-hand joint targets per hand. Together, the two sources comprise 248{,}642 episodes and 2{,}913.2 hours. The human camera view is replicated across the three input slots; each training window is paired with its episode-level instruction and aligned program slice. Table~\ref{tab:pretrain_totals} reports detailed statistics.

\begin{table}[h]
\centering\small
\caption{\textbf{Pretraining Corpus Statistics.}}
\label{tab:pretrain_totals}
\begin{tabular*}{\linewidth}{@{\extracolsep{\fill}}l r r r@{}}
\toprule
 & YAM & Egocentric Human & Total \\
\midrule
Episodes & 73{,}101 & 175{,}541 & 248{,}642 \\
Durations (h) & 1{,}919.30 & 993.91 & 2{,}913.22 \\
Tasks & 279 & 327 & 606 \\
Total subtask slices & 986{,}314 & 1{,}089{,}059 & 2{,}075{,}373 \\
Mean subtask slices per episode & 13.5 & 6.2 & 8.3 \\
\bottomrule
\end{tabular*}
\end{table}

\subsection{Skill Library Definitions}
\label{apdx:skill_library}

A skill is a reusable motion that the model can invoke by name. Each skill is a Python function written over the primitive API described in Appendix~\ref{app:low_level_primitives}, taking the objects it acts on as arguments, so that a subtask recurring across the corpus becomes a single call rather than being spelled out primitive by primitive. Following the annotation pipeline of Section~\ref{sec:skill_planning_code}, we generate a skill library for each dataset: 23 skills for YAM and 27 skills for the egocentric human data. Tables~\ref{tab:skills_yam} and~\ref{tab:skills_egocentric_human} list them, each with a one-line description of the motion it implements.

{\footnotesize
\begin{longtable}{p{0.38\linewidth} p{0.55\linewidth}}
\caption{\textbf{The Frozen YAM Skill Library} ($23$ skills).}
\label{tab:skills_yam} \\
\toprule
Skill & Definition \\
\midrule
\endfirsthead
\toprule
Skill & Definition \\
\midrule
\endhead
\texttt{adjust\_in\_place} & Grip an object where it lies, slide it a few centimetres without lifting, and let go. \\
\texttt{alternating\_smoothing\_pats} & Press the fabric flat with one arm and come off, then the other --- the alternating pats of a smoothing slice. \\
\texttt{approach\_best\_grasp} & Sample grasp candidates on an object, pick the best one reachable by this arm, and drive the end-effector onto it. \\
\texttt{brace\_and\_pull\_open} & Brace the body with one arm, take its lid, flap or handle with the other, draw it open, then release the brace. \\
\texttt{brace\_and\_push\_in} & Grip the same part twice, low to keep it aligned and high to push, and drive it home into its receptacle. \\
\texttt{flip\_object\_over} & Lift a held flat object clear, roll the wrist a half turn, release it inverted, then grasp it again to pin it down. \\
\texttt{fold\_edge\_over} & Both arms pinch a cloth edge, carry it across the item, and release --- one fold. \\
\texttt{free\_hand\_and\_pick} & Open the idle arm, then pick the object up with the working arm. \\
\texttt{hand\_over\_hand\_extract} & Two grip--pull--release bites that walk a long object out of a bag, sleeve or slot. \\
\texttt{lower\_onto\_target} & Hover the carried object above a detected target at the current wrist orientation and lower it in; the caller releases. \\
\texttt{pull\_part\_free} & Grip a part where it sits, drag it straight off along one bounded delta, and release. \\
\texttt{push\_closure\_shut} & Grip a drawer, lid, door or cover and drive it along its travel until shut, then let go. \\
\texttt{ratchet\_turn\_fastener} & Two ratchet strokes on a threaded fastener; the release in between is the wrist unwinding, not a drop. \\
\texttt{ratchet\_twist\_off} & Two ratchet bites on a threaded part; the sign of the angle sets whether it tightens or loosens. \\
\texttt{relocate\_object} & One-arm transport inside a single slice: pick the object up and place it at the named target. \\
\texttt{roll\_up\_one\_bite} & One reach-limited bite of rolling: pinch, push the roll forward, release, and travel back to the next pinch point. \\
\texttt{run\_zipper\_along\_seam} & Pin the case with one arm, take the zipper pull with the other, and draw it one run along the seam. \\
\texttt{scrub\_back\_and\_forth} & Bring an already-held cloth to a detected surface and scrub it once back and forth. \\
\texttt{spread\_taut\_and\_release} & With both arms already gripping, drag them apart to pull the item taut, then release both. \\
\texttt{stir\_braced\_vessel} & Stir inside a braced vessel, re-seating the brace once the tool tip starts loading the wall. \\
\texttt{tear\_off\_piece} & Pin the parent sheet, roll or bag with one arm, pinch the piece with the other, and pull along the tear line. \\
\texttt{tilt\_pour\_over\_target} & Carry a held vessel over a detected receptacle, clear the rim, tip once, and return to upright. \\
\texttt{wrap\_tension\_pull} & Two hand-over-hand bites on a strand being wound: regrasp further along and pull the slack taut, twice. \\
\bottomrule
\end{longtable}
}

{\footnotesize
\begin{longtable}{p{0.38\linewidth} p{0.55\linewidth}}
\caption{\textbf{The Frozen Egocentric Human Skill Library} ($27$ skills).}
\label{tab:skills_egocentric_human} \\
\toprule
Skill & Definition \\
\midrule
\endfirsthead
\toprule
Skill & Definition \\
\midrule
\endhead
\texttt{align\_and\_seat\_braced} & One hand pins the target so it cannot shift while the other pushes the part home and lets go. \\
\texttt{brace\_and\_pick} & Hold the source still with one hand and pick the object off or out of it with the other. \\
\texttt{brace\_and\_take\_tool} & Set up the two-handed working stance: the brace hand closes on the workpiece, the working hand on the tool. \\
\texttt{close\_closure} & Swing a closure shut with both hands on it and come off it; nothing is held afterwards. \\
\texttt{flatten\_by\_pulling\_taut} & Pin a slack flat item with one hand, drag it taut with the other, re-bite further along, and drag again. \\
\texttt{flip\_over\_between\_hands} & Turn an object face-over by passing it between the hands through a half turn. \\
\texttt{fold\_over\_and\_crease} & Brace a flat item, carry its free edge across in two steps, set it down, and press the crease --- one fold. \\
\texttt{hang\_item\_on\_support} & Bring a held item over a hook, rack or branch from above and let it down onto it. \\
\texttt{insert\_into\_receptacle} & Seat the held object into a receptacle, let go, and withdraw the now-empty hand. \\
\texttt{move\_item\_two\_handed} & Take a bulky item with both hands where it sits, carry it along a semantic direction, and release with both. \\
\texttt{open\_closure} & Swing a hinged closure open with one hand while the other braces the container body, then come off it. \\
\texttt{pick\_and\_bring} & Pick an object up and carry it to a semantic relation on where it is going next; the hand keeps hold of it. \\
\texttt{place\_at\_target} & Set the carried object down at a target with the working hand and free the second hand as well. \\
\texttt{pour\_into\_target} & Bring a held vessel's mouth over a target, tip it until it empties, and return it upright; the grip never changes. \\
\texttt{press\_onto\_target} & Press a target with both hands and come off it --- seating a lid or case home, or working a control. \\
\texttt{pull\_open\_and\_hold} & Draw a handle, leaf or free end open along a semantic direction and keep hold of it. \\
\texttt{pull\_part\_free} & Brace an assembly with one hand and pull a part off or out of it with the other. \\
\texttt{rotate\_grasped\_object} & Take an object where it sits and turn it about a semantic axis; the hand ends still on the object. \\
\texttt{scoop\_and\_carry} & Take a load out of a source container with a held utensil and bring it over the destination. \\
\texttt{scrub\_inside\_vessel} & One hand holds the vessel still while the other puts a brush down inside it and works a stroke pair. \\
\texttt{stir\_contents} & Drop the already-held tool inside the vessel and sweep it round. \\
\texttt{thread\_through\_and\_pull\_out} & Push a lace, cord or pin through a hole with one hand and catch the emerging end with the other to draw the slack through. \\
\texttt{tighten\_with\_ratchet\_regrasps} & Take up the last of a thread on an already-seated cap: turn, release the bite, re-bite, and turn again. \\
\texttt{turn\_over\_object} & Flip a held object face-over and bring the new face to a semantic relation on its destination. \\
\texttt{unscrew\_cap} & Twist a threaded part, let the wrist unwind and re-close, twist again, and lift the freed part clear. \\
\texttt{wipe\_surface\_back\_and\_forth} & Park a cloth or sponge already in one hand on an open surface and run two out-and-back strokes. \\
\texttt{wrap\_around\_target} & Lead a held wrapper --- ribbon, foil, film, gauze --- once around a target in a closed circuit. \\
\bottomrule
\end{longtable}
}

\subsection{Visualizations of YAM Skills}

We visualize four of the 23 YAM skills. For each, we select one representative slice and show a strip of key frames, the corresponding task and subgoal annotations, the program slice, and the skill function. In the listings, calls in \textbf{bold} are library skills; all others are primitive calls.

\needspace{20\baselineskip}
\noindent\textbf{Skill I: \texttt{fold\_edge\_over}}\\
\textit{Task:} ``wrap blanket roll with ribbon''\\
\textit{Sub-goal:} ``Fold the blanket at the center of the board''
\begin{center}
  \includegraphics[width=0.1995\linewidth]{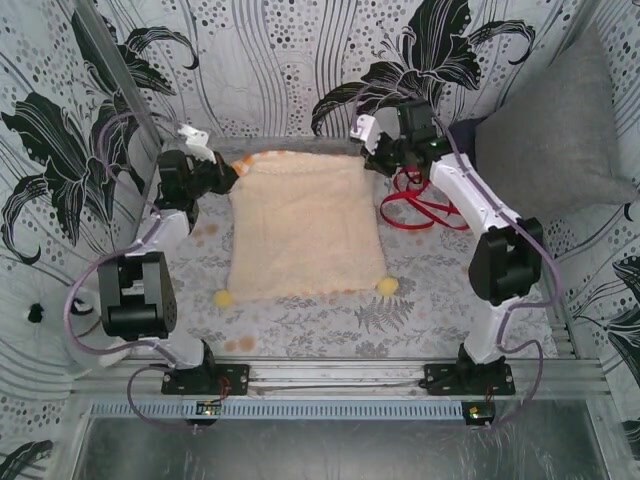}%
  \includegraphics[width=0.1995\linewidth]{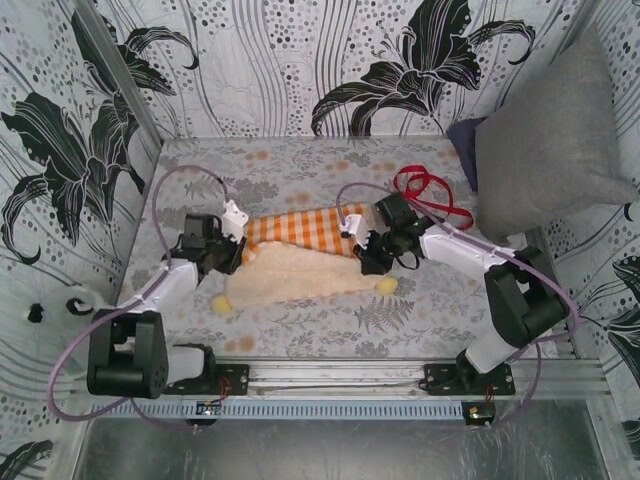}%
  \includegraphics[width=0.1995\linewidth]{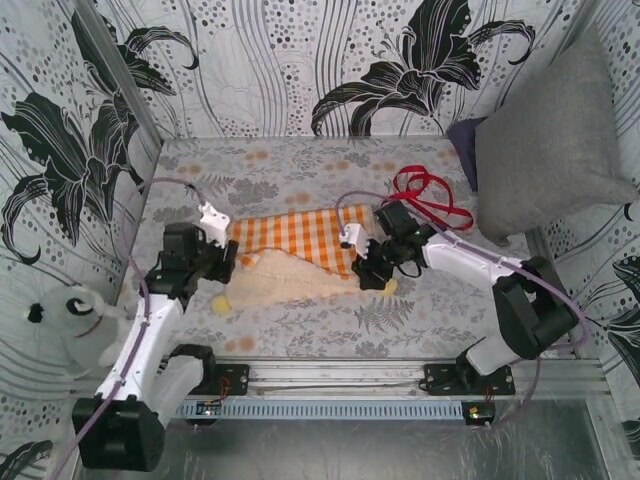}%
  \includegraphics[width=0.1995\linewidth]{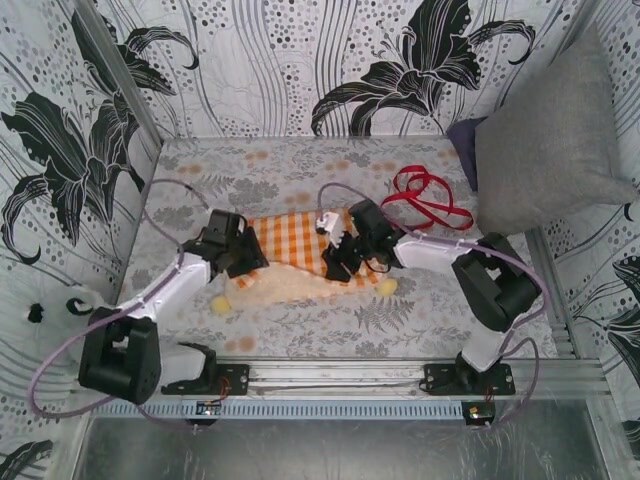}%
  \includegraphics[width=0.1995\linewidth]{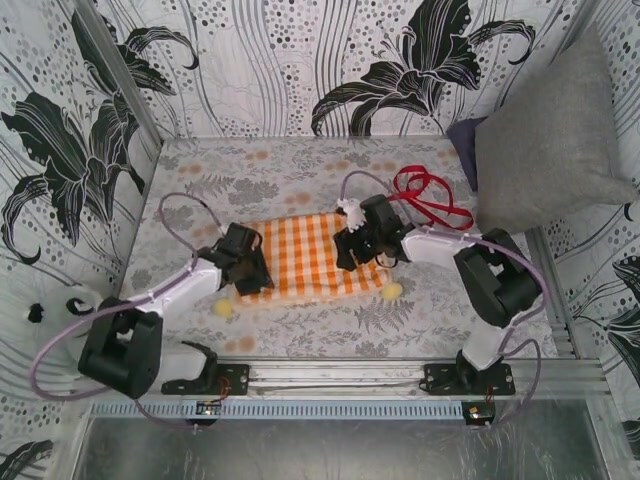}%
  \captionof{figure}{Key frames of Skill~I, \texttt{fold\_edge\_over}.}
  \label{fig:skill_fold_edge_over}
\end{center}

\noindent\textit{Skill definition:}
\begin{lstlisting}[style=skillcode]
def fold_edge_over(robot, item_name, left_delta, right_delta):
    # pinch the edge with both arms
    robot.grasp(item_name, 'left')
    robot.grasp(item_name, 'right')
    # carry the pinched edge across the item
    robot.servo_ee_delta('left', left_delta)
    robot.servo_ee_delta('right', right_delta)
    # let the fold lie
    robot.release('left')
    robot.release('right')
\end{lstlisting}

\noindent\textit{Program slice for this episode:}
\begin{lstlisting}[style=skillcode]
fold_edge_over(robot, 'blanket', [0.15, 0.0, 0.0], [0.15, 0.0, 0.0])
\end{lstlisting}

\needspace{20\baselineskip}
\noindent\textbf{Skill II: \texttt{alternating\_smoothing\_pats}}\\
\textit{Task:} ``roll towels''\\
\textit{Sub-goal:} ``Flatten the towel''
\begin{center}
  \includegraphics[width=0.1995\linewidth]{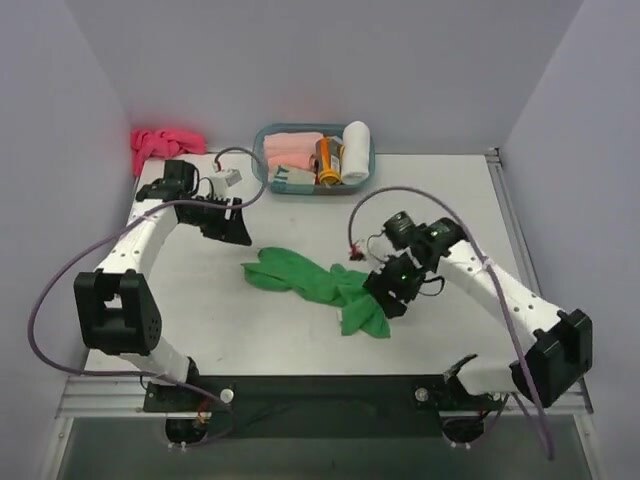}%
  \includegraphics[width=0.1995\linewidth]{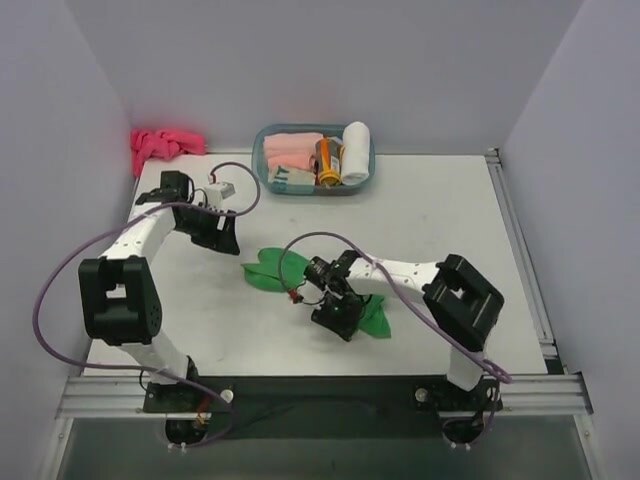}%
  \includegraphics[width=0.1995\linewidth]{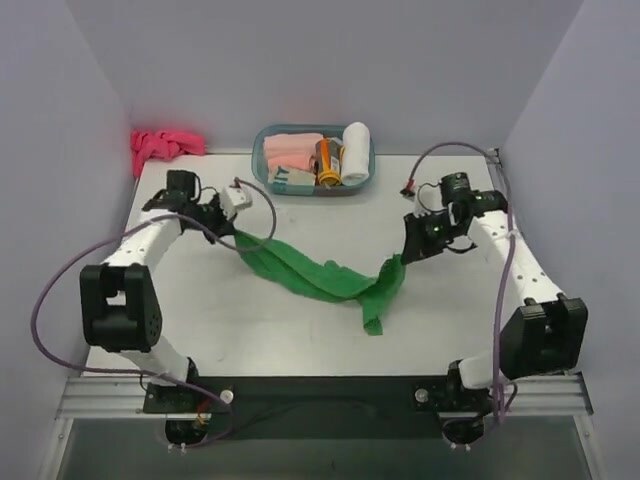}%
  \includegraphics[width=0.1995\linewidth]{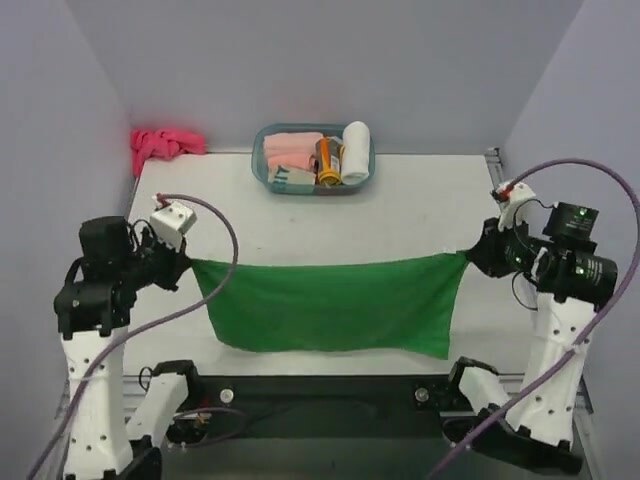}%
  \includegraphics[width=0.1995\linewidth]{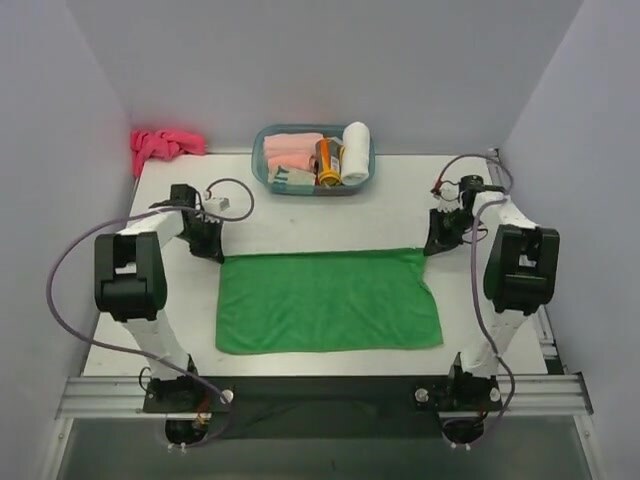}%
  \captionof{figure}{Key frames of Skill~II, \texttt{alternating\_smoothing\_pats}.}
  \label{fig:skill_alternating_smoothing_pats}
\end{center}

\noindent\textit{Skill definition:}
\begin{lstlisting}[style=skillcode]
def alternating_smoothing_pats(robot, fabric_name, first_side, second_side):
    # pat 1 -- press the fabric flat, then come off
    robot.grasp(fabric_name, first_side)
    robot.release(first_side)
    # pat 2 -- the other hand, on the other half of the sheet
    robot.grasp(fabric_name, second_side)
    robot.release(second_side)
\end{lstlisting}

\noindent\textit{Program slice for this episode:}
\begin{lstlisting}[style=skillcode]
# Smooth out the towel with alternating pats
alternating_smoothing_pats(robot, 'towel', 'left', 'right')
\end{lstlisting}

\needspace{20\baselineskip}
\noindent\textbf{Skill III: \texttt{flip\_object\_over}}\\
\textit{Task:} ``fold napkin into a case and place utensils in it''\\
\textit{Sub-goal:} ``Turn over the spoon at the center of the board''
\begin{center}
  \includegraphics[width=0.1995\linewidth]{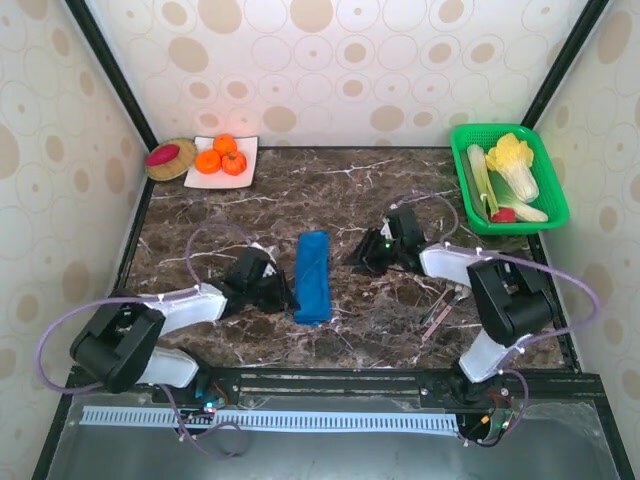}%
  \includegraphics[width=0.1995\linewidth]{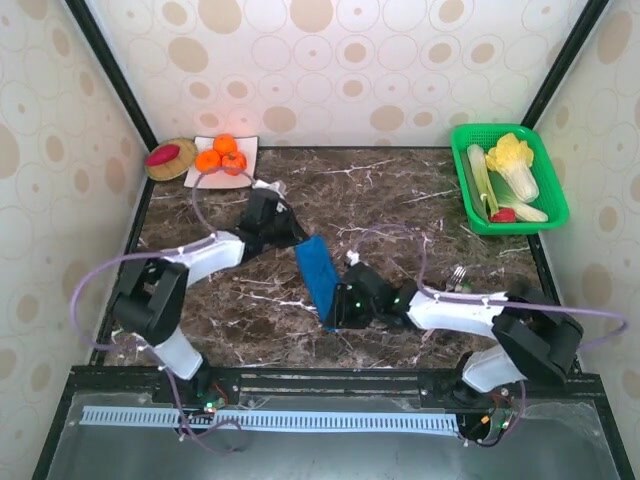}%
  \includegraphics[width=0.1995\linewidth]{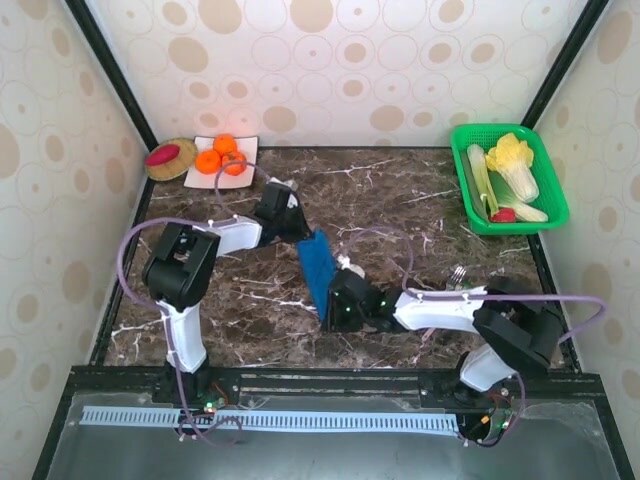}%
  \includegraphics[width=0.1995\linewidth]{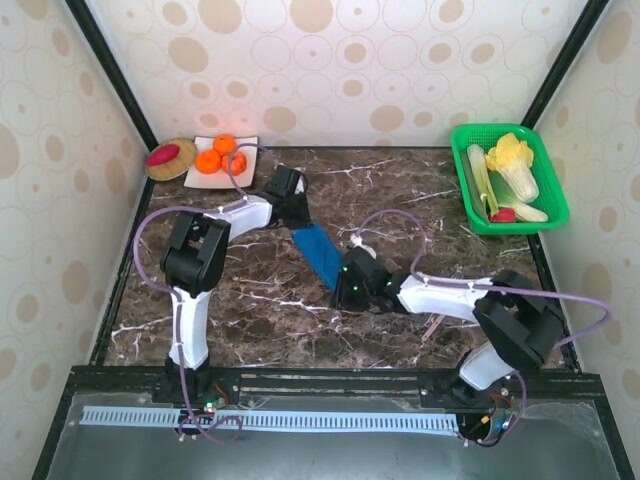}%
  \includegraphics[width=0.1995\linewidth]{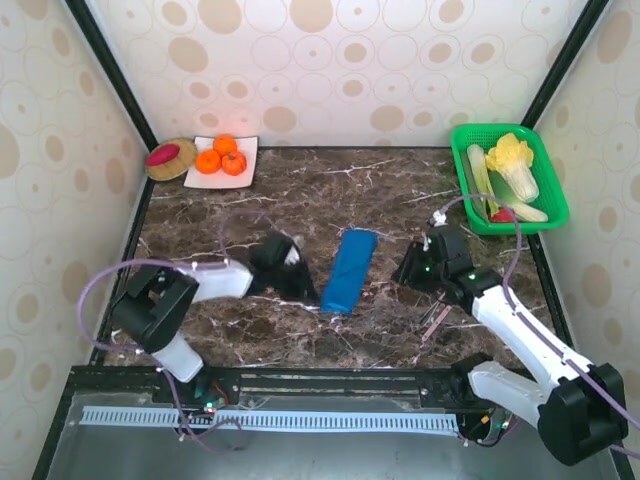}%
  \captionof{figure}{Key frames of Skill~III, \texttt{flip\_object\_over}.}
  \label{fig:skill_flip_object_over}
\end{center}

\noindent\textit{Skill definition:}
\begin{lstlisting}[style=skillcode]
def flip_object_over(robot, object_name, side, lift_delta, turn_deg=180.0):
    # lift just clear so the object can swing
    robot.servo_ee_delta(side, lift_delta)
    # roll the wrist -- this is the flip
    robot.rotate(object_name, turn_deg, side)
    # let it fall the last centimetre, landing inverted
    robot.release(side)
    # pin the flipped object down again for the next slice
    robot.grasp(object_name, side)
\end{lstlisting}

\noindent\textit{Program slice for this episode:}
\begin{lstlisting}[style=skillcode]
# Lift, roll the wrist to flip the spoon, drop it, and pin it
flip_object_over(robot, 'spoon', 'left', [0.0, 0.0, 0.1])
\end{lstlisting}

\needspace{20\baselineskip}
\noindent\textbf{Skill IV: \texttt{adjust\_in\_place}}\\
\textit{Task:} ``place bookmark into book''\\
\textit{Sub-goal:} ``Adjust the bookmark in the book''
\begin{center}
  \includegraphics[width=0.1995\linewidth]{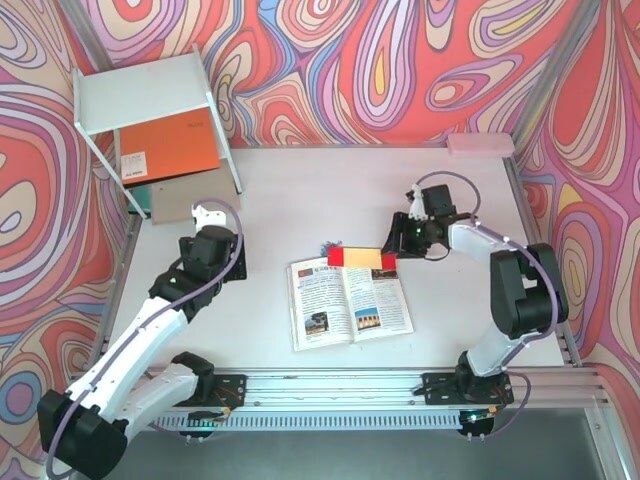}%
  \includegraphics[width=0.1995\linewidth]{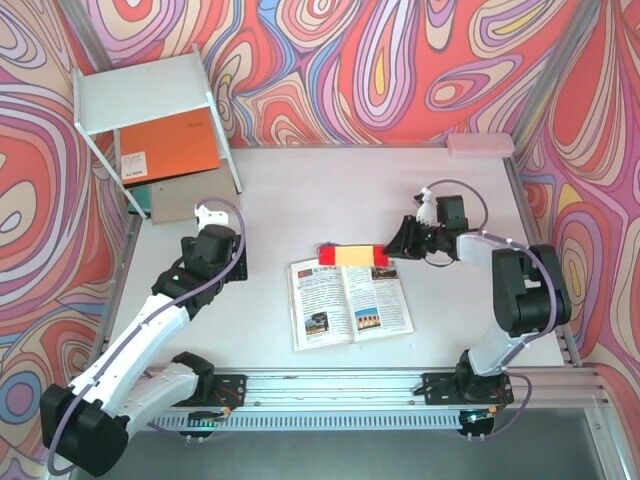}%
  \includegraphics[width=0.1995\linewidth]{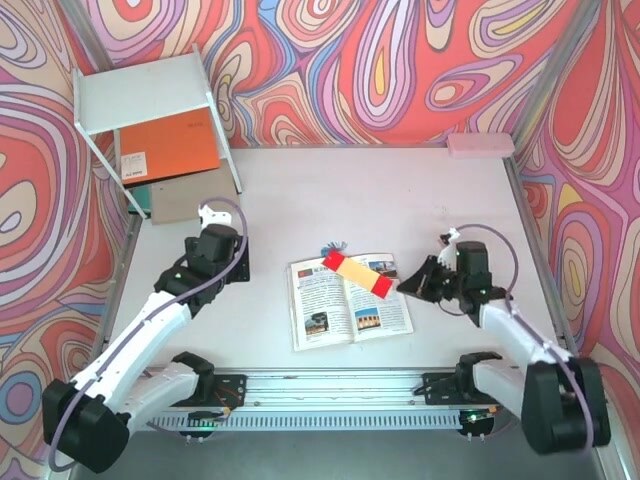}%
  \includegraphics[width=0.1995\linewidth]{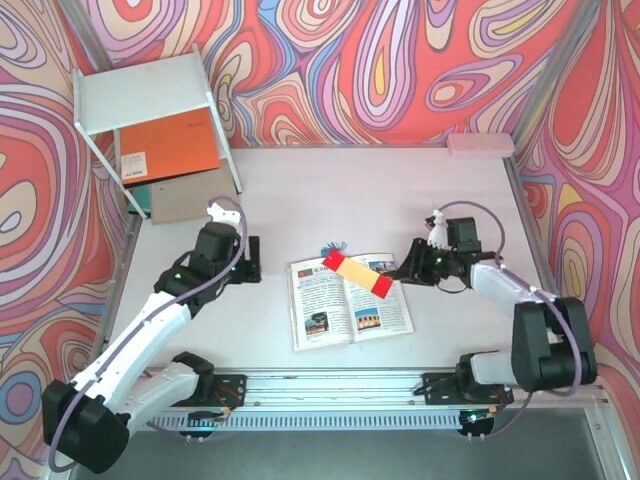}%
  \includegraphics[width=0.1995\linewidth]{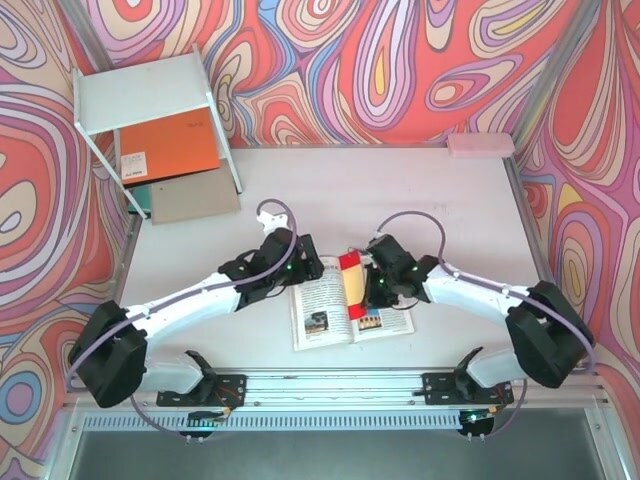}%
  \captionof{figure}{Key frames of Skill~IV, \texttt{adjust\_in\_place}.}
  \label{fig:skill_adjust_in_place}
\end{center}

\noindent\textit{Skill definition:}
\begin{lstlisting}[style=skillcode]
def adjust_in_place(robot, object_name, side, nudge_delta):
    # take hold where it sits -- no lift
    robot.grasp(object_name, side)
    # slide it by a small bounded amount
    robot.servo_ee_delta(side, nudge_delta)
    # let it settle
    robot.release(side)
\end{lstlisting}

\noindent\textit{Program slice for this episode:}
\begin{lstlisting}[style=skillcode]
adjust_in_place(robot, 'bookmark', 'right', [0.05, 0.0, 0.0])
\end{lstlisting}

\subsection{Visualizations of Egocentric Human Skills}
\label{apdx:skill_viz_human}

We visualize six of the 27 egocentric human skills. For each, we select one representative slice and show a strip of key frames, the corresponding task and sub-goal annotations, the program slice, and the skill function. In the listings, calls in \textbf{bold} are library skills; all others are primitive calls.

\needspace{20\baselineskip}
\noindent\textbf{Skill I: \texttt{open\_closure}}\\
\textit{Task:} ``assemble photo album''\\
\textit{Sub-goal:} ``Open the photo album''
\begin{center}
  \includegraphics[width=0.1995\linewidth]{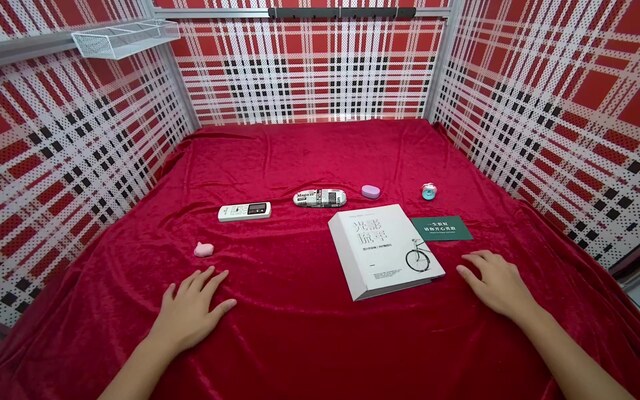}%
  \includegraphics[width=0.1995\linewidth]{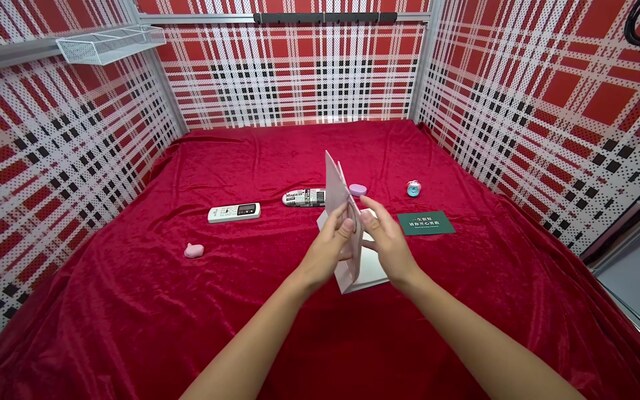}%
  \includegraphics[width=0.1995\linewidth]{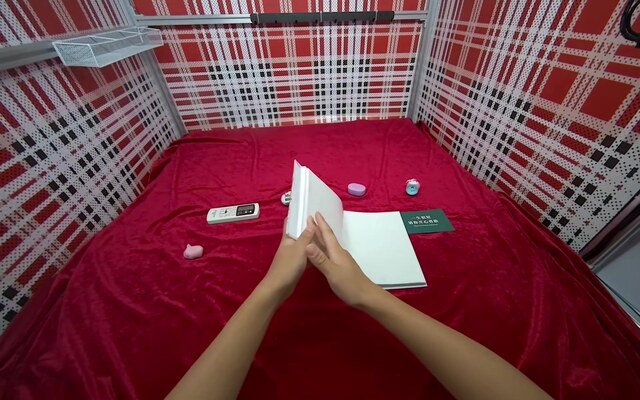}%
  \includegraphics[width=0.1995\linewidth]{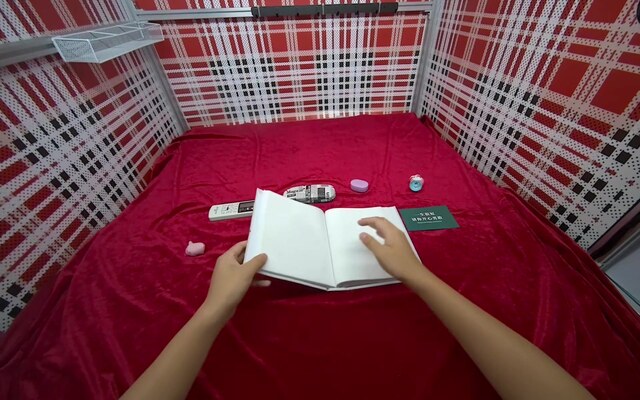}%
  \includegraphics[width=0.1995\linewidth]{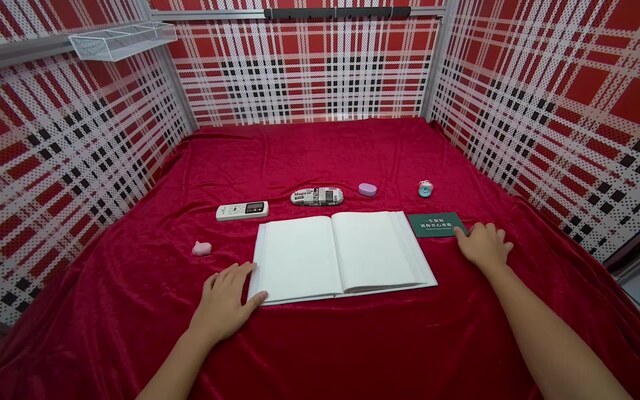}%
  \captionof{figure}{Key frames of Skill~I, \texttt{open\_closure}.}
  \label{fig:skill_open_closure}
\end{center}

\noindent\textit{Skill definition:}
\begin{lstlisting}[style=skillcode]
def open_closure(robot, container_name, closure_name, side, brace_side, axis, angle_deg):
    # Hold the container body down so the closure, not the container, moves.
    robot.grasp(container_name, brace_side)
    # Take the closure where it sits.
    robot.grasp(closure_name, side)
    # Swing it open about its hinge.
    robot.rotate(closure_name, axis, angle_deg, side)
    # Come off the closure; it stays open.
    robot.release(side)
\end{lstlisting}

\noindent\textit{Program slice for this episode:}
\begin{lstlisting}[style=skillcode]
open_closure(robot, 'photo album', 'photo album cover', 'left', 'right', 'object_long', 180.0)
\end{lstlisting}

\needspace{20\baselineskip}
\noindent\textbf{Skill II: \texttt{pour\_into\_target}}\\
\textit{Task:} ``serve rice into a bowl''\\
\textit{Sub-goal:} ``Pour a spoonful of rice into the bowl''
\begin{center}
  \includegraphics[width=0.1995\linewidth]{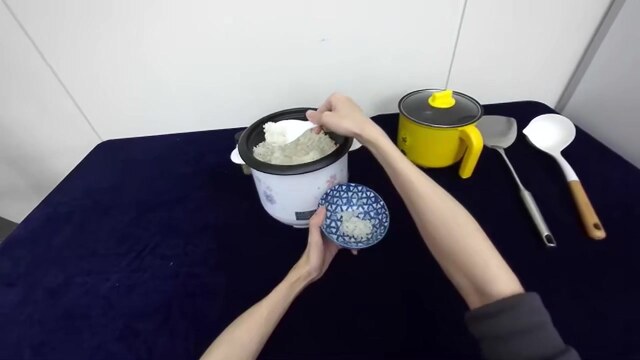}%
  \includegraphics[width=0.1995\linewidth]{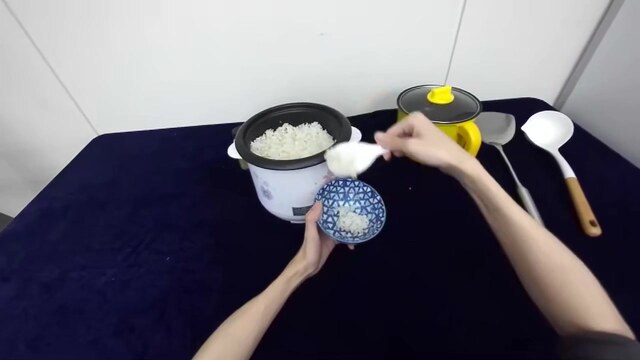}%
  \includegraphics[width=0.1995\linewidth]{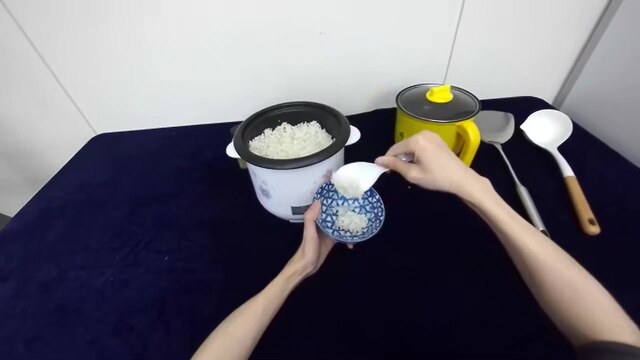}%
  \includegraphics[width=0.1995\linewidth]{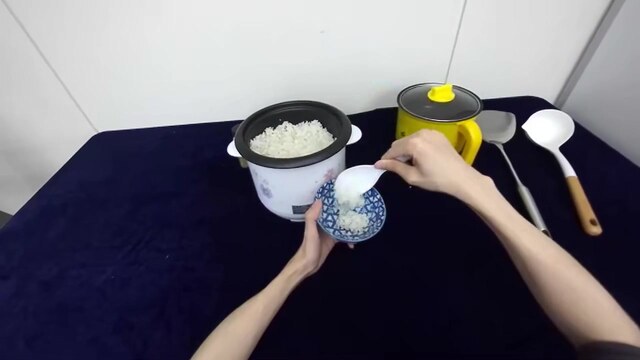}%
  \includegraphics[width=0.1995\linewidth]{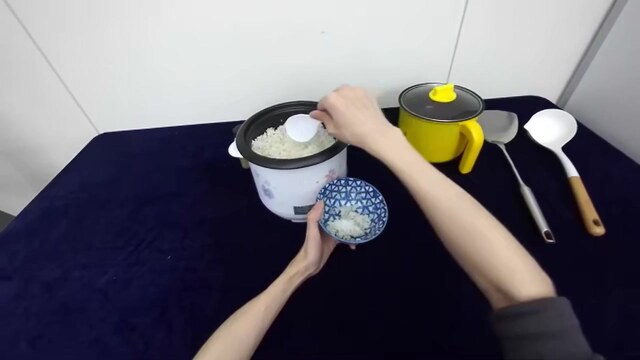}%
  \captionof{figure}{Key frames of Skill~II, \texttt{pour\_into\_target}.}
  \label{fig:skill_pour_into_target}
\end{center}

\noindent\textit{Skill definition:}
\begin{lstlisting}[style=skillcode]
def pour_into_target(robot, vessel_name, target_name, side, axis, angle_deg):
    # Bring the held vessel's mouth over the receiving target.
    robot.goto_relative(side, target_name, 'above')
    # Tip it until it pours.
    robot.rotate(vessel_name, axis, angle_deg, side)
    # Bring it back upright; the pour ends.
    robot.rotate(vessel_name, axis, -angle_deg, side)
\end{lstlisting}

\noindent\textit{Program slice for this episode:}
\begin{lstlisting}[style=skillcode]
pour_into_target(robot, 'spoon', 'bowl', 'right', 'object_long', 110.0)
\end{lstlisting}

\needspace{20\baselineskip}
\noindent\textbf{Skill III: \texttt{stir\_contents}}\\
\textit{Task:} ``mixing solutions with stirring rod''\\
\textit{Sub-goal:} ``Stir the solution in the beaker''
\begin{center}
  \includegraphics[width=0.1995\linewidth]{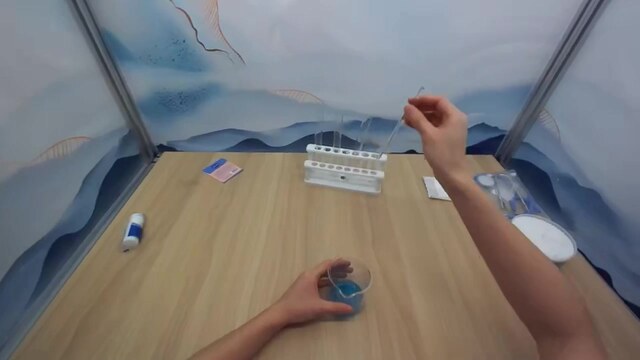}%
  \includegraphics[width=0.1995\linewidth]{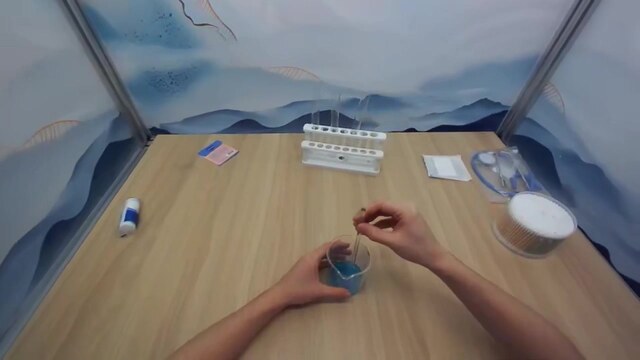}%
  \includegraphics[width=0.1995\linewidth]{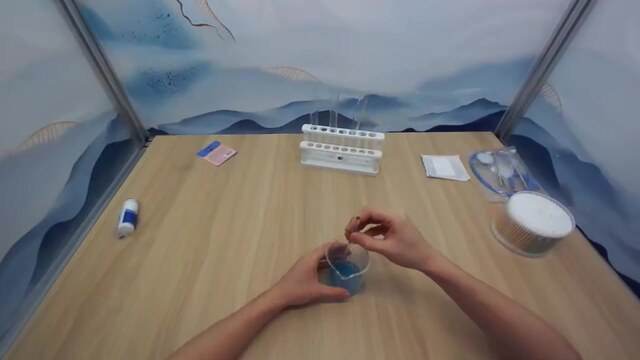}%
  \includegraphics[width=0.1995\linewidth]{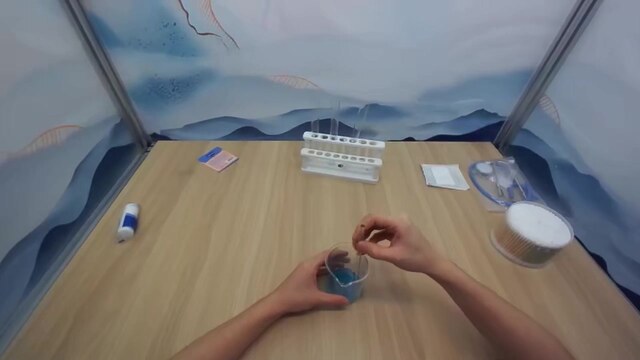}%
  \includegraphics[width=0.1995\linewidth]{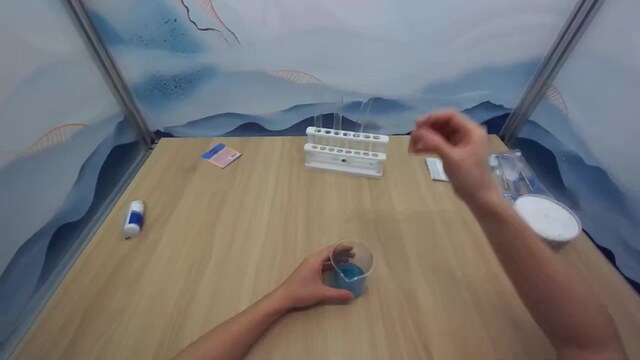}%
  \captionof{figure}{Key frames of Skill~III, \texttt{stir\_contents}.}
  \label{fig:skill_stir_contents}
\end{center}

\noindent\textit{Skill definition:}
\begin{lstlisting}[style=skillcode]
def stir_contents(robot, tool_name, vessel_name, side):
    # Bring the held tool down inside the vessel, among the contents.
    robot.goto_relative(side, vessel_name, 'inside')
    # Sweep it round the vessel; the grip never changes, nothing is let go.
    robot.rotate(tool_name, 'vertical', 360.0, side)
    robot.rotate(tool_name, 'vertical', 360.0, side)
\end{lstlisting}

\noindent\textit{Program slice for this episode:}
\begin{lstlisting}[style=skillcode]
brace_and_take_tool(robot, 'beaker', 'left', 'stirring rod', 'right')
stir_contents(robot, 'stirring rod', 'beaker', 'right')
\end{lstlisting}

\needspace{20\baselineskip}
\noindent\textbf{Skill IV: \texttt{wipe\_surface\_back\_and\_forth}}\\
\textit{Task:} ``use lint roller on pants''\\
\textit{Sub-goal:} ``Remove the lint from the pair of pants''
\begin{center}
  \includegraphics[width=0.1995\linewidth]{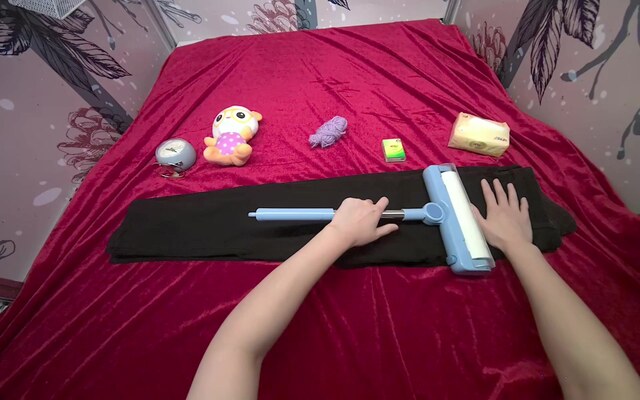}%
  \includegraphics[width=0.1995\linewidth]{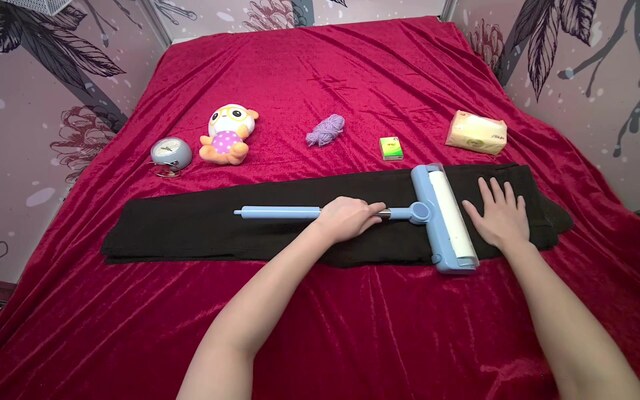}%
  \includegraphics[width=0.1995\linewidth]{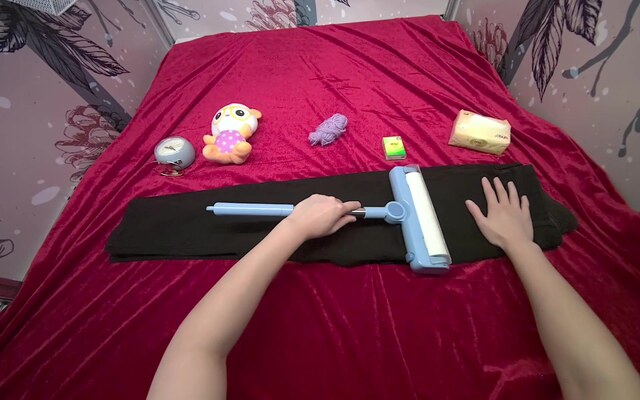}%
  \includegraphics[width=0.1995\linewidth]{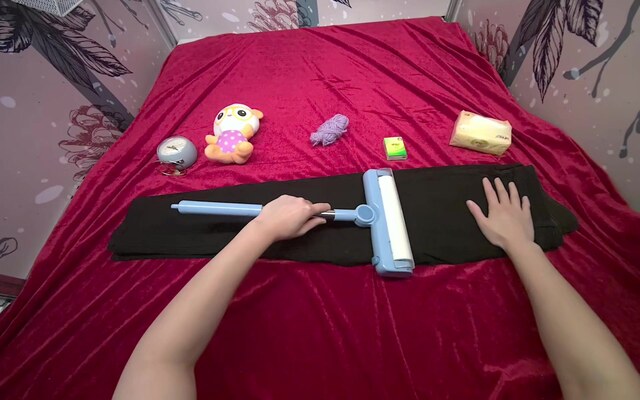}%
  \includegraphics[width=0.1995\linewidth]{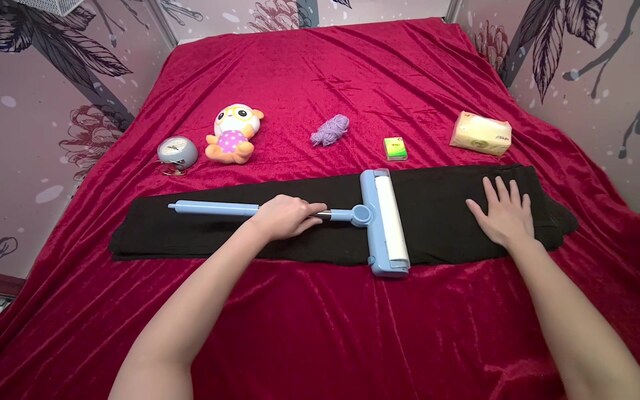}%
  \captionof{figure}{Key frames of Skill~IV, \texttt{wipe\_surface\_back\_and\_forth}.}
  \label{fig:skill_wipe_surface_back_and_forth}
\end{center}

\noindent\textit{Skill definition:}
\begin{lstlisting}[style=skillcode]
def wipe_surface_back_and_forth(robot, surface_name, relation, stroke_direction, return_direction, side):
    # Park the cloth on the surface.
    robot.goto_relative(side, surface_name, relation)
    # Out and back.
    robot.move_arm(side, stroke_direction)
    robot.move_arm(side, return_direction)
    # And again.
    robot.move_arm(side, stroke_direction)
    robot.move_arm(side, return_direction)
\end{lstlisting}

\noindent\textit{Program slice for this episode:}
\begin{lstlisting}[style=skillcode]
# Roll the lint roller back and forth over the pants (4 pairs of strokes)
for _ in range(2):
    wipe_surface_back_and_forth(robot, 'pants', 'against', 'right', 'left', 'left')
\end{lstlisting}

\needspace{20\baselineskip}
\noindent\textbf{Skill V: \texttt{fold\_over\_and\_crease}}\\
\textit{Task:} ``fold long pants''\\
\textit{Sub-goal:} ``Fold a pair of pants''
\begin{center}
  \includegraphics[width=0.1995\linewidth]{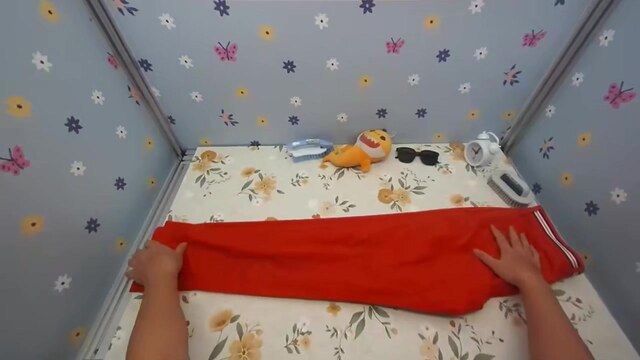}%
  \includegraphics[width=0.1995\linewidth]{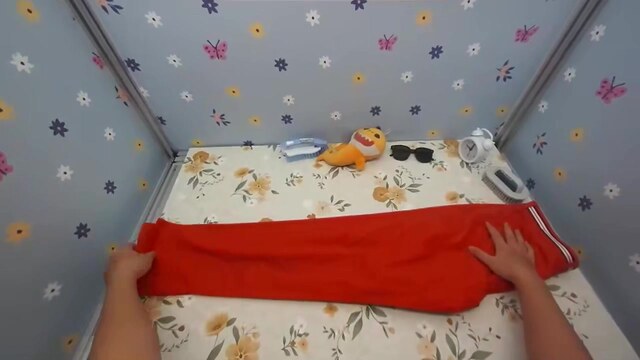}%
  \includegraphics[width=0.1995\linewidth]{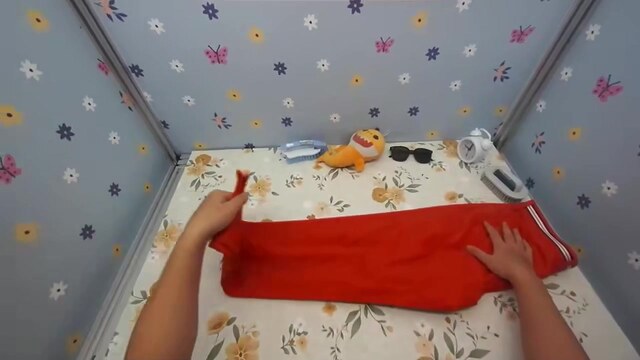}%
  \includegraphics[width=0.1995\linewidth]{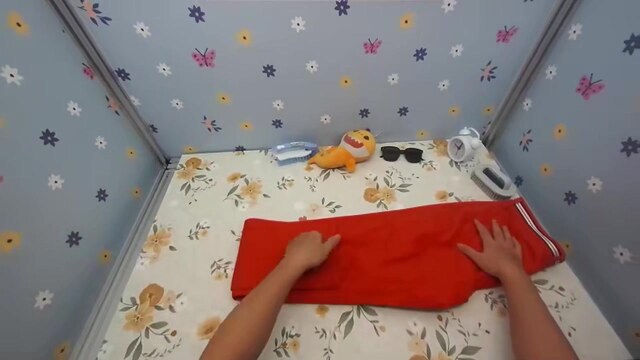}%
  \includegraphics[width=0.1995\linewidth]{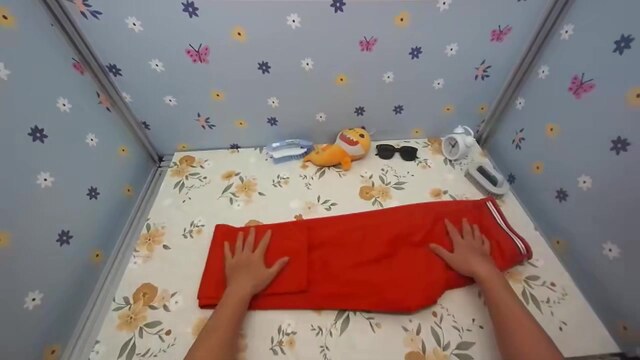}%
  \captionof{figure}{Key frames of Skill~V, \texttt{fold\_over\_and\_crease}.}
  \label{fig:skill_fold_over_and_crease}
\end{center}

\noindent\textit{Skill definition:}
\begin{lstlisting}[style=skillcode]
def fold_over_and_crease(robot, item_name, side, brace_side, direction, fold_direction):
    # Hold the item down so the fold, not the whole item, travels.
    robot.grasp(item_name, brace_side)
    # Take the free edge with the working hand.
    robot.grasp(item_name, side)
    # Carry the edge across the item ...
    robot.move_arm(side, direction)
    # ... and land it on the far side.
    robot.move_arm(side, fold_direction)
    # Let the folded edge down.
    robot.release(side)
    # Press the crease flat.
    robot.press(item_name, side)
\end{lstlisting}

\noindent\textit{Program slice for this episode:}
\begin{lstlisting}[style=skillcode]
fold_over_and_crease(robot, 'pants', 'left', 'right', 'up', 'right')
\end{lstlisting}

\needspace{20\baselineskip}
\noindent\textbf{Skill VI: \texttt{wrap\_around\_target}}\\
\textit{Task:} ``use a curling iron to curl hair''\\
\textit{Sub-goal:} ``Wrap a section of hair around the curling iron''
\begin{center}
  \includegraphics[width=0.1995\linewidth]{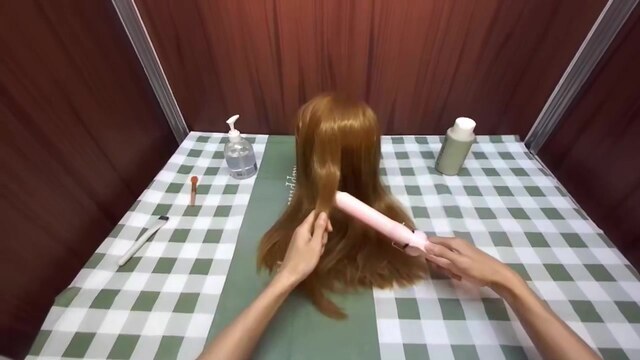}%
  \includegraphics[width=0.1995\linewidth]{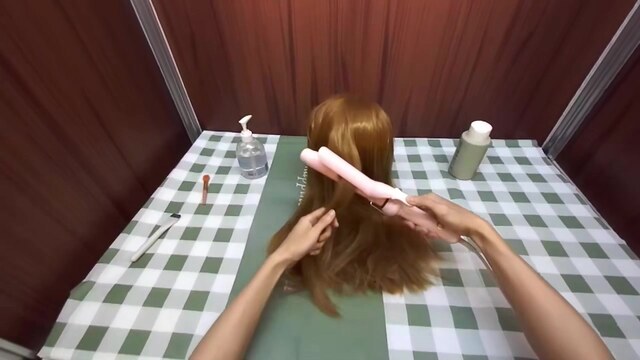}%
  \includegraphics[width=0.1995\linewidth]{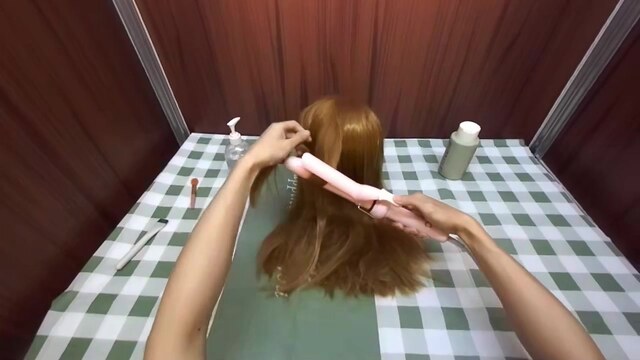}%
  \includegraphics[width=0.1995\linewidth]{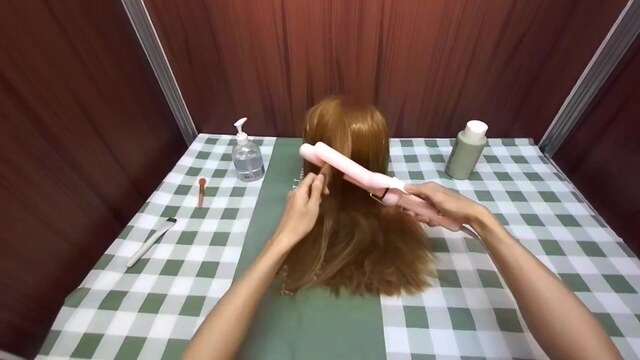}%
  \includegraphics[width=0.1995\linewidth]{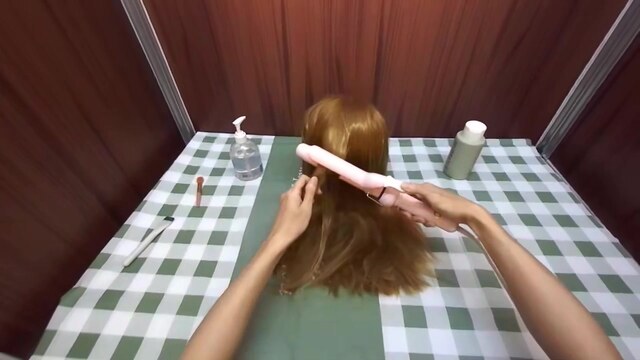}%
  \captionof{figure}{Key frames of Skill~VI, \texttt{wrap\_around\_target}.}
  \label{fig:skill_wrap_around_target}
\end{center}

\noindent\textit{Skill definition:}
\begin{lstlisting}[style=skillcode]
def wrap_around_target(robot, target_name, side):
    # Start over the top of the target.
    robot.goto_relative(side, target_name, 'above')
    # Over the far side.
    robot.move_arm(side, 'away_from_body')
    # Down its far face.
    robot.move_arm(side, 'down')
    # Back underneath, toward the body.
    robot.move_arm(side, 'toward_body')
    # And up to close the circuit.
    robot.move_arm(side, 'up')
\end{lstlisting}

\noindent\textit{Program slice for this episode:}
\begin{lstlisting}[style=skillcode]
# Wrap the hair around the curling iron twice.
for _ in range(2):
    wrap_around_target(robot, 'curling iron', 'left')
\end{lstlisting}

\subsection{Skill Distribution in the Embedding Space}

In Figure~\ref{fig:skill_embed_yam} and~\ref{fig:skill_embed_human}, we visualize the skill distribution of variant skills in the pretrained code expert's embedding space. Each point represents one episode slice, placed by a two-dimensional t-SNE projection of its embedding and colored by the skill assigned to it. Dashed outlines mark the region covered by each skill, and the frames along the edges show a representative short span from an episode for selected skills. Figure~\ref{fig:skill_embed_yam} covers the 18 most common skills in the YAM library and Figure~\ref{fig:skill_embed_human} the 16 most common skills in the egocentric human library. In both, slices assigned to different skills occupy largely separate regions, suggesting that the skill label is predictive of the motion in a slice and can serve as a conditioning signal for action generation.

\needspace{18\baselineskip}
\begin{center}
  \includegraphics[width=\linewidth]{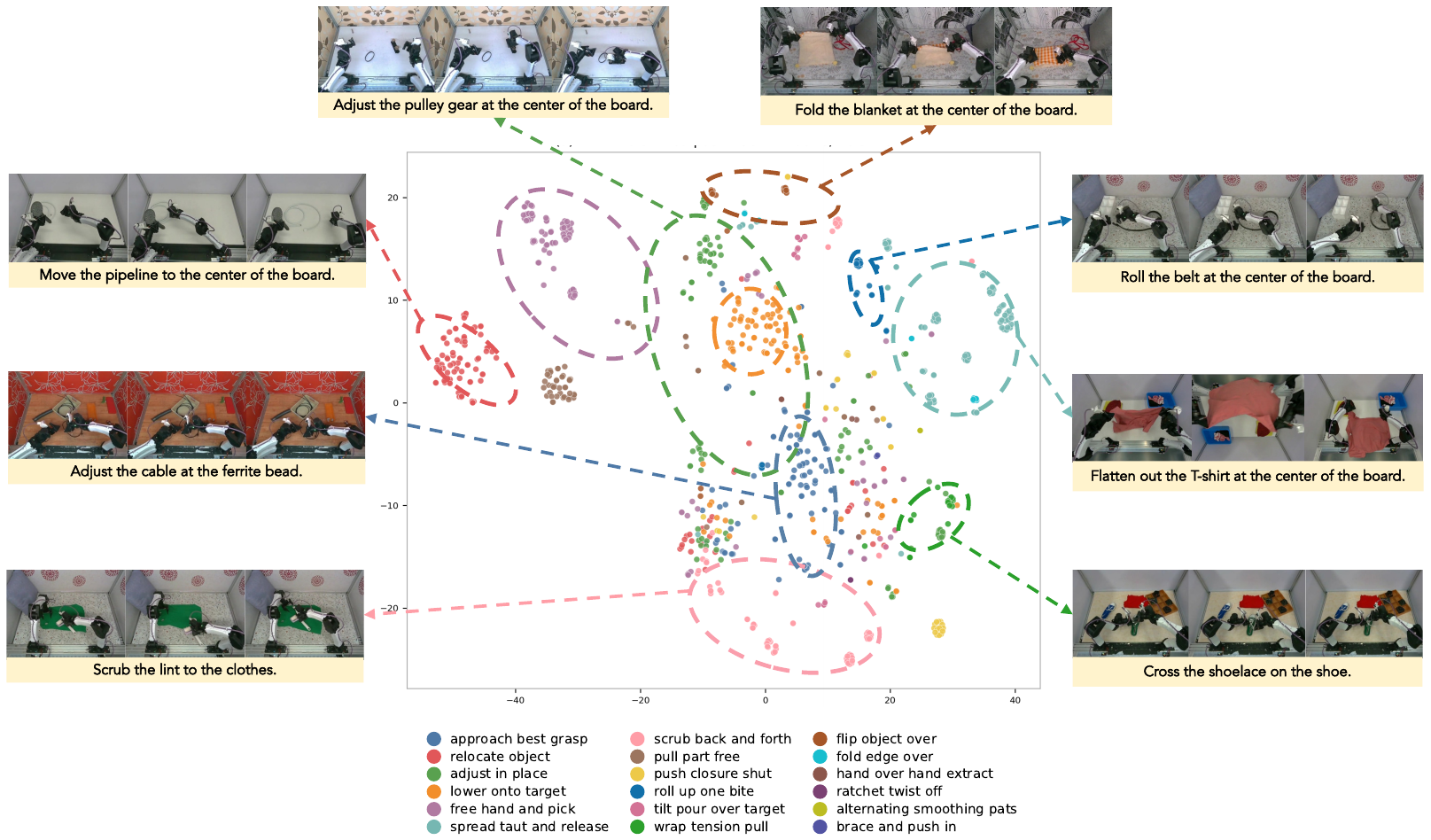}%
  \captionof{figure}{\textbf{Skills of the YAM Skill Library in the Embedding Space.}}
  \label{fig:skill_embed_yam}
\end{center}

\needspace{18\baselineskip}
\begin{center}
  \includegraphics[width=\linewidth]{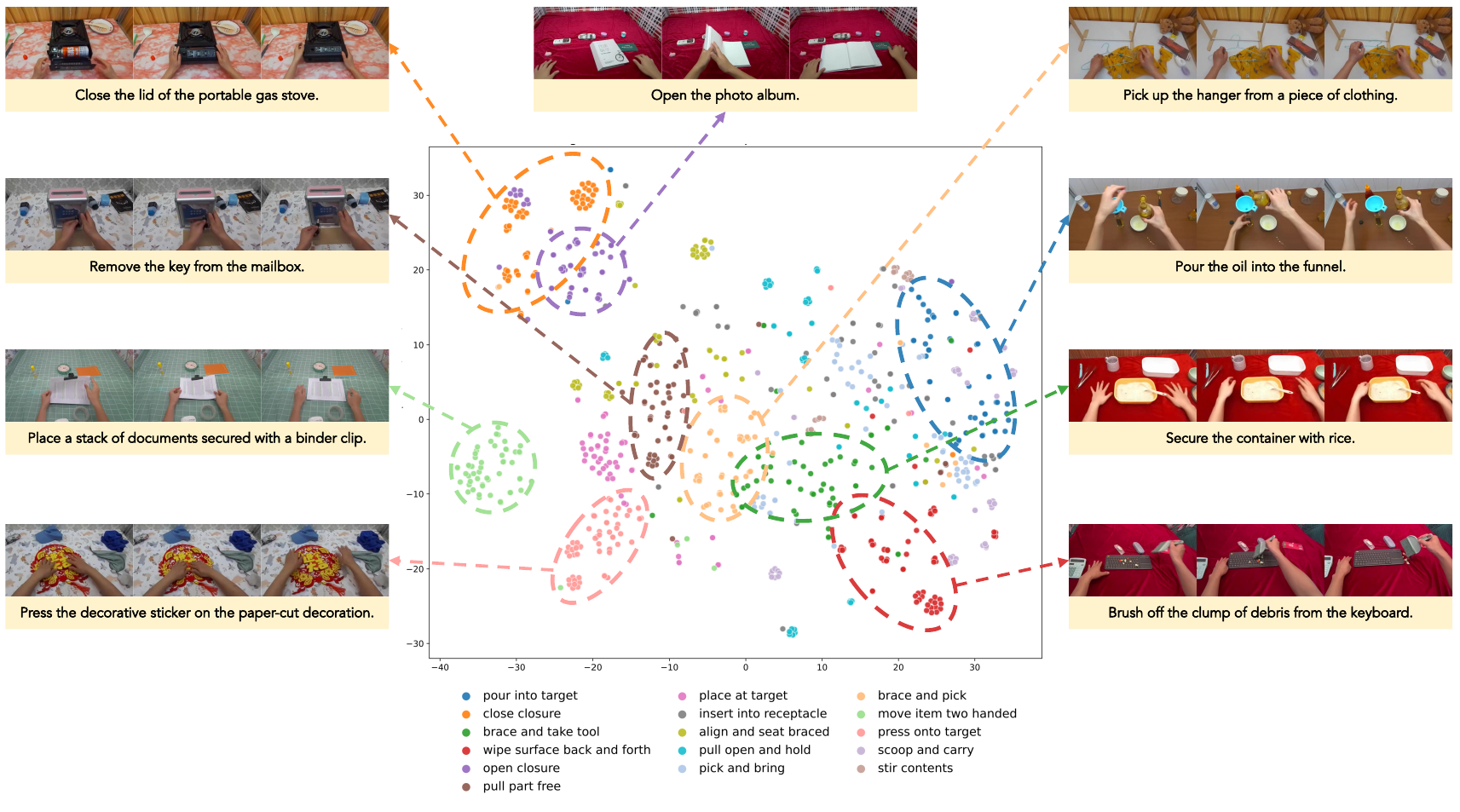}%
  \captionof{figure}{\textbf{Skills of the Egocentric Human Skill Library in the Embedding Space.}}
  \label{fig:skill_embed_human}
\end{center}

\needspace{10\baselineskip}
\section{Baseline Implementation Details}
\label{app:baseline_details}

In Sections~\ref{sec:exp_sim} and~\ref{sec:exp_real}, we compare \Ours{} with four baselines: $\pi_{0.5}$, DreamZero, EgoScale, and WLA. Here, we describe how each is reproduced and adapted. On RoboTwin~2.0, we train a single multi-task policy per baseline on 2,500 \texttt{demo\_clean} demonstrations (50 tasks $\times$ 50 demonstrations). For each of the 8 real-world tasks, we train a separate policy per baseline using task-specific demonstrations sampled at 30\,Hz. Every baseline receives a language instruction, the robot state, and three RGB views (one head and two wrist cameras). All baselines follow the same evaluation protocol as \Ours{}. All real-world rollouts are scored using the task-specific rubrics in Appendix~\ref{apdx:task_rubrics}.

$\bm{\pi_{0.5}}$~\citep{black2025pi05}
is a vision-language-action model that combines a pretrained vision-language backbone with a flow-matching action expert. We follow the official OpenPI implementation\footnote{\url{https://github.com/Physical-Intelligence/openpi}} and initialize all policies from the released $\pi_{0.5}$ pretrained checkpoint. For each real-world task, we fine-tune a separate policy for 50 epochs on the task-specific demonstrations. We use 14-dimensional joint-space actions on RoboTwin and YAM, and 62-dimensional actions on Tianji--Sharpa consisting of a 9-dimensional end-effector pose and 22 hand joints per side. Arm actions are relative to the current state, while gripper and hand actions are absolute targets. We reinitialize the action input and output projections to accommodate these dimensions and fine-tune all model parameters. Following the official implementation, we use an action horizon of 50 for RoboTwin and 32 for the real robots, together with quantile normalization and standard image augmentation. All models are trained with AdamW on eight H100 GPUs using FSDP and a global batch size of 256. We use a peak learning rate of $5\times10^{-5}$ with cosine decay to $5\times10^{-6}$ and 2,000 warmup steps.

\textbf{DreamZero}~\citep{ye2026dreamzero}
is a world-action model that jointly generates future video and actions using a pretrained video diffusion backbone. We follow the official implementation\footnote{\url{https://github.com/dreamzero0/dreamzero}} and initialize from the DreamZero-AgiBot checkpoint. We reproduce DreamZero on RoboTwin and all 8 real-world tasks, training a separate policy for 10 epochs on each real-world task. We fully fine-tune the diffusion transformer and state/action modules, while keeping the text encoder, image encoder, and video VAE frozen. Following the original design, the three camera views are tiled into one frame, and each training sample contains up to four action chunks of 24 steps. Actions are predicted relative to the state at the beginning of each chunk, with absolute hand-joint targets on Tianji--Sharpa. The state and action projections are adapted to the target embodiment, and the video and action losses are weighted equally. All policies are trained with AdamW on 32 H100 GPUs using a global batch size of 32. We use the same training hyperparameters across all tasks, with a peak learning rate of $2\times10^{-5}$ and 5\% linear warmup followed by cosine decay.

\textbf{EgoScale} (using the GR00T N1.7-3B checkpoint)~\citep{zheng2026egoscale}
studies how pretraining on egocentric human demonstrations scales to dexterous robot manipulation. We reproduce this baseline using the GR00T N1.7 implementation\footnote{\url{https://github.com/NVIDIA/Isaac-GR00T}} and initialize from the pretrained \texttt{nvidia/GR00T-N1.7-3B} checkpoint. For each real-world task, we fine-tune a separate policy for 50 epochs while keeping the vision-language backbone frozen and training the projection layers and diffusion action head. We use relative joint actions for the arms on RoboTwin and YAM, relative end-effector actions on Tianji--Sharpa, and absolute targets for the grippers and 22-DoF dexterous hands. Following the official fine-tuning recipe, we apply state dropout with probability 0.2, random image cropping, and color jitter. We use an action horizon of 20 steps and train with AdamW on eight H100 GPUs with a global batch size of 640. The peak learning rate is $10^{-4}$, with 5\% linear warmup followed by cosine decay and a weight decay of $10^{-5}$.

\textbf{WLA}~\citep{yang2026worldlanguageactionmodelunifiedworld}
combines a vision-language backbone, a world expert, and an action expert to jointly model visual observations, language, and robot actions. We follow the official implementation\footnote{\url{https://github.com/SJTU-DENG-Lab/WLA}} and initialize the backbone and world expert from pretrained RynnBrain-2B and Sana-600M weights, respectively. The connector and action expert are randomly initialized. We train a separate policy for 50 epochs on each real-world task, updating all modules except the image VAE. Following the original design, the policy observes the current head and wrist views together with the head view from 8 steps earlier. We use 64 meta-queries and a 32-step action chunk, with absolute action targets normalized to $[-1,1]$. The world expert predicts the head-camera image at the end of the chunk, and its flow-matching loss is weighted by 0.1 relative to the action loss. All models are trained with AdamW on eight H100 GPUs using a global batch size of 256. We use a peak learning rate of $5\times10^{-5}$ with cosine decay to $5\times10^{-6}$ and 1,000 warmup steps.

\section{Real-world Experimental Setup}
\label{app:real_world_setup}

\begin{figure}[htbp]
    \centering
    \begin{subfigure}[b]{0.437\linewidth}
        \centering
        \includegraphics[width=\linewidth]{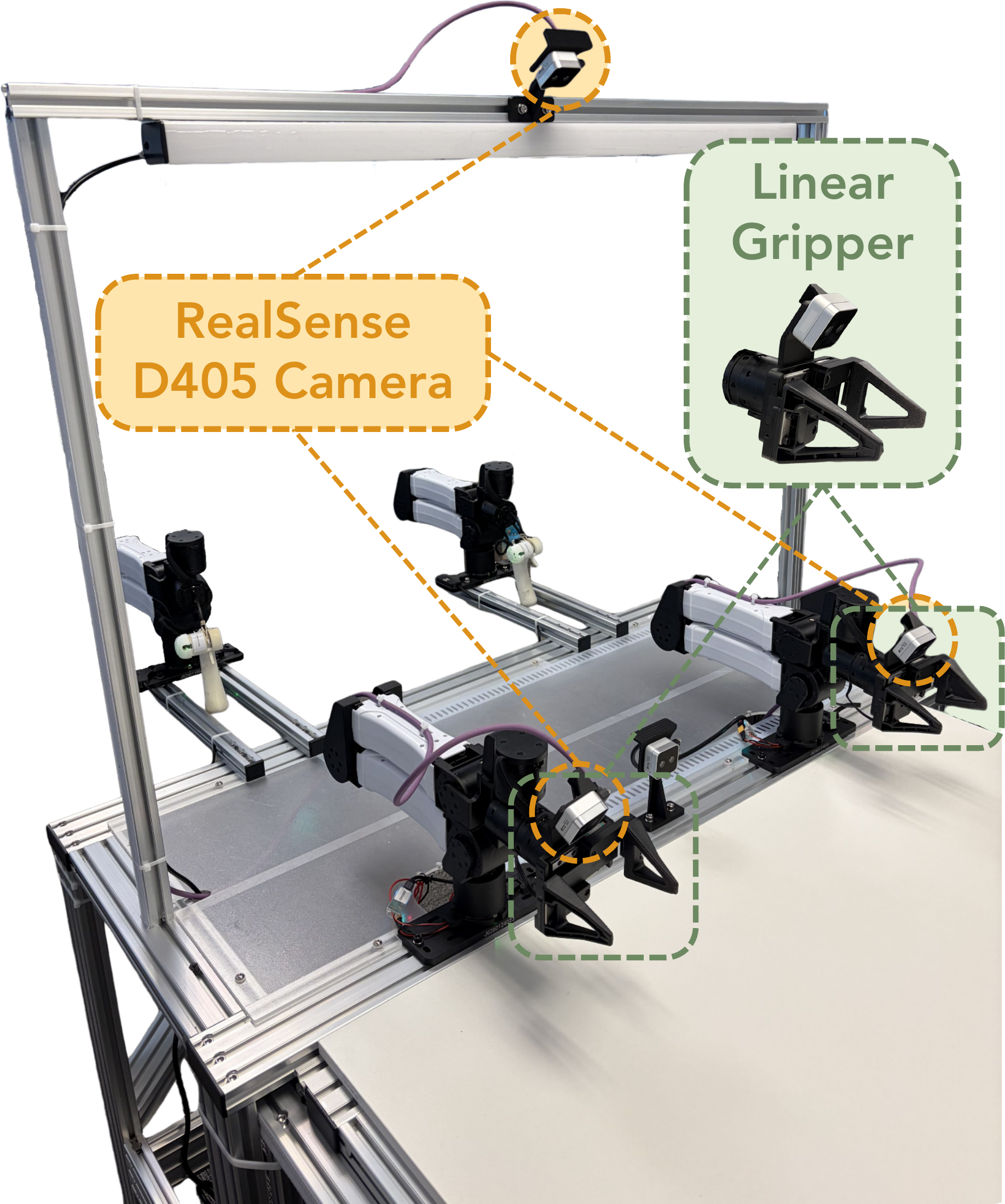}
        \caption{YAM Station}
        \label{fig:hardware_setup_yam}
    \end{subfigure}\hfill
    \begin{subfigure}[b]{0.533\linewidth}
        \centering
        \includegraphics[width=\linewidth]{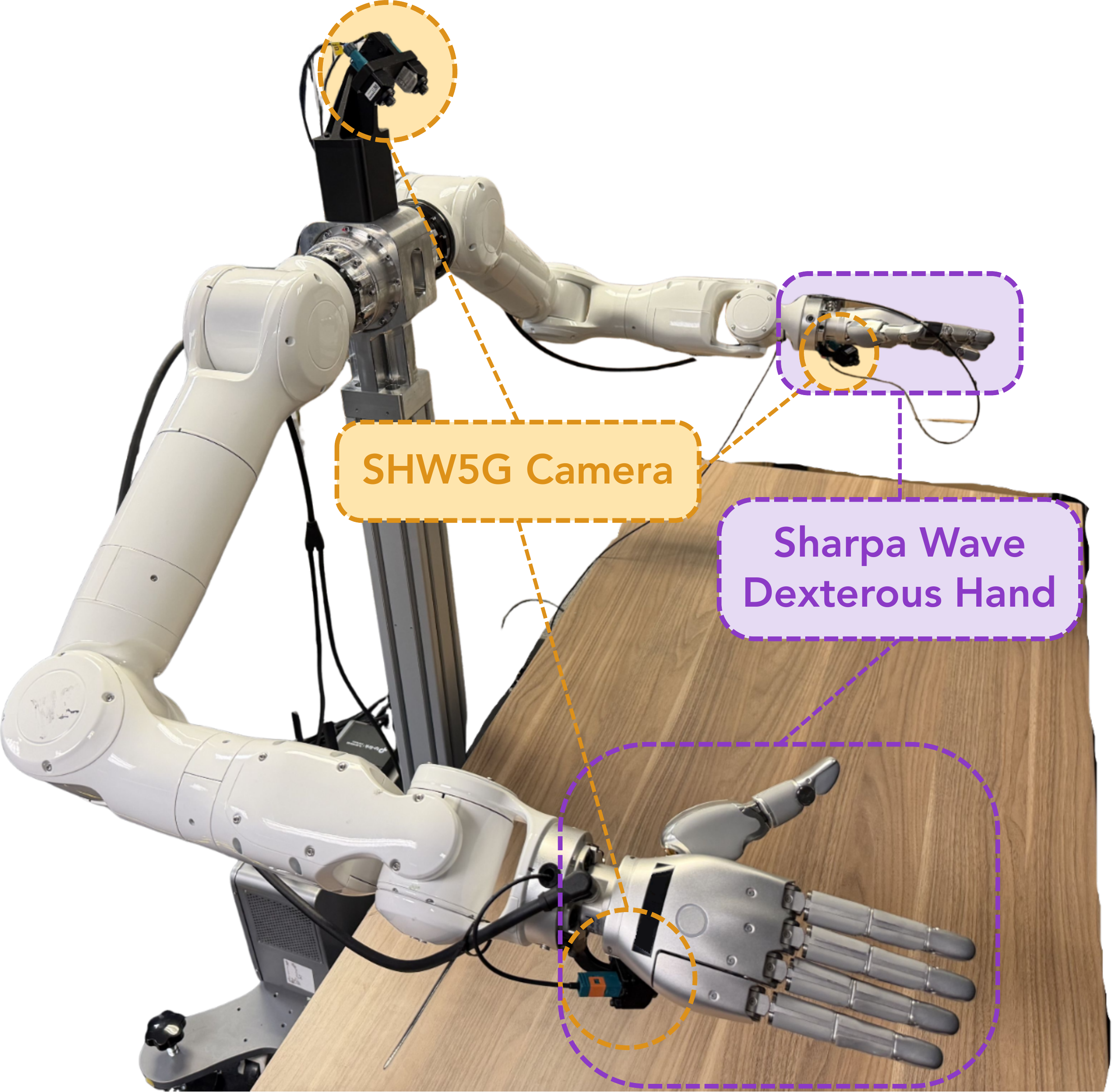}
        \caption{Tianji--Sharpa Station}
        \label{fig:hardware_setup_tianji_sharpa}
    \end{subfigure}
    \caption{\textbf{Real-world Robot Setups.} (a) The YAM station: two arms with linear grippers, with a RealSense D405 camera overhead and one on each wrist. The bottom camera is not used. (b) The Tianji--Sharpa station: a Tianji torso with two Sharpa Wave dexterous hands, observed by an SHW5G camera on the head and one on each wrist.}
    \label{fig:real_world_setup}
\end{figure}

We conduct data collection and policy evaluation on two robot embodiments: the YAM station with parallel grippers and the Tianji--Sharpa station with dexterous hands. This section describes the robot hardware, visual observations, and teleoperation interfaces used in our experiments. Figure~\ref{fig:real_world_setup} gives an overview of the two stations.

\textbf{Robot Hardware and Control.}
The YAM station is a fixed-base bimanual system with two 6-DoF arms, each equipped with a parallel gripper. The Tianji--Sharpa station consists of a Tianji humanoid torso with two arms and two 22-DoF Sharpa dexterous hands. For \Ours{}, both embodiments use the 64-dimensional state--action representation described in Appendix~\ref{app:model_training_details}. The policy predicts relative end-effector poses together with absolute gripper targets on YAM or absolute finger-joint targets on Tianji--Sharpa. 

\textbf{Perception System.}
Both stations provide three RGB views: a head or overhead camera observes the tabletop workspace, and a camera at each wrist provides a local view of the gripper or hand and the manipulated objects. Each policy receives these images together with the task instruction and robot state. For \Ours{}, images are resized to $224\times320$ pixels, and the state contains the end-effector poses and the gripper positions or hand-joint angles. The same camera streams are provided to all baselines, with method-specific preprocessing described in Appendix~\ref{app:baseline_details}.

\textbf{Teleoperation and Data Collection.}
We collect YAM demonstrations using the standard YAM station configuration, with two master arms and two puppet arms. The operator moves the master arms, whose measured joint angles are sent as joint-position targets to the corresponding puppet arms. The triggers on the teaching handles control the parallel gripper openings, yielding six arm-joint targets and one gripper target per side. For Tianji--Sharpa, we use Manus gloves with PICO motion trackers attached to the gloves, together with an Apple Vision Pro worn around the operator's neck. The tracked wrist positions and orientations define poses in $\mathrm{SE}(3)$, which are mapped to end-effector targets for the two Tianji arms. In parallel, finger-motion measurements from the Manus gloves are retargeted to 22-dimensional joint-position targets for each Sharpa hand. At each teleoperation step, the arm-pose and hand-joint targets are passed to the corresponding robot controllers to coordinate arm and finger motion. During collection, we record the three RGB camera streams, measured robot states, and commanded action targets. We use their timestamps to pair visual observations with robot states and action commands, and associate each demonstration with a task instruction for post-training.

\section{Real-world Task Progress}
\label{apdx:task_rubrics}

We evaluate 8 real-world tasks that span two robot platforms. \textit{Arrange Fruits}, \textit{Pack Backpack}, \textit{Replace Lightbulb}, and \textit{Fold and Stack Towels} run on the YAM bimanual setup (two 6-DoF arms with parallel grippers);  \textit{Arrange Fruits with Tongs}, \textit{Deal Poker}, \textit{Flip Scoreboard}, and \textit{Solve Rubik's Cube} run on the Tianji humanoid torso with two Sharpa dexterous hands. For each task we show the progress with the key frames of one training episode together with its code label; every color-coded block of the program is linked to the range of frames it explains. Each task is graded with an additive rubric that awards independent partial credit for each completed sub-step.

\subsection{YAM Tasks}

\needspace{14\baselineskip}
\noindent\textbf{Task I: Arrange Fruits.}
\begin{center}
    \includegraphics[width=\linewidth]{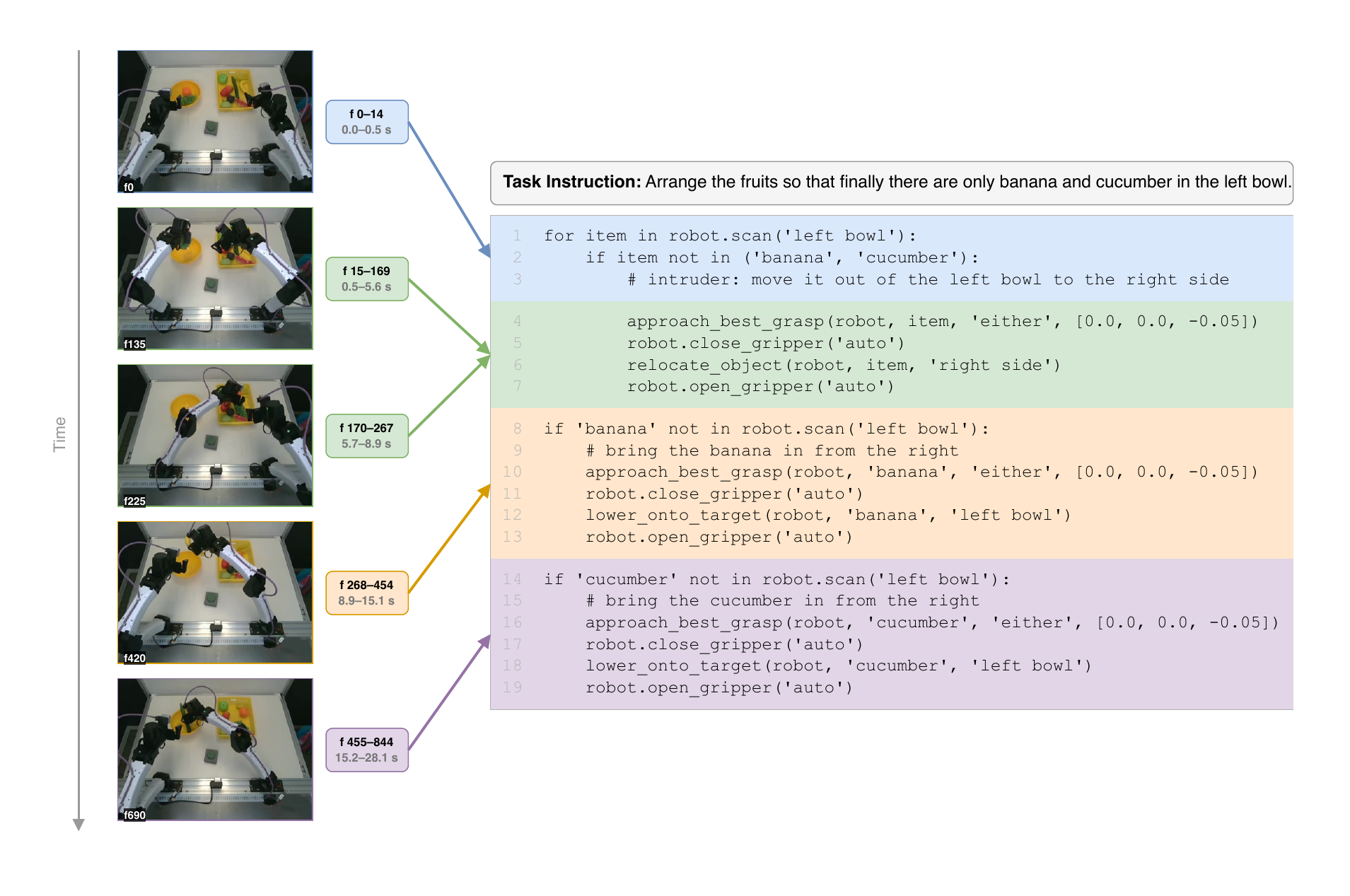}
    \captionof{figure}{Key frames and code label of \textit{Arrange Fruits}.}
    \label{fig:task_arrange_fruits}
\end{center}
\textit{Grading rubric (additive):}
\begin{itemize}
    \item $+0.3$: \textbf{(a)} Moves each fruit not named in the instruction from the left bowl into the right basket (or none to remove).
    \item $+0.3$: \textbf{(b)} Moves each named fruit not yet in the left bowl from the right basket into the bowl (or none to fetch).
    \item $+0.4$: \textbf{(c)} Final state matches the instruction and no fruit is dropped.
\end{itemize}

\needspace{14\baselineskip}
\noindent\textbf{Task II: Pack Backpack.}
\begin{center}
    \includegraphics[width=\linewidth]{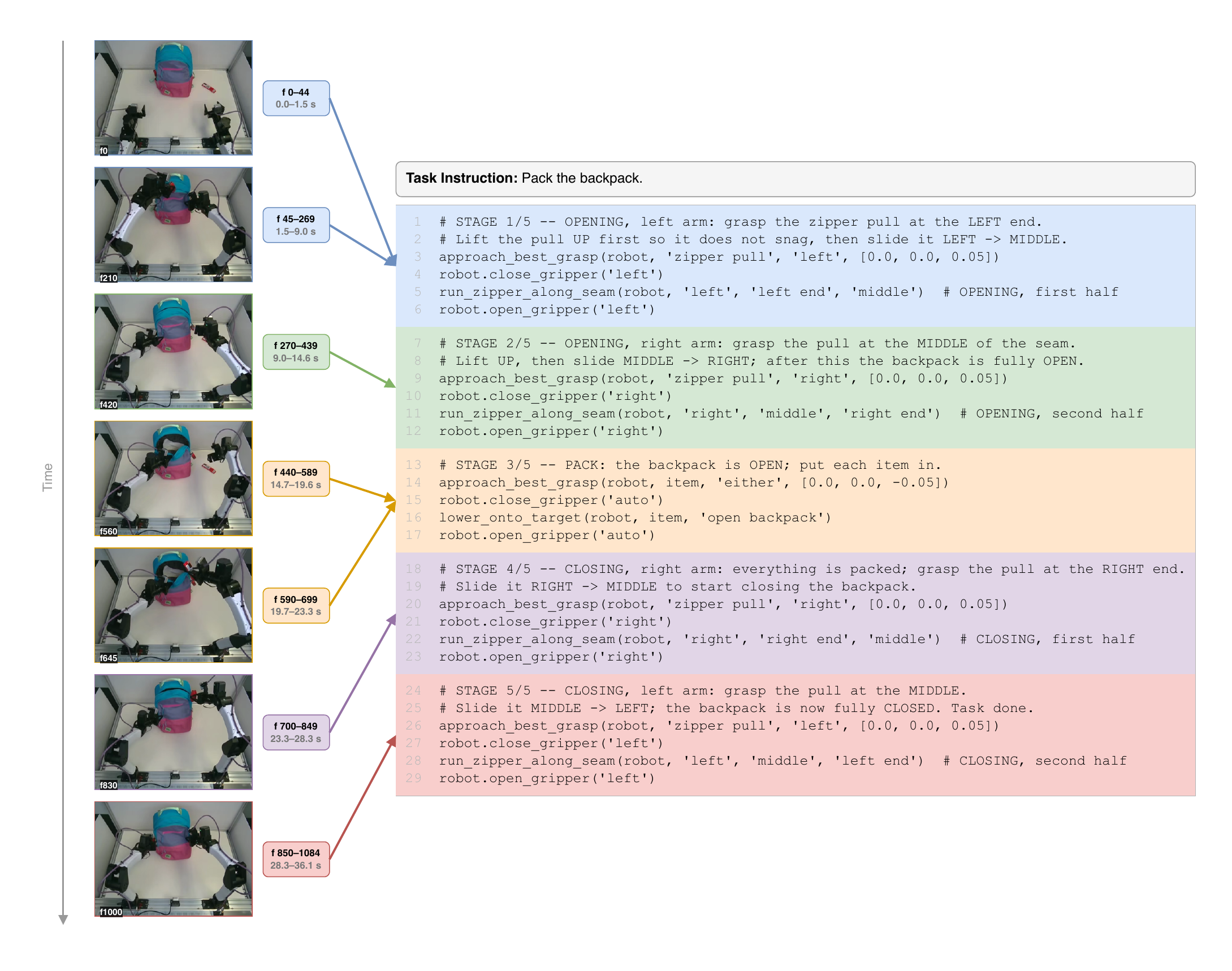}
    \captionof{figure}{Key frames and code label of \textit{Pack Backpack}.}
    \label{fig:task_pack_backpack}
\end{center}
\textit{Grading rubric (additive):}
\begin{itemize}
    \item $+0.15$: \textbf{(a)} Left gripper grasps the zipper pull at the left end and slides it to the middle.
    \item $+0.15$: \textbf{(b)} Right gripper re-grasps the pull at the middle and slides it to the right end.
    \item $+0.2$: \textbf{(c)} Open the backpack with the left gripper.
    \item $+0.2$: \textbf{(d)} Picks up the item from the table and places it inside the open backpack.
    \item $+0.15$: \textbf{(e)} Right gripper grasps the pull at the right end and slides it back to the middle.
    \item $+0.15$: \textbf{(f)} Left gripper slides the pull from the middle to the left end;
          the backpack is fully closed with the item inside.
\end{itemize}

\needspace{14\baselineskip}
\noindent\textbf{Task III: Replace Lightbulb.}
\begin{center}
    \includegraphics[width=\linewidth]{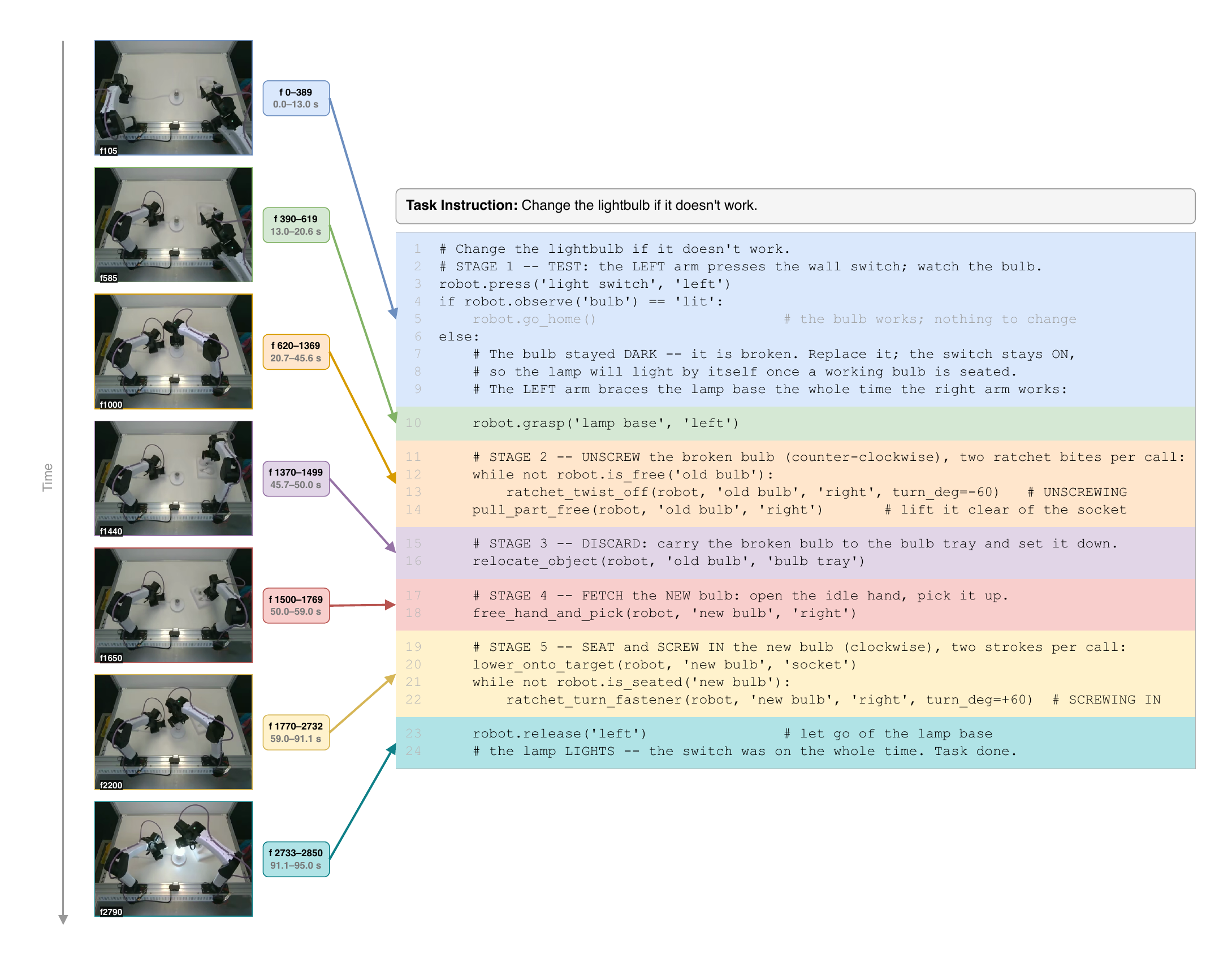}
    \captionof{figure}{Key frames and code label of \textit{Replace Lightbulb}.}
    \label{fig:task_replace_lightbulb}
\end{center}
\textit{Condition A -- dead bulb, replacement required. Grading rubric (additive):}
\begin{itemize}
    \item $+0.1$: \textbf{(a)} Left gripper presses the inline switch and the bulb stays dark.
    \item $+0.1$: \textbf{(b)} Left gripper braces the lamp base.
    \item $+0.2$: \textbf{(c)} Right gripper unscrews the dead bulb and lifts it out of the socket.
    \item $+0.1$: \textbf{(d)} Dead bulb is set down in the bulb tray.
    \item $+0.1$: \textbf{(e)} Right gripper picks up the spare bulb.
    \item $+0.3$: \textbf{(f)} Spare bulb is seated in the socket and screwed in clockwise until lit.
    \item $+0.1$: \textbf{(g)} When the bulb is lit, release both arms to their home positions.
\end{itemize}

\textit{Condition B -- working bulb, no replacement needed. Grading rubric (additive):}
\begin{itemize}
    \item $+0.3$: Left gripper presses the inline switch and the bulb lights up.
    \item $+0.7$: The lit bulb shows the lamp still works, so both arms move back to their initial pose and stay still until the episode ends.
\end{itemize}

\needspace{14\baselineskip}
\noindent\textbf{Task IV: Fold and Stack Towels.}
\begin{center}
    \includegraphics[width=\linewidth]{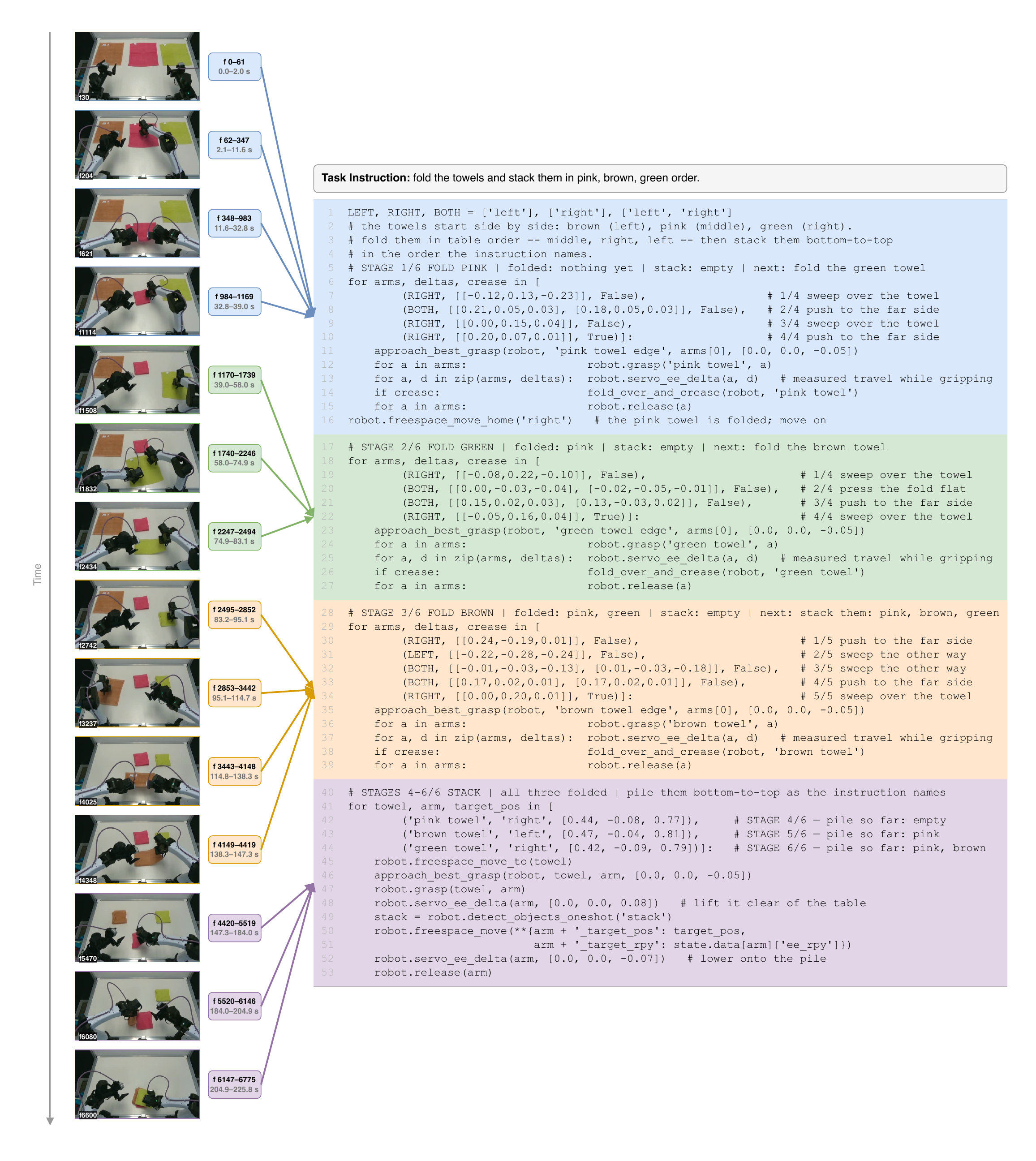}
    \captionof{figure}{Key frames and code label of \textit{Fold and Stack Towels}.}
    \label{fig:task_fold_and_stack_towels}
\end{center}
\textit{Grading rubric (additive):}
\begin{itemize}
    \item $+0.2$: \textbf{(a)} Grasps a towel edge and completes one fold.
    \item $+0.2$: \textbf{(b)} Each of the three towels is folded flat.
    \item $+0.1$: \textbf{(c)} Picks a folded towel up and lays it on the pile.
    \item $+0.2$: \textbf{(d)} All three folded towels end up in a single pile, none dropped or
          left behind.
    \item $+0.3$: \textbf{(e)} The pile reads, bottom to top, in the colour order the instruction
          names.
\end{itemize}

\subsection{Tianji--Sharpa Tasks}

\needspace{14\baselineskip}
\noindent\textbf{Task I: Arrange Fruits with Tongs.}
\begin{center}
    \includegraphics[width=\linewidth]{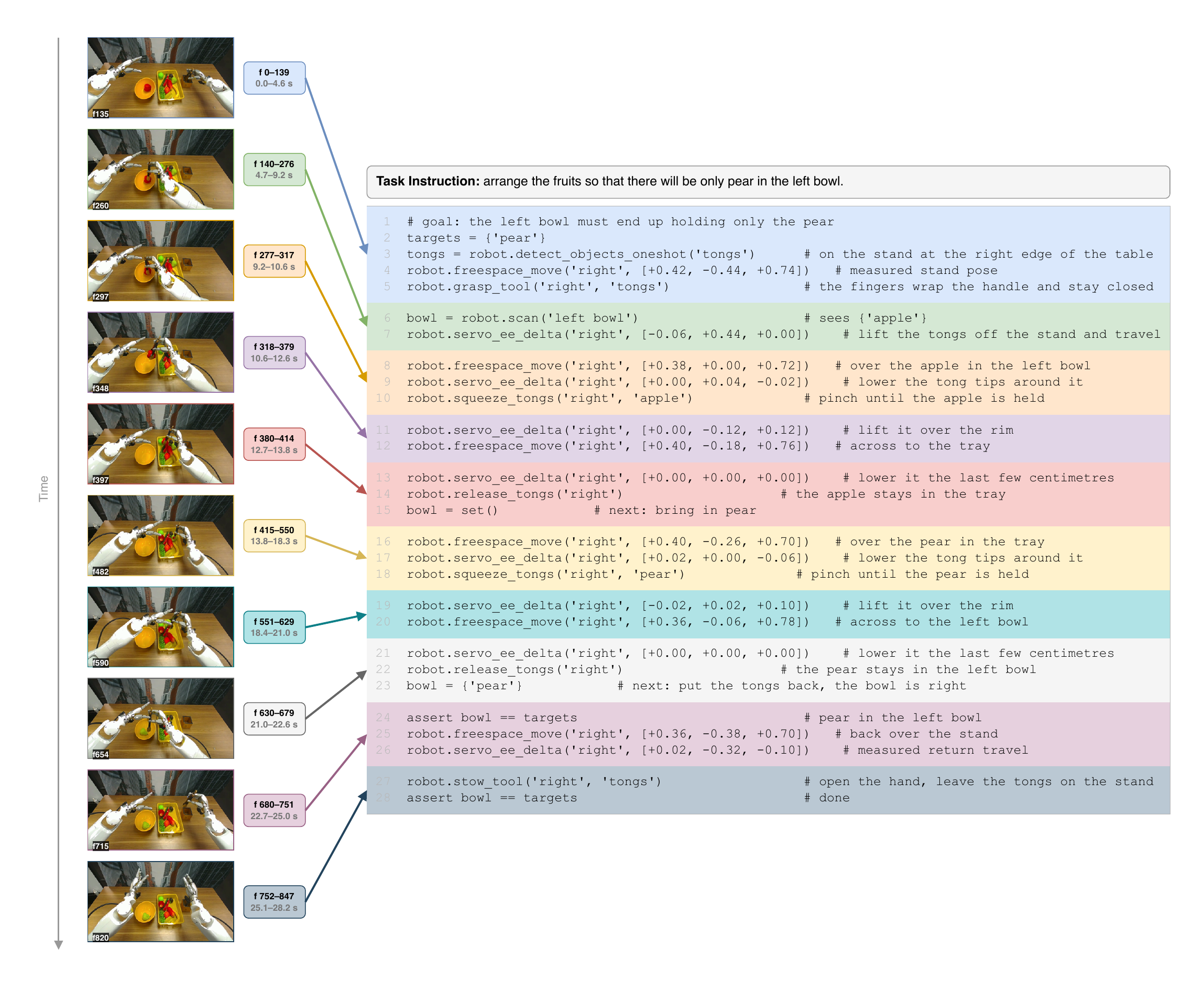}
    \captionof{figure}{Key frames and code label of \textit{Arrange Fruits with Tongs}.}
    \label{fig:task_arrange_fruits_sharpa}
\end{center}
\textit{Grading rubric (additive):}
\begin{itemize}
\item $+0.2$: \textbf{(a)} Right hand takes the tongs off their stand and keeps hold of them.
\item $+0.25$: \textbf{(b)} Moves each fruit not named in the instruction from the left bowl into the tray (or none to remove).
\item $+0.25$: \textbf{(c)} Moves each named fruit not yet in the left bowl from the tray into the bowl (or none to fetch).
\item $+0.2$: \textbf{(d)} Final state matches the instruction and no fruit is dropped.
\item $+0.1$: \textbf{(e)} Tongs are put back on their stand.
\end{itemize}

\needspace{14\baselineskip}
\noindent\textbf{Task II: Deal Poker.}
\begin{center}
    \includegraphics[width=\linewidth]{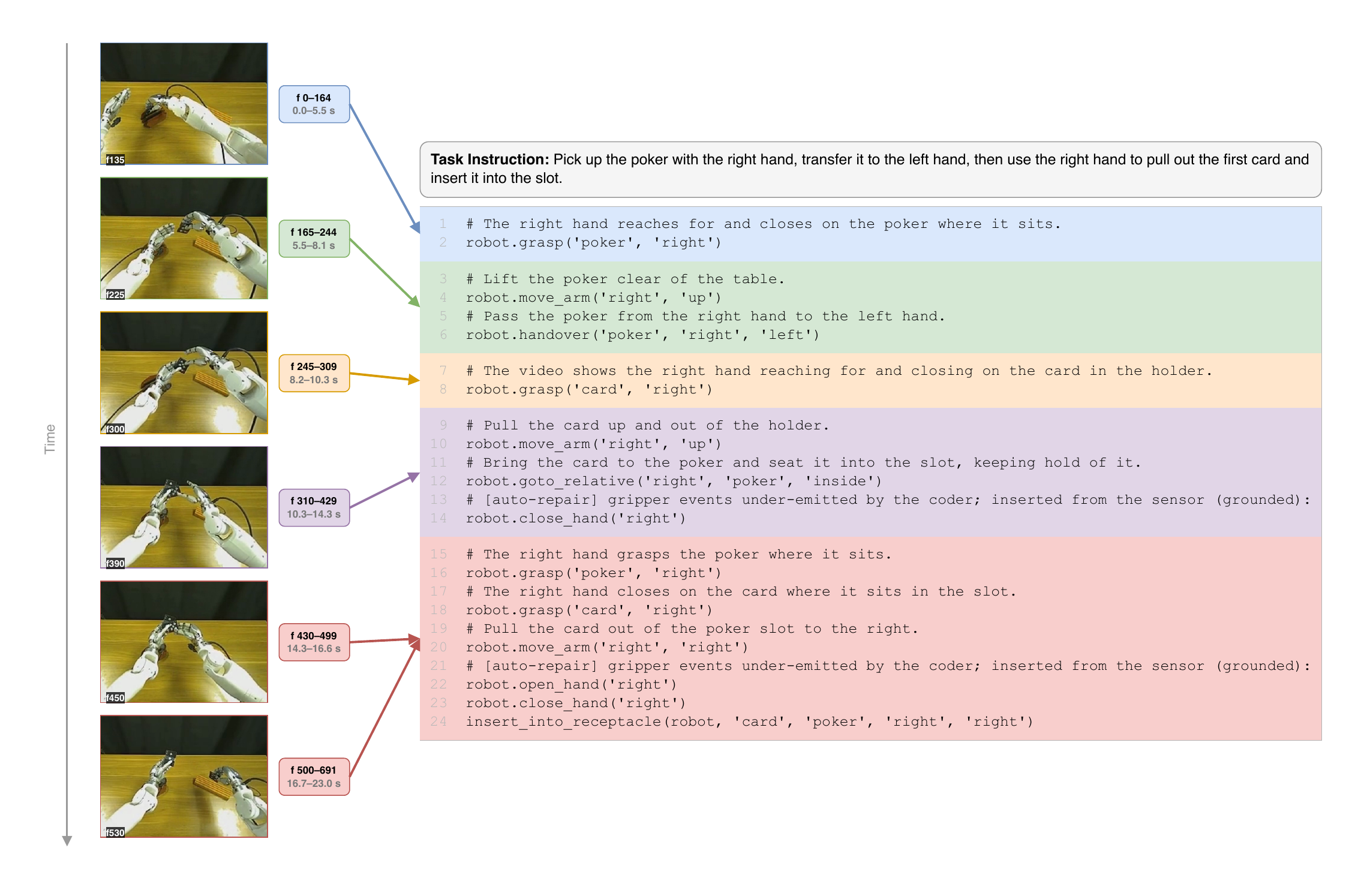}
    \captionof{figure}{Key frames and code label of \textit{Deal Poker}.}
    \label{fig:task_deal_poker}
\end{center}
\textit{Grading rubric (additive):}
\begin{itemize}
    \item $+0.1$: \textbf{(a)} Right hand grasps the card stack from above.
    \item $+0.2$: \textbf{(b)} Right hand transfers the stack to the left hand (handover).
    \item $+0.3$: \textbf{(c)} Right thumb flicks and adjusts until exactly one card protrudes from the top.
    \item $+0.3$: \textbf{(d)} Right hand grasps the single protruding card.
    \item $+0.1$: \textbf{(e)} Right hand inserts the card vertically into the card slot.
\end{itemize}

\needspace{14\baselineskip}
\noindent\textbf{Task III: Flip Scoreboard.}
\begin{center}
    \includegraphics[width=\linewidth]{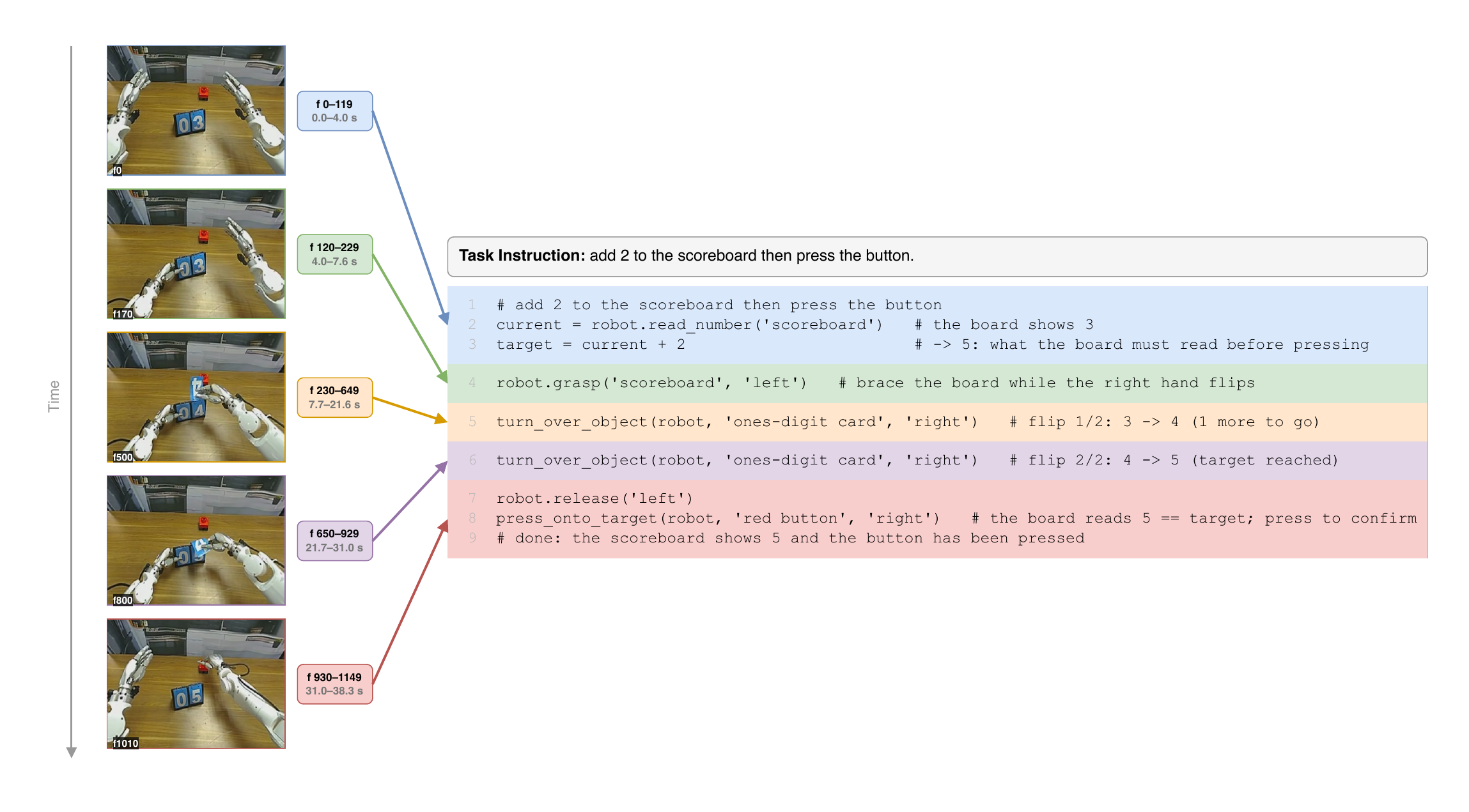}
    \captionof{figure}{Key frames and code label of \textit{Flip Scoreboard}.}
    \label{fig:task_flip_scoreboard}
\end{center}
\textit{Grading rubric (additive):}
\begin{itemize}
    \item $+0.1$: \textbf{(a)} Left hand holds the scoreboard steady.
    \item $+0.2$: \textbf{(b)} Right hand flips a digit card over (at least one complete
          flip).
    \item $+0.4$: \textbf{(c)} Flips exactly $k$ times, so the board shows the initial score plus $k$.
    \item $+0.1$: \textbf{(d)} Left hand releases the board.
    \item $+0.2$: \textbf{(e)} Right hand presses the red button while the target score is
          showing.
\end{itemize}

\needspace{14\baselineskip}
\noindent\textbf{Task IV: Solve Rubik's Cube.}
\begin{center}
    \includegraphics[width=\linewidth]{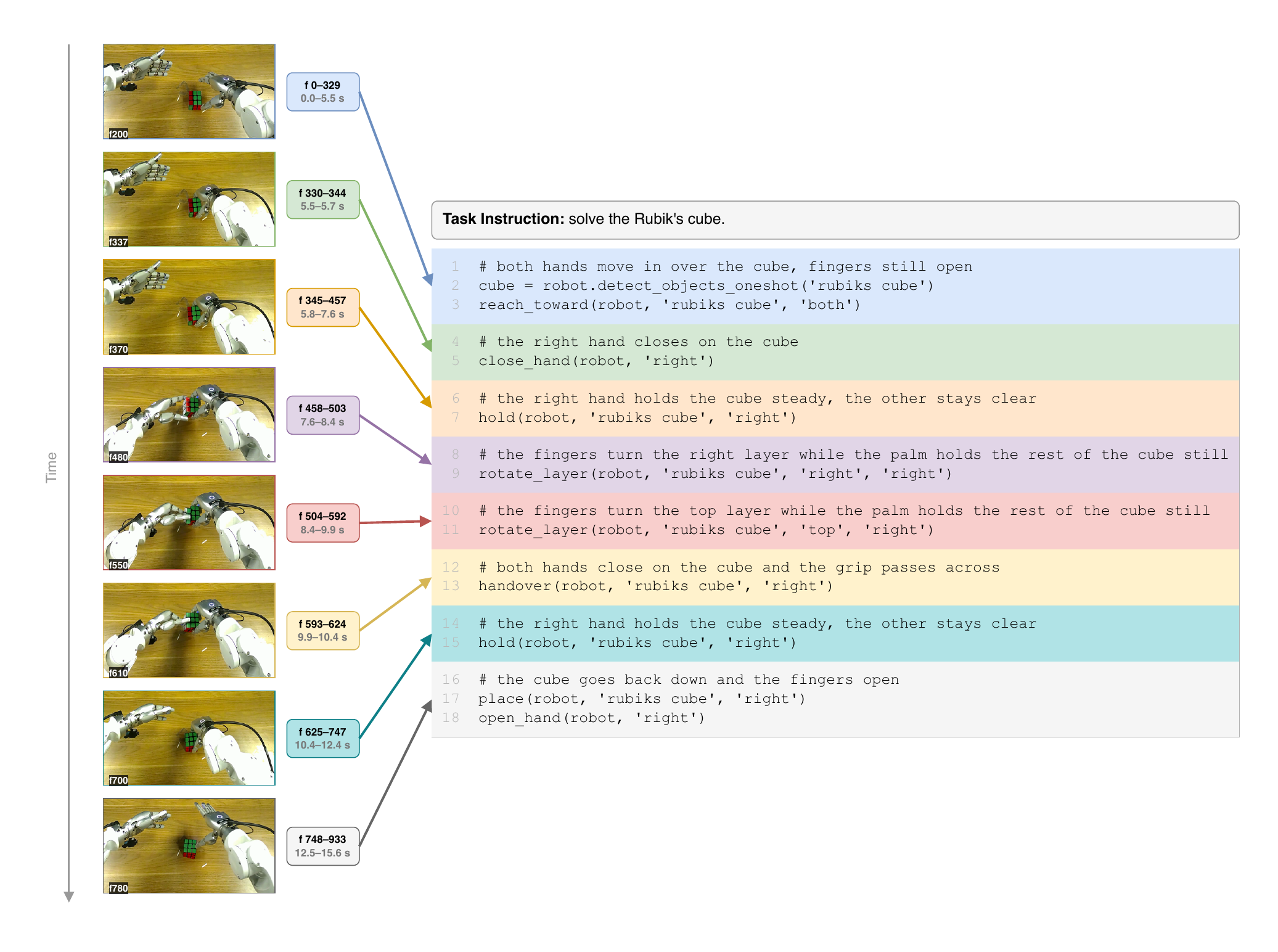}
    \captionof{figure}{Key frames and code label of \textit{Solve Rubik's Cube}.}
    \label{fig:task_solve_rubiks_cube}
\end{center}
For each rollout, one randomly selected face is scrambled clockwise or counterclockwise; the visible face colors and placement angle are fully randomized, with no fixed layer or turn sequence.

\textit{Grading rubric (additive):}
\begin{itemize}
\item $+0.2$: \textbf{(a)} Right hand grasps the cube and lifts it off the table.
\item $+0.3$: \textbf{(b)} Fingers turn the scrambled layer a full quarter turn, leaving no layer half-turned.
\item $+0.3$: \textbf{(c)} The cube is solved: every face shows a single colour.
\item $+0.2$: \textbf{(d)} The cube is put back on the table and both hands let go.
\end{itemize}

\section{Analysis on Code Generation Quality}
\label{app:code_generation_quality}

\subsection{Pre-trained Code Generation Quality}
\label{app:pretrained_code_quality}

\begin{figure}[htbp]
    \centering
    \includegraphics[width=\linewidth]{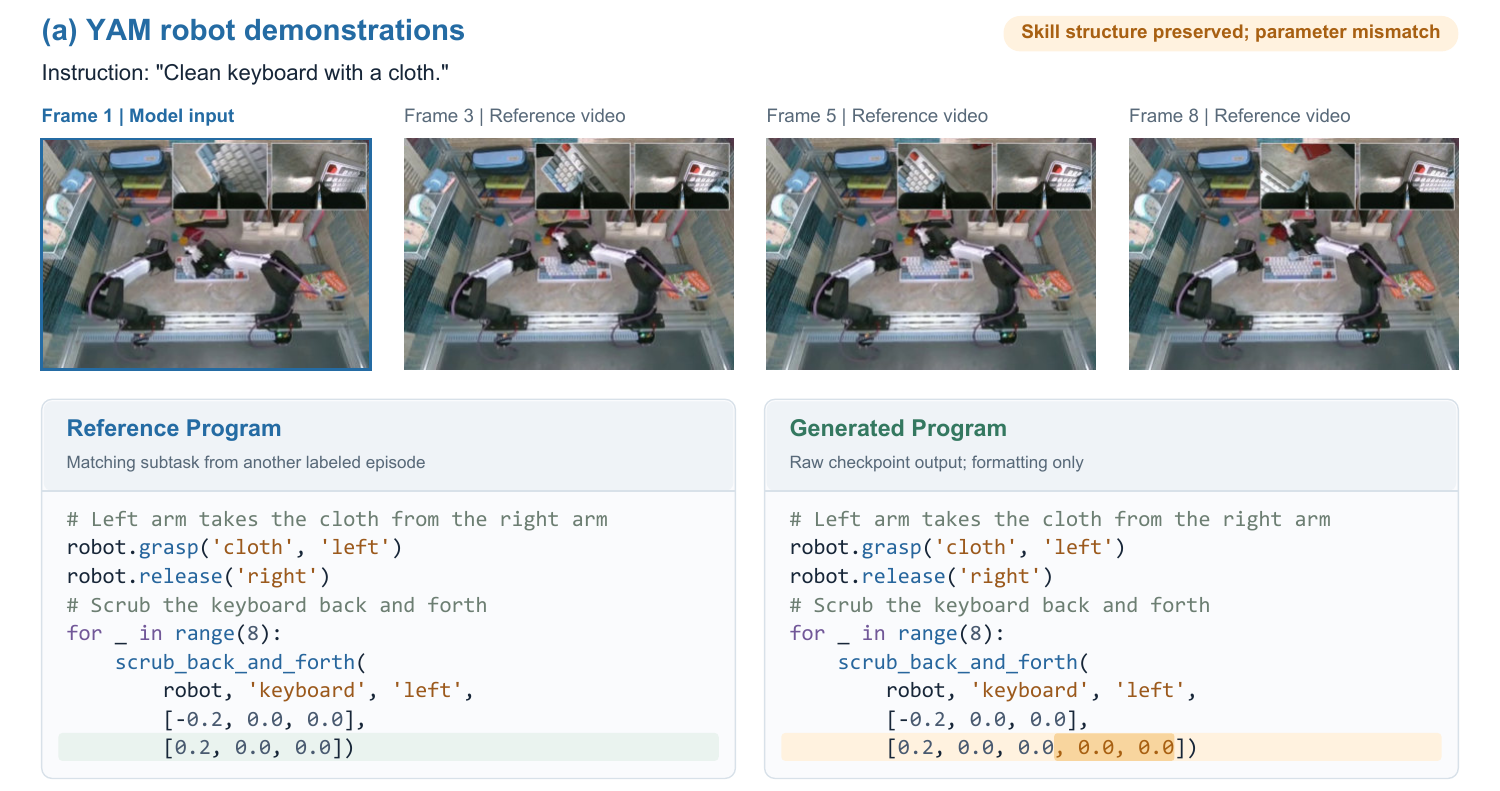}
    \caption{\textbf{Pretrained Code Generation on a YAM Validation Example.} The generated program is compared with a reference program from another labeled episode of the same task.}
    \label{fig:pretrained_code_yam}
\end{figure}

\begin{figure}[htbp]
    \centering
    \includegraphics[width=\linewidth]{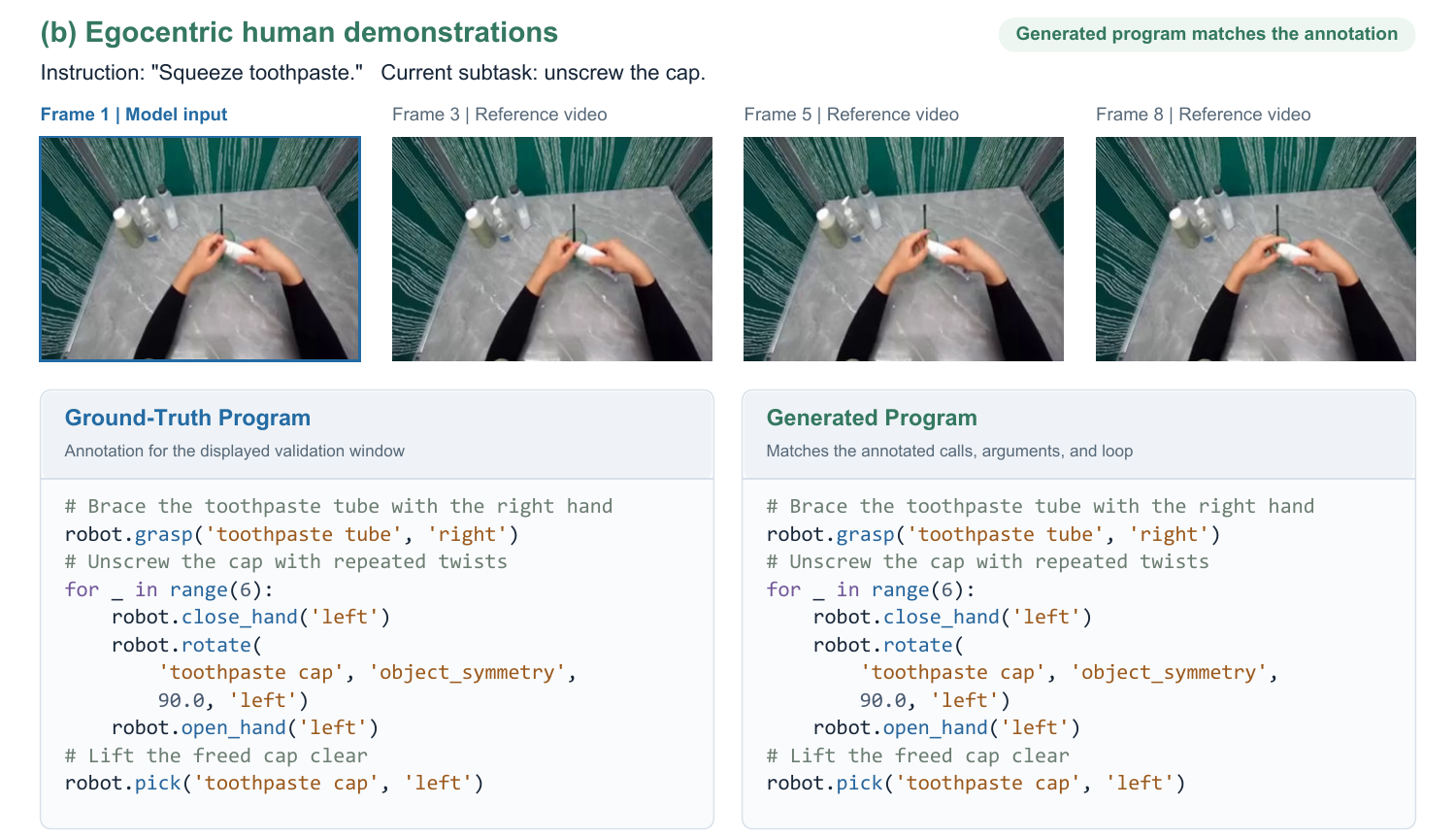}
    \caption{\textbf{Pretrained Code Generation on an Egocentric Human Validation Example.} Under the instruction \textit{``Squeeze toothpaste,''} the model generates the cap-removal subtask, matching the annotated program for the held-out window.}
    \label{fig:pretrained_code_human}
\end{figure}

We evaluate the pretrained code expert on the validation split of the pretraining dataset, before downstream task-specific fine-tuning. Table~\ref{tab:pretrained_code_quality} reports quantitative metrics on this split, while Figures~\ref{fig:pretrained_code_yam} and~\ref{fig:pretrained_code_human} present generated programs for YAM and egocentric human validation examples. We examine visual grounding, program structure, and semantic agreement with the reference annotations.

\begin{table}[htbp]
\centering
\caption{\textbf{Validation Metrics for the Pretrained Code Expert.} TF denotes teacher-forced evaluation. The grounding gap is the absolute increase in code loss when visual observations are shuffled across validation samples.}
\label{tab:pretrained_code_quality}
\small
\setlength{\tabcolsep}{3pt}
\begin{tabular*}{\linewidth}{@{\extracolsep{\fill}}ccccccc@{}}
\toprule
\shortstack{TF\\loss} & \shortstack{TF\\accuracy} & \shortstack{Grounding\\gap} & \shortstack{Parse\\rate} & \shortstack{Entity\\F1} & \shortstack{Primitive\\F1} & \shortstack{Skill-call\\rate} \\
\midrule
0.535 & 0.897 & 1.774 & 0.875 & 0.534 & 0.365 & 0.292 \\
\bottomrule
\end{tabular*}
\end{table}

\textbf{Quantitative Assessment.}
The model achieves 89.7\% teacher-forced token accuracy and an 87.5\% parse rate under free generation. Shuffling visual observations increases the code loss by 1.774, indicating that code prediction depends on the visual context. Entity and primitive F1 are 0.534 and 0.365, respectively, measuring overlap of quoted strings and robot API names with the reference programs. These scores indicate partial semantic agreement despite the higher syntactic validity. In addition, 29.2\% of generated programs contain skill-style helper calls, identified by calls that take \texttt{robot} as their first argument. Together, these results show that pretraining learns visually conditioned program generation and skill invocation, while precise semantic and argument prediction remain areas for improvement.

\textbf{YAM Robot Demonstrations.}
Figure~\ref{fig:pretrained_code_yam} shows a held-out YAM demonstration of \textit{``Clean keyboard with a cloth"}. The generated program expresses a cloth handover followed by repeated scrubbing through \texttt{scrub\_back\_and\_forth}, preserving the object arguments, arm assignment, and loop structure of the reference subtask. This illustrates how a reusable skill can encode a repeated motion compactly. The generated displacement argument nevertheless contains two extra entries, illustrating that an appropriate skill-level plan can still contain parameter errors that prevent reliable direct execution.

\textbf{Egocentric Human Demonstrations.}
Figure~\ref{fig:pretrained_code_human} illustrates subtask-level planning under the instruction \textit{``Squeeze toothpaste.''} The generated program describes the preparatory subtask of removing the cap: stabilizing the tube, repeatedly twisting the cap, and lifting it away. It matches the annotated calls, arguments, and loop structure for this validation window. The two examples illustrate how the pretrained model represents manipulation through explicit object references, ordered calls, and repetition across robot and human observations, while the aggregate metrics distinguish this representational capability from uniformly correct or executable code.

\begingroup
\raggedbottom
\subsection{Post-trained Code Generation Quality}
\label{app:posttrained_code_quality}

We examine code generated online by the post-trained model during real-world deployment on the YAM and Tianji--Sharpa platforms. Figures~\ref{fig:posttrained_code_yam} and~\ref{fig:posttrained_code_sharpa} show examples from \textit{Pack Backpack} and \textit{Solve Rubik's Cube}, respectively. Each example pairs the observation at a replanning point with the generated program and subsequent recorded frames from the same deployment rollout. Reference programs are drawn from separate annotated demonstrations of the same task and subtask, providing qualitative comparisons of skill selection, object arguments, and call structure.

\textbf{YAM Deployment Rollout.}
Figure~\ref{fig:posttrained_code_yam} shows \textit{Pack Backpack} at the stage where the right arm grasps the zipper pull at the middle of the seam. The generated program invokes \texttt{approach\_best\_grasp} for the zipper pull, followed by \texttt{robot.close\_gripper('right')}. The generated program uses the same grasping skill, target object, arm assignment, and numerical arguments as the reference program. This example illustrates how the post-trained model expresses a task-specific manipulation step through a reusable grasping skill and an explicit gripper command.

\begin{figure}[H]
    \centering
    \includegraphics[width=\linewidth]{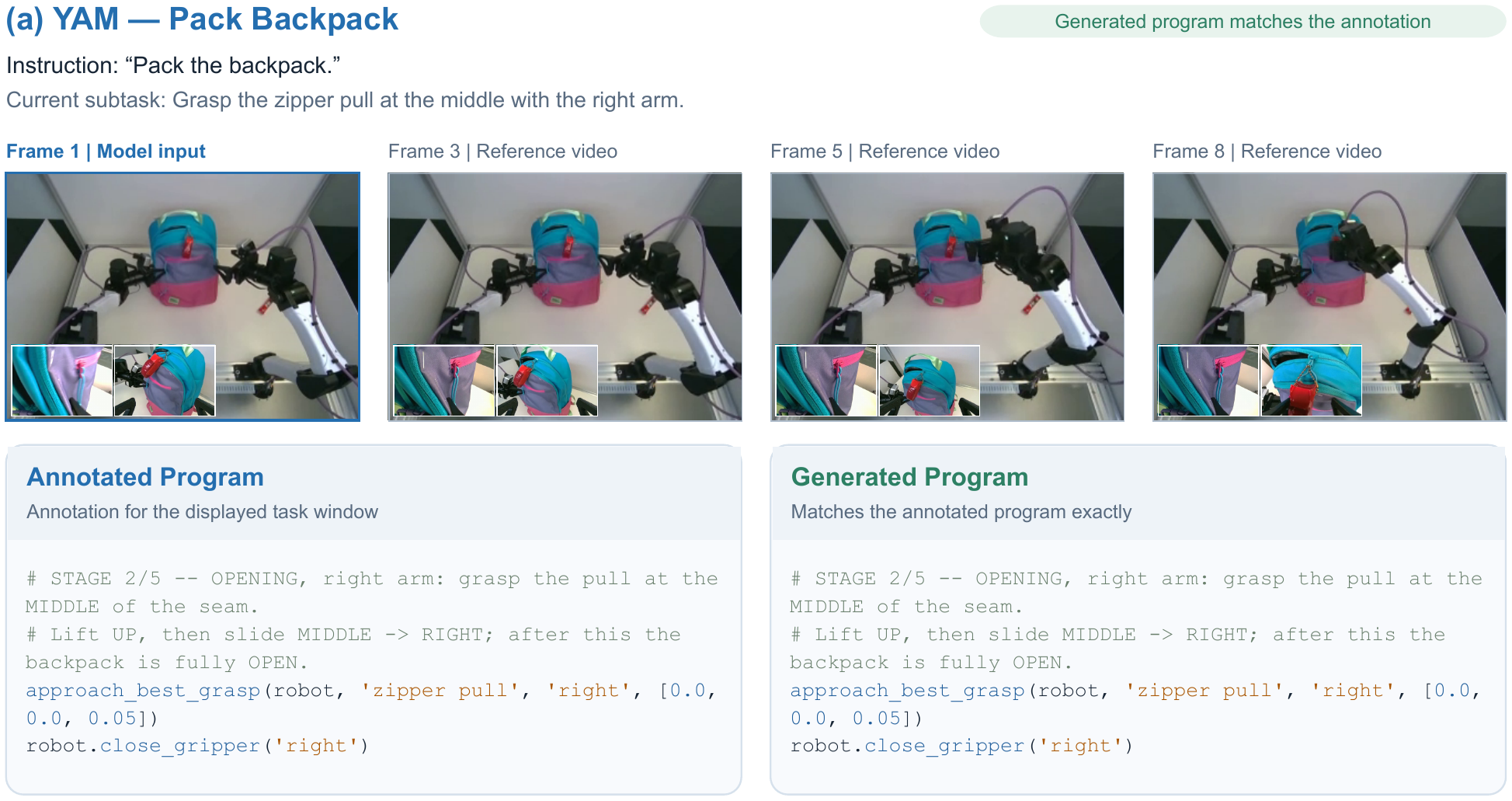}
    \caption{\textbf{Online Code Generation during Real-world Deployment on \textit{Pack Backpack.}} The panel labeled “Annotated Program” shows a reference program from a separate demonstration of the same task and subtask. “Generated Program” is the model output recorded online during deployment. The comparison illustrates semantic correspondence in skill selection and arguments.}
    \label{fig:posttrained_code_yam}
\end{figure}

\textbf{Tianji--Sharpa Deployment Rollout.}
Figure~\ref{fig:posttrained_code_sharpa} illustrates subtask-level planning for \textit{Solve Rubik's Cube} while the right hand holds the cube steady. The generated program reproduces \texttt{hold(robot, 'rubiks cube', 'right')}, matching the annotated skill, object, and hand assignment. The generated skill call agrees with the reference program in skill choice, object, and hand assignment. The output nevertheless contains a malformed comment: the word \texttt{cube} is omitted, and the comment continues on a new line without a comment marker. This formatting error prevents direct execution of the raw output despite the correct skill call. Together, the two examples illustrate task-specific semantic agreement across parallel grippers and dexterous hands, while distinguishing appropriate skill selection from uniformly valid executable code.

\begin{figure}[H]
    \centering
    \includegraphics[width=\linewidth]{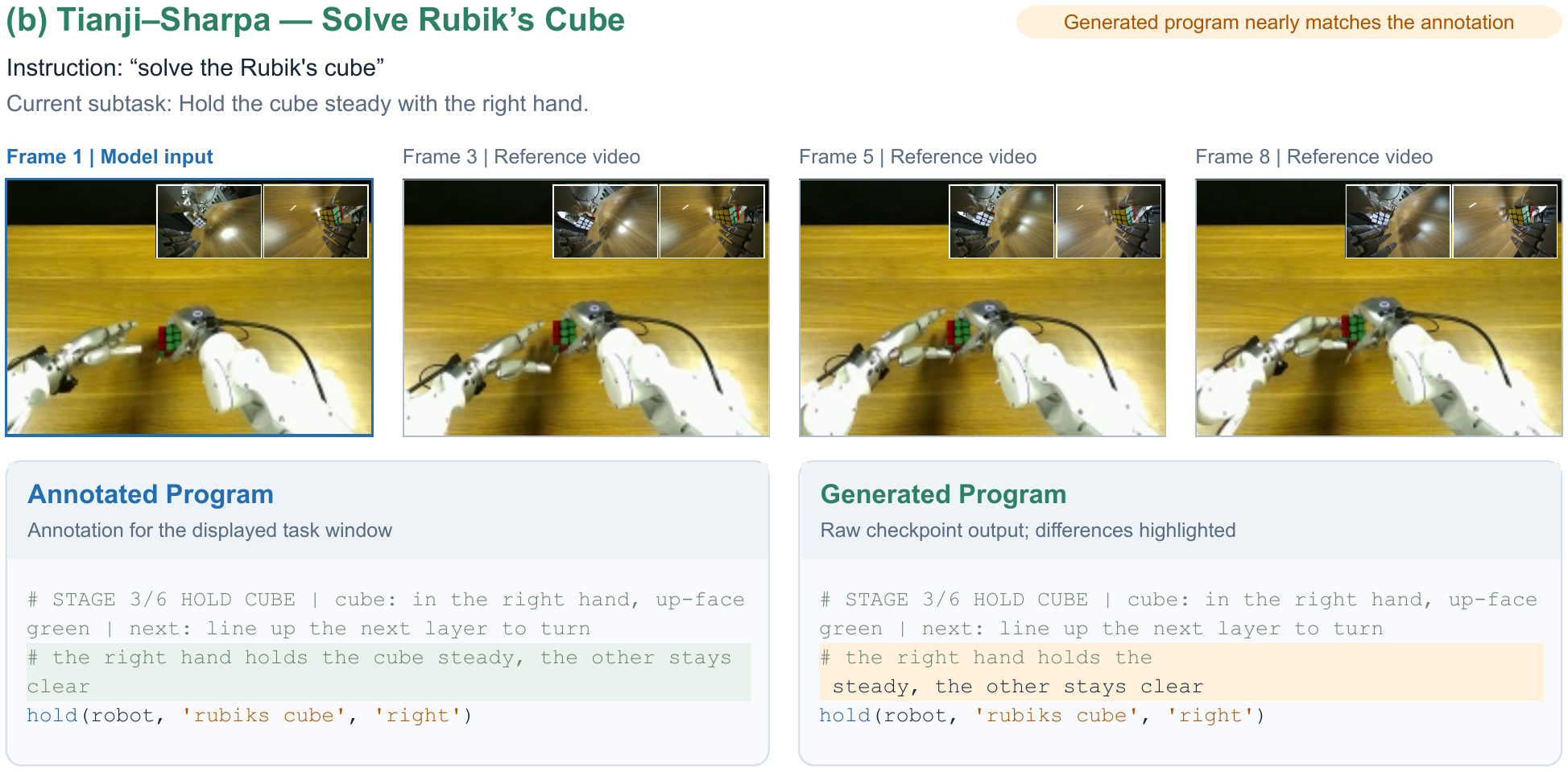}
    \caption{\textbf{Online Code Generation during Real-world Deployment on \textit{Solve Rubik’s Cube}.} The panel labeled “Annotated Program” shows a reference program from a separate demonstration of the same task and subtask. “Generated Program” is the model output recorded online during deployment. The comparison illustrates semantic correspondence in skill selection and arguments.}
    \label{fig:posttrained_code_sharpa}
\end{figure}
\clearpage
\endgroup

\section{Additional Analysis on Real-world Results}
\label{app:additional_real_results}

\subsection{Impact of Code Expert on Language Following Capability}

\label{apdx:language_following}

\begin{figure}[t]
\centering
\includegraphics[width=\linewidth]{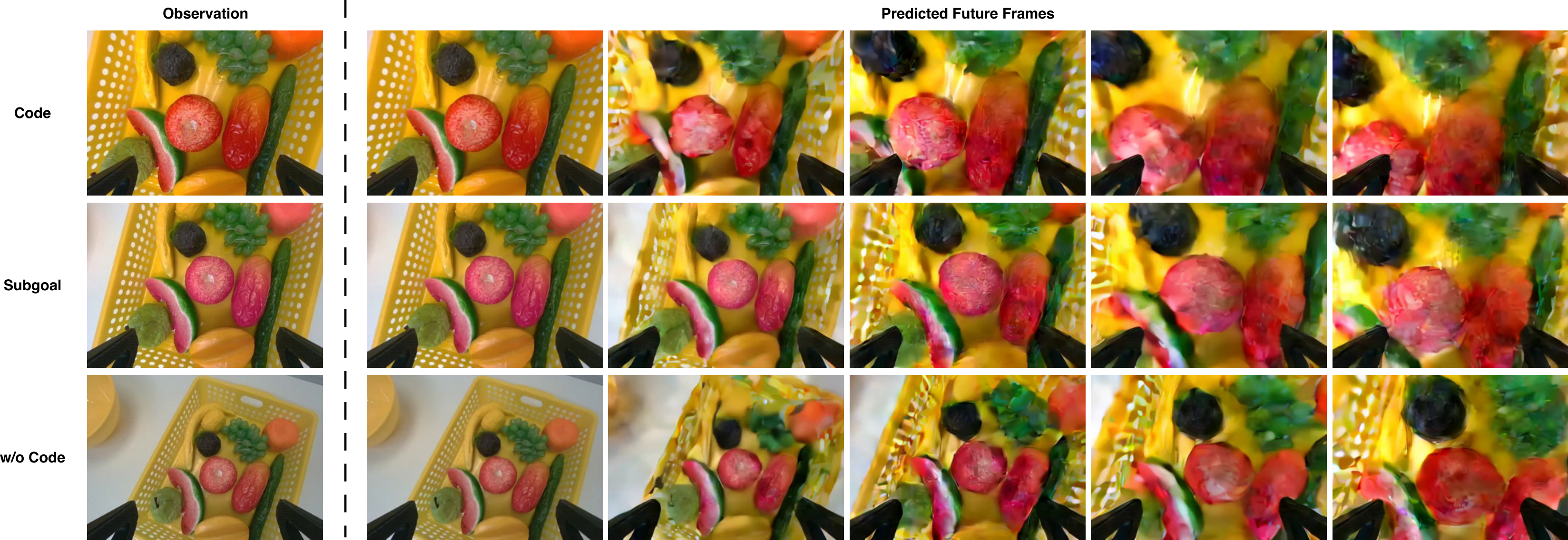}
\caption{\textbf{Future-frame Prediction on Arrange Fruits for Three Models.}}
\label{fig:video_quality_comparison}
\end{figure}

We further examine whether code-based planning improves the alignment of predicted visual dynamics with task semantics. As shown in Figure~\ref{fig:video_quality_comparison}, we compare the video generation quality of the right wrist camera of \Ours{} with a natural-language subgoal variant (Subgoal) and a model without a planning stream (\Ours{} w/o code), under the instruction \textit{``Arrange the fruits so that finally there is only mango in the left bowl.''} This task requires identifying the intended object and grounding the desired outcome in an appropriate interaction. The three variants are evaluated under the same task and scene configuration, with similar but non-identical initial observations from separate rollouts.

The predicted futures reveal distinct differences in task understanding: \Ours{} directs the interaction toward the mango, whereas Subgoal exhibits ambiguous targeting between the apple and mango, and \Ours{} w/o code targets the apple. These observations suggest that structured skill calls and explicit object arguments provide a more specific planning condition for video prediction than a natural-language subgoal alone. By connecting instruction semantics to concrete interactions, the code stream helps align predicted dynamics with the intended task. Since the action expert also conditions on predictive video representations, this alignment provides a pathway for improving language following, complementing the execution results in our main experiments.

\subsection{Impact of Code Expert on Skill Robustness}
\label{apdx:video_quality}

\begin{figure}[t]
\centering
\includegraphics[width=\linewidth]{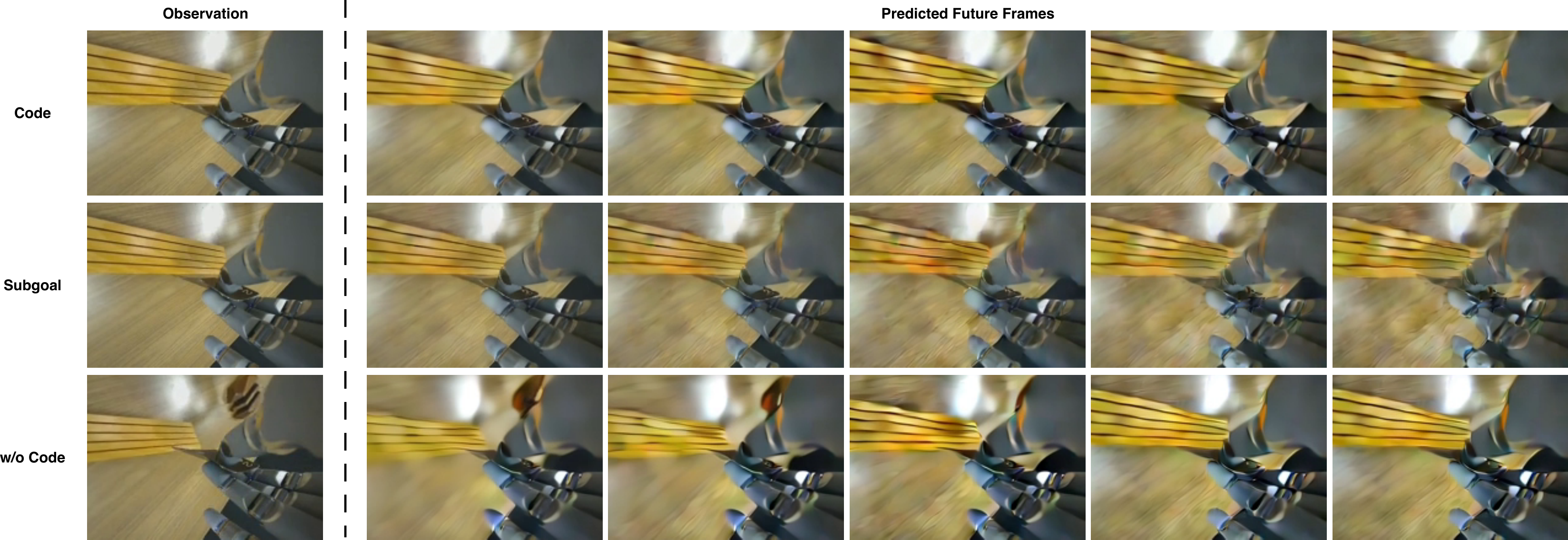}
\caption{\textbf{Future-frame Prediction on Deal Poker for Three Models.}}
\label{fig:video_quality_deal_poker}
\end{figure}

We further examine whether code-based planning improves the prediction of physical dynamics and skill robustness during precise manipulation. As shown in Figure~\ref{fig:video_quality_deal_poker}, we compare the future frames predicted by \Ours{}, Subgoal, and \Ours{} w/o code for the right wrist camera on \textit{Deal Poker}, focusing on the final stage in which the right hand inserts a card into a slot. This interaction requires aligning the held card with the narrow opening and guiding it into the slot, making the relative motion of the hand, card, and receptacle critical to an accurate prediction. The three variants receive similar but non-identical initial observations sampled at nearby time steps from the same episode.

The predicted futures reveal distinct differences in the insertion motion: \Ours{} depicts the card aligning with and entering the slot more accurately, whereas Subgoal and \Ours{} w/o code produce less coherent insertion sequences, with blur and distortions around the hand and manipulated objects. These observations suggest that explicit skill calls and object arguments help the video expert better predict how the hand and card should move relative to the receptacle. The implementation of the code expert supports more accurate video generation, and therefore provides the action expert with more reliable predictive guidance for precise manipulation.

\subsection{Hybrid Actions Evaluation Details}
\label{app:hybrid_action_eval}

\begin{figure}[t]
\centering
\includegraphics[width=\linewidth]{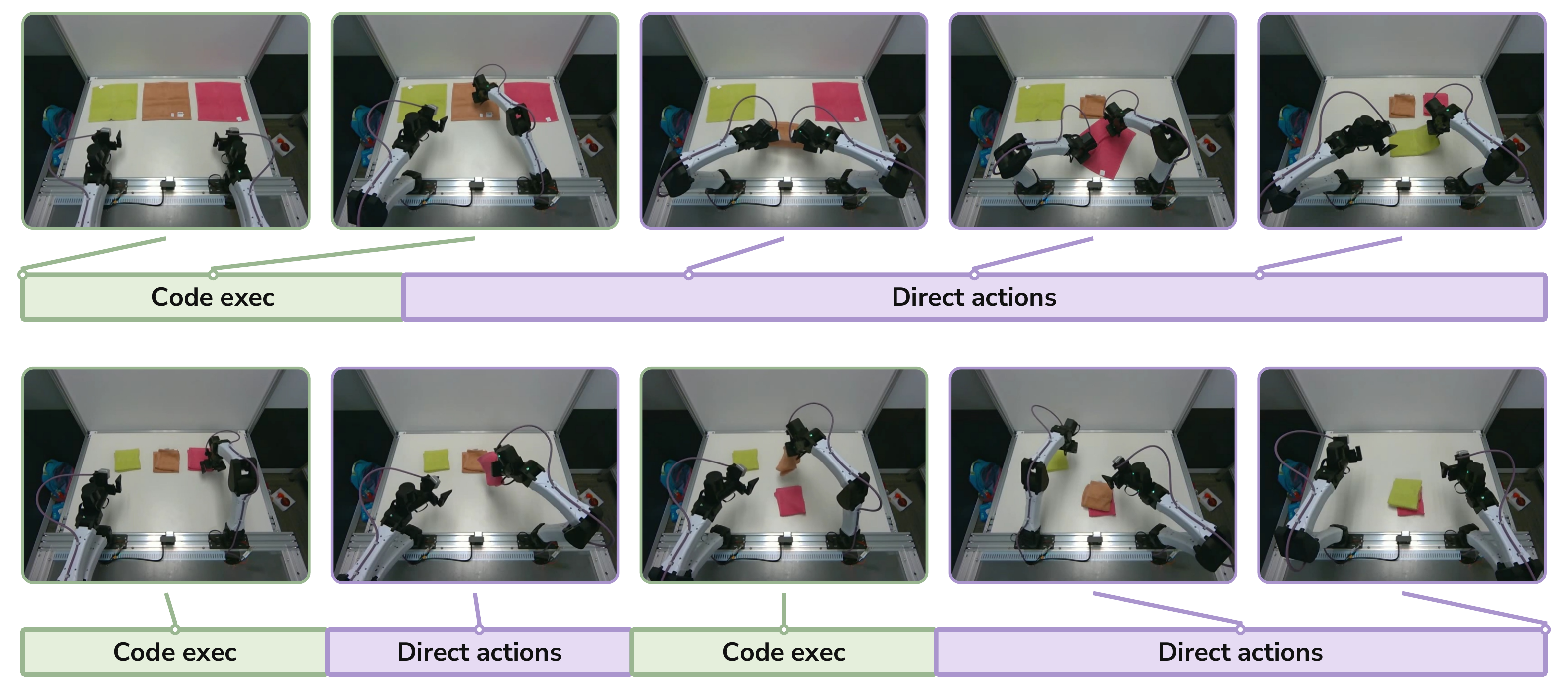}
\caption{\textbf{Qualitative Example of Hybrid Actions for the \textit{``Fold and Stack Towels"} Task.} A single episode continues from the upper to the lower row, with green and purple indicating code execution and direct actions, respectively.}
\label{fig:real_hybrid_progress}
\end{figure}
Figure~\ref{fig:real_hybrid_progress} is a qualitative example of hybrid evaluation on the \textit{``Fold and Stack Towels"} task, showing a single episode that continues from the upper to the lower row. We observe that dexterous folding motions rely primarily on direct actions from the action expert of \Ours{}, while code execution handles more structured motions, such as approaching and stacking the towels, which provide direct guidance on the stacking order and motion direction, improving the semantic reasoning capability for the whole system.

\section{Failure Case Analysis}
\label{app:failure_case_analysis}

\begin{figure}[t]
\centering
\includegraphics[width=\linewidth]{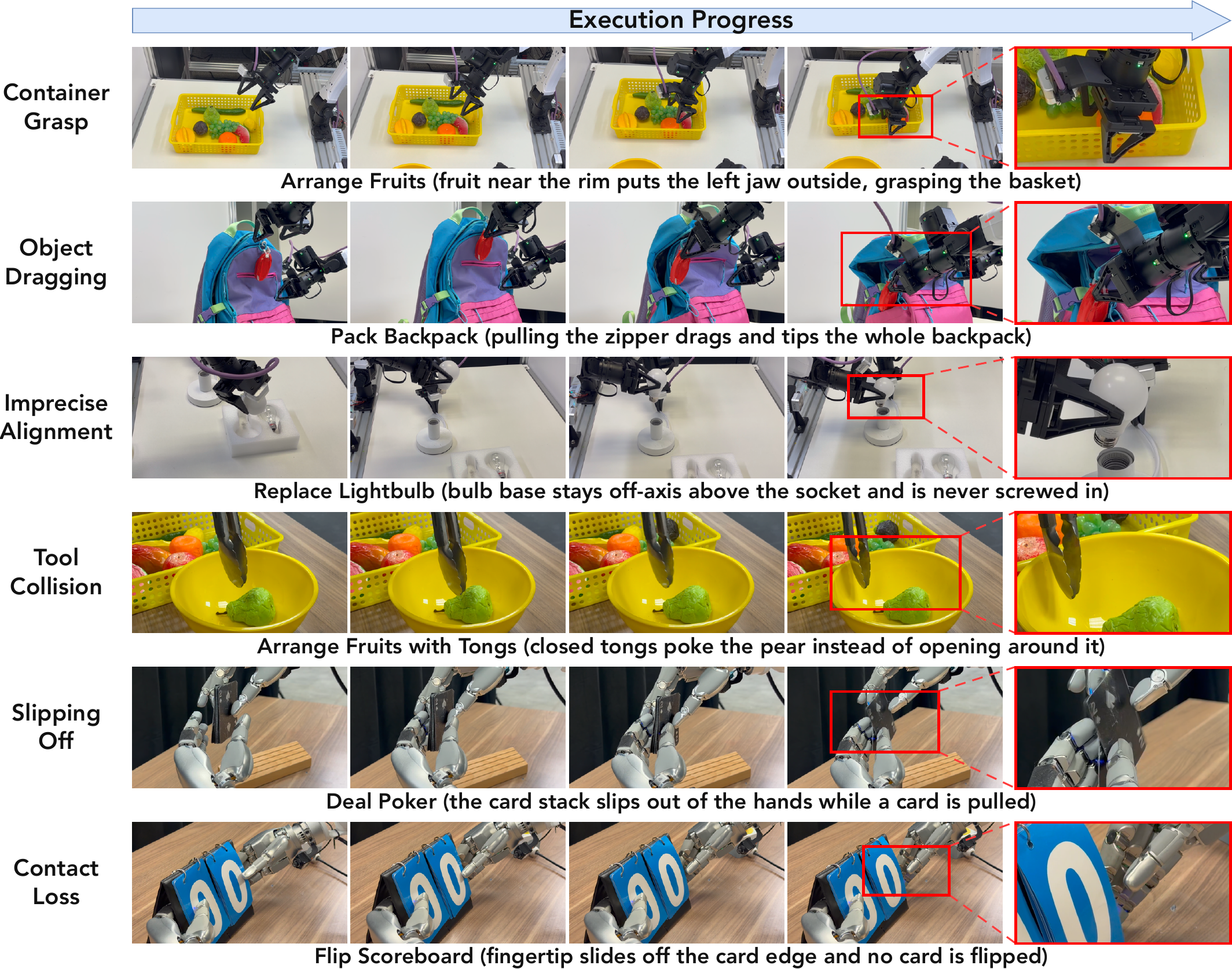}
\caption{\textbf{Failure Case Analysis.} Each row shows one representative failure case. Frames are ordered from left to right by execution progress. The red box on the last frame marks the region enlarged in the rightmost panel, where the failure occurs.}
\label{fig:failure_case_analysis}
\end{figure}
Figure~\ref{fig:failure_case_analysis} shows representative failure cases across various scenarios. The red boxes highlight the local interactions where execution fails. These examples illustrate some remaining challenges despite the overall performance gains of \Ours{}.

\textbf{1) Container Grasp.}
In \textit{Arrange Fruits} in the first row, the target fruit lies close to the basket rim. As the gripper approaches, its left jaw moves outside the basket, causing the closing motion to grasp the basket wall rather than only the fruit. This case highlights the need to account for surrounding geometry when selecting a grasp. Explicitly estimating the rim geometry and checking clearance along both the approach and closing paths could help reject obstructed grasps and select a more accurate approach direction.

\textbf{2) Object Dragging.}
In \textit{Pack Backpack} in the second row, when the robot pulls the zipper, the backpack moves and tips over rather than remaining stable. The visible failure suggests insufficient stabilization of the deformable object during manipulation. A potential improvement is to coordinate the supporting and pulling motions through more frequent visual feedback: the supporting hand could adjust its grasp as the bag deforms, while the pulling hand could slow down or pause when the bag starts to shift or tip.

\textbf{3) Imprecise Alignment.}
In \textit{Replace Lightbulb} in the third row, the robot grasps the bulb and brings it above the socket, but the bulb base remains misaligned with the socket opening and is not screwed in. Completing this step requires both positional and angular alignment before the threads can engage. The case illustrates a gap between moving an object to the correct region and achieving the precise relative pose required for insertion and rotation.

\textbf{4) Tool Collision.}
In \textit{Arrange Fruits with Tongs} in the fourth row, the tongs approach the pear with their tips closed and poke it rather than opening around it to establish a grasp. Successful tool use requires coordinating the tool's opening with its approach trajectory, in addition to holding and positioning the tool itself. Explicitly tracking the tong tips and opening width could help condition actions on the tool's state. A feedback check that the tips are open and positioned on opposite sides of the fruit before closing could further reduce unsuccessful contact.

\textbf{5) Slipping Off.}
In \textit{Deal Poker} in the fifth row, the left hand fails to secure the card stack during handover from the right hand, causing the cards to slip. A successful handover requires the receiving hand to establish a stable grasp before the giving hand releases the stack. This failure illustrates the difficulty of coordinating grasping and release across the two hands while maintaining support for the cards throughout the transfer.

\textbf{6) Contact Loss.}
In \textit{Flip Scoreboard} in the sixth row, the fingertip reaches the edge of a score card but slides off before turning it over, leaving the displayed score unchanged. Unlike a brief touch, a successful flip requires sustained contact along the card's rotation about the binding. This case illustrates the difficulty of sliding the fingertip along the first card to lift its edge while maintaining continuous contact.

\section{Hybrid Actions Algorithm}
\label{app:hybrid_action_algorithm}

Algorithms~\ref{alg:hybrid_episode} and~\ref{alg:hybrid_arbitrate} outline the workflow of the hybrid mode. From an observation $o$ and instruction $\ell$, one forward pass of the policy $\pi$ returns a direct action chunk $a_{1:H}$ over the replan horizon $H$ and a code slice $c$. Every slice executes in a shared namespace $\mathcal{G}$ containing the environment's skill API, reusable helper definitions, and its setup block (episode-start perception), which persists throughout the episode. At each replanning point, \textsc{Arbitrate} decides which of the two channels drives the robot. The returned flag indicates whether code execution was attempted, not whether it succeeded. If execution is attempted, the action chunk from the same policy invocation is discarded, even if an exception occurs; the next iteration observes the current state and replans. Otherwise, the chunk is played for $H$ steps as in the default mode.

The original generated slice $c$ is checked without repair. It is executed only if it passes the following gates in order: 
\begin{enumerate}
    \item \textbf{Stage Gate}: its stage is not complete, is not below a stage already attempted, and has fewer than $R$ failed execution attempts;
    \item \textbf{Dependency Gate}: every variable it reads is defined in $\mathcal{G}$; and
    \item \textbf{Syntax, Whitelist, and Vacuity Gates}: it compiles as generated, calls only whitelisted functions, and contains an executable statement.
\end{enumerate}

\begin{algorithm}[t]
\caption{Hybrid control: generated code first, direct actions otherwise}
\label{alg:hybrid_episode}
\begin{algorithmic}[1]
\REQUIRE policy $\pi$, environment $E$, execution namespace $\mathcal{G}$, instruction $\ell$, replan horizon $H$, per-stage failure limit $R$
\STATE $\mathcal{D} \leftarrow \emptyset$, \ $n_k \leftarrow 0\ \forall k$ \COMMENT{stages completed by code; failed attempts per stage}
\WHILE{episode not terminated}
  \STATE $o \leftarrow E.\mathrm{observe}()$
  \STATE $a_{1:H},\, c \leftarrow \pi(o, \ell)$ \COMMENT{direct action chunk $a_{1:H}$ and code slice $c$ for the next stage}
  \STATE $\mathit{code\_attempted} \leftarrow \textsc{Arbitrate}(c,\, \mathcal{G},\, \mathcal{D},\, n,\, R)$ \COMMENT{whether code execution was attempted}
  \IF{\NOT $\mathit{code\_attempted}$}
    \FOR{$t = 1$ \TO $H$}
      \STATE $E.\mathrm{step}(a_t)$ \COMMENT{fall back to the direct actions}
    \ENDFOR
  \ENDIF
\ENDWHILE
\end{algorithmic}
\end{algorithm}

\begin{algorithm}[t]
\caption{\textsc{Arbitrate}$(c, \mathcal{G}, \mathcal{D}, n, R)$: execute the slice only if it passes every gate}
\label{alg:hybrid_arbitrate}
\begin{algorithmic}[1]
\STATE $k \leftarrow$ stage index declared in the header of $c$
\IF{$k \in \mathcal{D}$ \OR $k$ is below a stage already attempted \OR $n_k \ge R$}
  \RETURN \FALSE \COMMENT{stage gate}
\ENDIF
\IF{$c$ uses a variable that is not defined in $\mathcal{G}$}
  \RETURN \FALSE \COMMENT{dependency gate}
\ENDIF

\IF{$c$ does not compile \OR $c$ calls a function not in $\mathcal{G}$ \OR $c$ has nothing to execute}
  \RETURN \FALSE \COMMENT{syntax, whitelist, and vacuity gates}
\ENDIF
\IF{$c$ already failed on stage $k$ and nothing has changed since}
  \RETURN \FALSE \COMMENT{skip a hopeless retry}
\ENDIF
\STATE execute $c$ in $\mathcal{G}$ on the robot
\IF{it raised no exception}
  \STATE $\mathcal{D} \leftarrow \mathcal{D} \cup \{k\}$ \COMMENT{stage $k$ is done}
\ELSE
  \STATE $n_k \leftarrow n_k + 1$
\ENDIF
\RETURN \TRUE \COMMENT{code execution was attempted, even if an exception occurred}
\end{algorithmic}
\end{algorithm}

\section{Additional Simulation Experiment Details}
\label{app:additional_sim_experiment}

\subsection{RoboTwin Per-Task Breakdown}
Table~\ref{tab:robotwin_pertask} reports the per-task breakdown of the aggregate results in Table~\ref{tab:robotwin_main}.

\begin{table*}[t]
\centering
\tiny
\caption{\textbf{Per-task RoboTwin~2.0 Results under the \textit{Clean} and \textit{Random} Evaluation Settings (50 Tasks).}
Success rate (\%) over 50 rollouts per task per setting.
Best result per task and setting is shown in bold.}
\label{tab:robotwin_pertask}
\setlength{\tabcolsep}{2pt}
\resizebox{\textwidth}{!}{
\begin{tabular}{lcc|cc|cc|cc|cc|cc}
\toprule
Task & \multicolumn{2}{c}{$\pi_{0.5}$} & \multicolumn{2}{c}{EgoScale} & \multicolumn{2}{c}{DreamZero} & \multicolumn{2}{c}{WLA} & \multicolumn{2}{c}{\textbf{\Ours{} (ours)}} & \multicolumn{2}{c}{\textbf{\Ours{} (hybrid)}} \\
\cmidrule(lr){2-3}\cmidrule(lr){4-5}\cmidrule(lr){6-7}\cmidrule(lr){8-9}\cmidrule(lr){10-11}\cmidrule(lr){12-13}
 & Clean & Rand. & Clean & Rand. & Clean & Rand. & Clean & Rand. & Clean & Rand. & Clean & Rand. \\
\midrule
Adjust Bottle & 62 & 18 & 26 & 0 & 0 & 6 & \textbf{100} & 0 & \textbf{100} & \textbf{50} & 90 & 48 \\
Beat Block Hammer & 48 & 16 & 22 & 0 & 26 & 12 & 82 & 0 & \textbf{94} & \textbf{28} & 46 & 24 \\
Blocks Ranking RGB & 74 & 16 & 0 & 0 & 0 & 0 & 54 & 0 & \textbf{96} & 32 & 94 & \textbf{82} \\
Blocks Ranking Size & 32 & 0 & 0 & 0 & 16 & 0 & 32 & 0 & 56 & 0 & \textbf{64} & \textbf{50} \\
Click Alarmclock & 44 & 56 & 88 & 16 & \textbf{100} & 24 & \textbf{100} & 42 & 96 & 56 & \textbf{100} & \textbf{62} \\
Click Bell & 32 & 40 & 92 & 36 & \textbf{100} & 66 & \textbf{100} & 48 & 96 & 18 & \textbf{100} & \textbf{72} \\
Dump Bin Bigbin & 72 & \textbf{70} & 32 & 18 & 0 & 0 & \textbf{100} & 12 & 98 & 20 & 78 & 14 \\
Grab Roller & 94 & 46 & 96 & 6 & 64 & 22 & \textbf{100} & 0 & \textbf{100} & 32 & 92 & \textbf{50} \\
Handover Block & 16 & \textbf{0} & 4 & \textbf{0} & 0 & \textbf{0} & 8 & \textbf{0} & \textbf{82} & \textbf{0} & 4 & \textbf{0} \\
Handover Mic & 26 & 4 & 0 & 0 & 16 & 0 & 68 & 0 & \textbf{98} & \textbf{12} & 34 & 2 \\
Hanging Mug & 10 & \textbf{4} & 0 & 0 & 0 & 0 & 8 & 0 & \textbf{30} & 0 & 2 & 0 \\
Lift Pot & 8 & 2 & 0 & 0 & 8 & 0 & 88 & 0 & \textbf{90} & \textbf{4} & 66 & 0 \\
Move Can Pot & 28 & 0 & 24 & 0 & 4 & 4 & 64 & 0 & 76 & 6 & \textbf{82} & \textbf{52} \\
Move Pillbottle Pad & 60 & \textbf{44} & 36 & 0 & 34 & 16 & 72 & 0 & 92 & 6 & \textbf{96} & 6 \\
Move Playingcard Away & 90 & \textbf{52} & 46 & 0 & 0 & 4 & 94 & 0 & \textbf{100} & 46 & 86 & 30 \\
Move Stapler Pad & 22 & \textbf{6} & 0 & 0 & 0 & 0 & 2 & 0 & \textbf{38} & 0 & 26 & 2 \\
Open Laptop & 68 & 20 & 20 & 0 & 66 & 22 & \textbf{90} & 0 & 84 & 38 & 72 & \textbf{48} \\
Open Microwave & 26 & 8 & 8 & 2 & 12 & 8 & 0 & 0 & 20 & 4 & \textbf{46} & \textbf{24} \\
Pick Diverse Bottles & 56 & 18 & 44 & 0 & 40 & 16 & \textbf{76} & 12 & \textbf{76} & \textbf{36} & 74 & 32 \\
Pick Dual Bottles & 82 & 14 & 38 & 4 & 54 & \textbf{42} & 80 & 0 & 80 & 28 & \textbf{96} & 28 \\
Place A2B Left & 60 & 10 & 0 & 0 & 22 & 14 & 48 & 0 & 84 & 30 & \textbf{98} & \textbf{58} \\
Place A2B Right & 58 & 14 & 4 & 0 & 24 & 8 & 54 & 0 & 80 & 26 & \textbf{96} & \textbf{80} \\
Place Bread Basket & 68 & \textbf{44} & 44 & 0 & 62 & 12 & 84 & 0 & \textbf{92} & 20 & 80 & 6 \\
Place Bread Skillet & 86 & \textbf{46} & 32 & 0 & 36 & 34 & 64 & 0 & \textbf{90} & 14 & 24 & 2 \\
Place Burger Fries & 90 & \textbf{78} & 78 & 0 & 70 & 62 & \textbf{100} & 0 & 94 & 20 & 66 & 22 \\
Place Can Basket & 40 & 0 & 6 & 0 & 26 & \textbf{12} & 42 & 6 & \textbf{70} & 8 & 26 & 8 \\
Place Cans Plasticbox & 94 & \textbf{74} & 40 & 0 & 0 & 12 & \textbf{100} & 0 & 98 & 12 & 80 & 34 \\
Place Container Plate & 96 & \textbf{58} & 46 & 0 & 96 & 44 & \textbf{100} & 0 & 94 & 26 & 86 & 22 \\
Place Dual Shoes & 54 & \textbf{12} & 8 & 0 & 20 & 0 & 32 & 0 & \textbf{64} & \textbf{12} & 2 & 2 \\
Place Empty Cup & 74 & \textbf{56} & 72 & 0 & 76 & 50 & 94 & 0 & \textbf{98} & 16 & 96 & 14 \\
Place Fan & 36 & \textbf{30} & 12 & 0 & 0 & 0 & 8 & 0 & \textbf{70} & 4 & 16 & 10 \\
Place Mouse Pad & 18 & 0 & 2 & 0 & 10 & 0 & 26 & 0 & \textbf{64} & 0 & 18 & \textbf{4} \\
Place Object Basket & 24 & 4 & 0 & 0 & 10 & 0 & 42 & 2 & \textbf{84} & \textbf{14} & 50 & 4 \\
Place Object Scale & 58 & 44 & 8 & 0 & 36 & 8 & 36 & 0 & 70 & 18 & \textbf{88} & \textbf{50} \\
Place Object Stand & \textbf{94} & 40 & 26 & 0 & 50 & 26 & 92 & 6 & 92 & 26 & 62 & \textbf{56} \\
Place Phone Stand & 40 & 4 & 8 & 0 & 22 & 0 & 86 & 0 & \textbf{90} & 22 & 82 & \textbf{54} \\
Place Shoe & 64 & 24 & 22 & 6 & 18 & 6 & 72 & 0 & \textbf{78} & 22 & 60 & \textbf{30} \\
Press Stapler & 42 & 32 & \textbf{100} & 20 & 88 & 44 & 82 & 0 & 86 & 8 & \textbf{100} & \textbf{68} \\
Put Bottles Dustbin & 14 & 2 & 0 & 0 & 4 & 6 & 16 & 0 & \textbf{86} & \textbf{12} & 14 & 0 \\
Put Object Cabinet & 22 & 0 & 12 & 0 & 4 & 0 & \textbf{68} & 0 & 56 & \textbf{6} & 66 & 0 \\
Rotate QRcode & 50 & 2 & 32 & 0 & 0 & 0 & 22 & 0 & \textbf{70} & 4 & 40 & \textbf{28} \\
Scan Object & 62 & \textbf{38} & 4 & 0 & 40 & 20 & \textbf{80} & 0 & 68 & 10 & 16 & 10 \\
Shake Bottle Horizontally & 98 & 74 & 72 & 30 & \textbf{100} & \textbf{86} & \textbf{100} & 26 & \textbf{100} & 62 & \textbf{100} & 26 \\
Shake Bottle & 96 & 72 & 80 & 6 & \textbf{100} & \textbf{76} & \textbf{100} & 24 & \textbf{100} & 58 & 94 & 30 \\
Stack Blocks Three & 68 & 16 & 0 & 0 & 10 & 0 & 12 & 0 & \textbf{92} & 18 & 62 & \textbf{56} \\
Stack Blocks Two & 80 & 56 & 6 & 0 & 58 & 46 & \textbf{90} & 0 & 70 & 18 & 80 & \textbf{82} \\
Stack Bowls Three & 38 & 36 & 0 & 0 & 14 & 14 & \textbf{88} & 0 & 78 & \textbf{40} & 44 & 4 \\
Stack Bowls Two & 82 & 44 & 0 & 0 & 88 & 24 & \textbf{100} & 0 & 98 & \textbf{56} & 52 & 44 \\
Stamp Seal & 32 & 2 & 6 & 0 & 40 & \textbf{16} & 40 & 0 & \textbf{78} & 2 & 58 & 4 \\
Turn Switch & \textbf{56} & \textbf{40} & 12 & 0 & 14 & 18 & 42 & 0 & 44 & 28 & 52 & 28 \\
\midrule
\textbf{Average} & 54.9 & 27.7 & 26.2 & 2.9 & 33.6 & 17.6 & 64.8 & 3.6 & \textbf{80.8} & 20.6 & 63.1 & \textbf{29.2} \\
\bottomrule
\end{tabular}
}
\end{table*}

\clearpage

\subsection{Hybrid Actions Evaluation Details}
\label{app:robotwin_hybrid_details}

\textbf{Perception and Control Primitives.} During hybrid evaluation, generated code executes through perception and control primitives implemented using Molmo for visual grounding~\citep{deitke2024molmo}, SAM~3 for segmentation~\citep{carion2025sam3}, Contact-GraspNet for grasp generation~\citep{sundermeyer2021contactgraspnet}, and RoboTwin's native motion-planning and gripper-control interfaces~\citep{chen2025robotwin20scalabledata}. Neither the generated programs nor the underlying primitives access privileged simulator information during evaluation. For demonstration collection, we preserve the same API interface but use privileged object states and annotated grasps to improve collection efficiency.

\textbf{Scope of Comparison.} The default and hybrid configurations differ in both their training demonstration collection pipelines and execution modes to support their training requirements. Unlike the default configuration, which uses demonstrations collected with RoboTwin's native motion planner, the hybrid configuration uses CaP-based data collection because it naturally pairs each demonstration with an aligned program executable in the RoboTwin simulator, supporting the code execution component of hybrid actions. Therefore, our resulting comparison evaluates the two configuration as a whole and does not isolate the effect of hybrid execution alone.

\textbf{Code and Action Usage.} Table~\ref{tab:channel_occupancy} reports the fraction of simulation steps driven by each of the two control channels during the hybrid evaluation in Table~\ref{tab:robotwin_main}: execution of code produced by the code expert and direct action chunks generated by the action expert. On successful episodes, code drives 14.5\% of steps in the clean setting and 35.0\% in the random setting. The gap follows from Table~\ref{tab:robotwin_main}. In the clean setting, the action expert alone is already strong (80.8\%), so there is little for code to add. Admitting code in fact lowers success to 63.1\%, which we attribute to the code channel inheriting the failure modes of the low-level perception and control primitives it calls, and the arbiter admits a code slice on validity alone, without predicting whether it will outperform the action chunk it displaces. In the random setting, where direct action degrades to 20.6\%, the code channel carries a third of the physical time on the episodes that succeed and lifts success to 29.2\%. Executing the code expert's generation thus complements the action expert especially where learned actions are less reliable, here under visually out-of-distribution scenes. Pooled over all episodes, code execution accounts for only 8.5\% of steps in the clean setting and 5.6\% in the random setting. The pooled share is dominated by failed episodes, which run to the step cap and accumulate roughly five times as many steps as successful ones (52.7k vs.\ 11.5k in clean, 55.5k vs.\ 8.4k in random), almost all under direct action. 

\begin{figure*}[t]
\centering
\includegraphics[width=\textwidth]{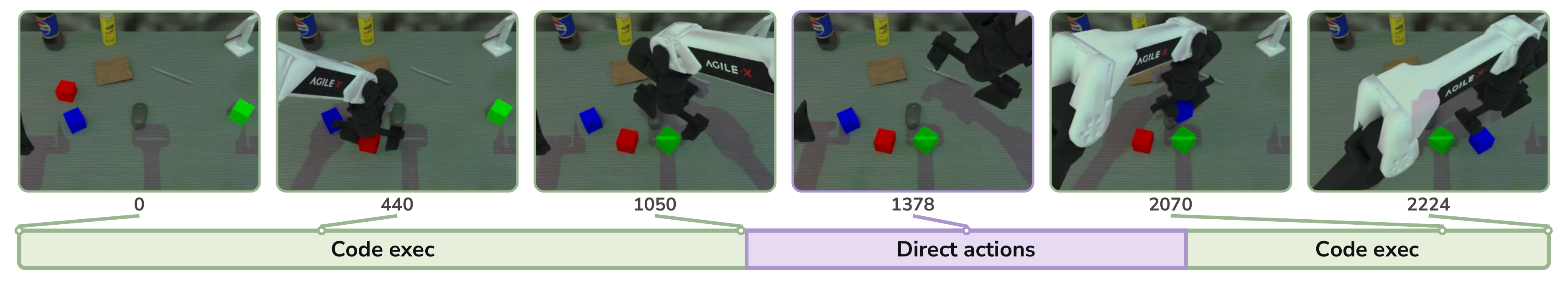}\\[4pt]
\includegraphics[width=\textwidth]{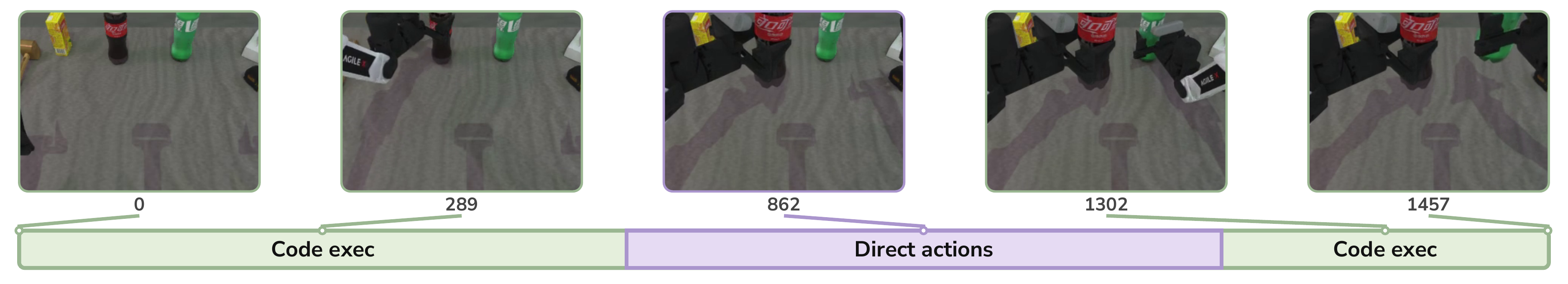}
\caption{\textbf{Qualitative examples of hybrid actions for RoboTwin~2.0 tasks.} Each row shows one successful episode from the hybrid evaluation in Table~\ref{tab:robotwin_main}. Top: \texttt{blocks\_ranking\_rgb}. Bottom: \texttt{pick\_dual\_bottles}.}
\label{fig:hybrid_progress}
\end{figure*}

\textbf{Qualitative Examples of Code--Action Interleaving.} Figure~\ref{fig:hybrid_progress} shows two qualitative examples of how code execution and direct action generation interleave within an episode. The top progress bar shows an episode of \texttt{blocks\_ranking\_rgb}, where the three blocks must be arranged in a row in red, green, blue order. We observe that the three pick-and-place cycles are almost entirely carried out by code execution, with direct action generation taking over around a stage transition. For tasks of this kind that involve regularly shaped objects and well-structured motions, the integration is able to utilize the stable and smooth behavior yields by the CaP actions. The bottom progress bar shows an episode of \texttt{pick\_dual\_bottles}, where the two bottles must be picked up and held together. Here, code execution correctly localized the bottle and moves the gripper to it but fails to secure the grasp due to its less regular shape. Direct action generated from the model then takes over and completes the grasp, after which code execution resumes to finish the episode.

\begin{table}[t]
\centering
\small
\caption{\textbf{Share of simulation steps driven by each control channel in the hybrid evaluation of RoboTwin tasks in Table~\ref{tab:robotwin_main}.} Code exec denotes steps driven by executing the generated code from the code expert; Direct action denotes steps directly generated by the action expert.}
\begin{tabular}{llrrr}
\toprule
& & \multicolumn{2}{c}{Share of sim.\ steps} & \\
\cmidrule(lr){3-4}
Setting & Episode set & Code exec & Direct action & Steps / ep. \\
\midrule
\multirow{3}{*}{Clean}
 & All         &  8.5\% & 91.5\% & 26,720 \\
 & Successful  & 14.5\% & 85.5\% & 11,549 \\
 & Failed      &  6.3\% & 93.7\% & 52,687 \\
\midrule
\multirow{3}{*}{Random}
 & All         &  5.6\% & 94.4\% & 41,716 \\
 & Successful  & 35.0\% & 65.0\% &  8,400 \\
 & Failed      &  3.7\% & 96.3\% & 55,483 \\
\bottomrule
\end{tabular}
\label{tab:channel_occupancy}
\end{table}

\section{Agentic Annotation Details}
\label{app:agentic_annotation_details}


\subsection{Low-level Primitives}
\label{app:low_level_primitives}
\begin{figure*}[t]
\centering
\includegraphics[width=\textwidth]{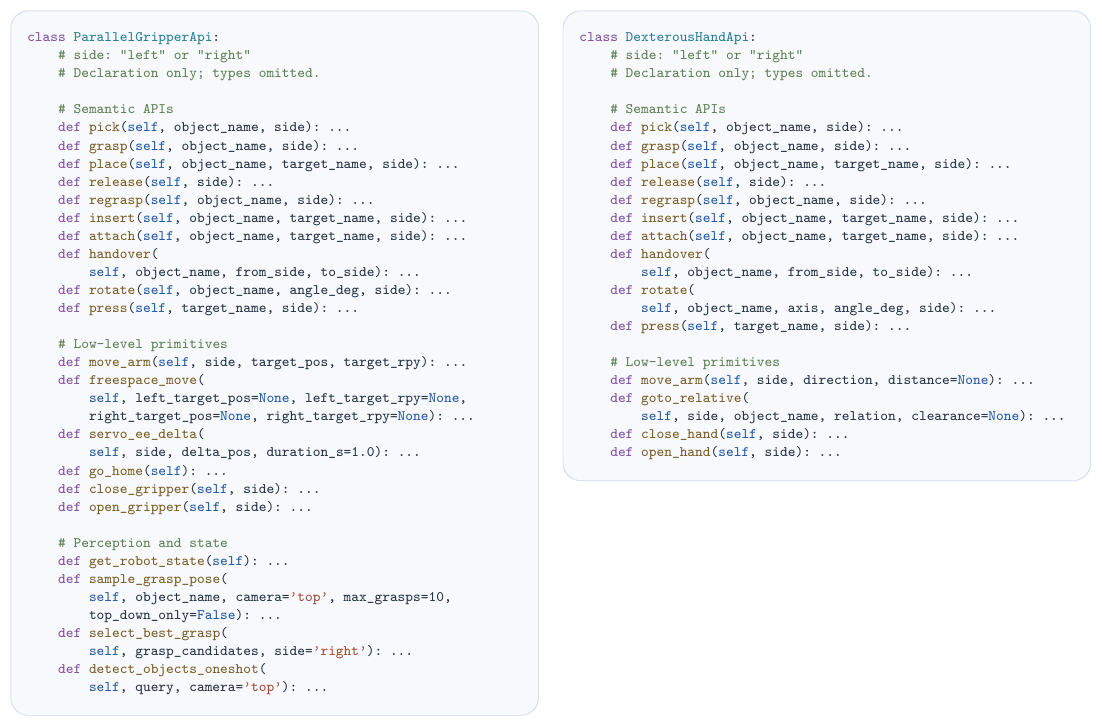}
\caption{\textbf{Declaration-only APIs for parallel grippers and dexterous hands.}}
\label{fig:declarative_apis}
\end{figure*}

Figure~\ref{fig:declarative_apis} shows the two API surfaces for the parallel-gripper (left) and the dexterous-hand (right) embodiments. They share ten semantic contact verbs but differ in their low-level primitives: end-effector poses for the gripper and relative motion for the dexterous hand. Each API call declares the contact events it produces, which is what allows a program to be grounded against a demonstration.

\subsection{Annotation Agent Prompts and Coordinator Protocol}
\label{app:agent-prompts}

The framework has two agents: a VLM \emph{annotator} that writes program lines for one subtask slice per call, and an agentic \emph{coordinator} that audits annotator output between rounds and grows the skill library (Stage~1). After the library is frozen, the same annotator runs over the remaining corpus and over downstream demonstrations, reusing cached subtask code where the embodiment's evidence verifies it. The full API docstrings and skill documentation are appended to the system prompt and omitted here.

\begin{agentprompt}{Shared annotator system instructions}
\begin{promptrules}

\item[Role and output]
Read the frames of one subtask and express the observed actions in Python using only the \texttt{robot} API and documented skills. Output only new code for this subtask, continuing the preceding program.

\item[Grounding and coverage]
The subtask annotation states \emph{what} happens; the frames ground \emph{which} objects and \emph{how} they move. Cover the whole frame span and invent nothing that neither source supports.

\item[Abstraction level]
Prefer the single highest-level call whose contact effects match; never wrap a semantic verb in the primitives it already contains. Use bare open/close primitives only when no verb fits.

\item[Repetition and literals]
Write repeated in-place twisting as one literal loop with the count read from the frames. Hand sides and loop counts must be literals. End on a complete statement.

\item[Static validation]
Code is checked without execution: syntax in program context, existence of called functions, argument binding, and vocabulary membership of literals. On failure, the annotator sees its output along with the identified problems and rewrites the slice.

\end{promptrules}
\end{agentprompt}

\begin{agentprompt}{System prompt for egocentric human demonstrations}
\begin{promptrules}

\item[Setting]
Two hands, \texttt{`left'} and \texttt{`right'}, seen from a head-mounted camera. No contact-event list is given; infer every contact from the frames and the annotation, judging object motion relative to the scene.

\item[Contact semantics]
\texttt{pick}/\texttt{grasp} close a hand (pick lifts, grasp holds in place); \texttt{place}/\texttt{release}/\texttt{insert}/\texttt{attach} open it; \texttt{regrasp} opens then closes; \texttt{press} closes then opens; \texttt{handover} closes the receiver before opening the giver.

\item[Motion and naming]
Motion is purely semantic: \texttt{move\_arm} along a documented direction, \texttt{goto\_relative} at a documented relation to a named object, \texttt{rotate} about a documented axis. No coordinates; objects are named in plain language.

\item[Conventions]
Calls return \texttt{None}. Distances only when visibly measurable; angles as semantic constants.

\end{promptrules}
\end{agentprompt}

\begin{agentprompt}{System prompt for YAM robot demonstrations}
\begin{promptrules}

\item[Setting]
Two arms seen from a top-down camera. The sensor-derived gripper-event list is authoritative: emitted code must reproduce every open/close transition on the named arm, in order, with nothing added, dropped, or reordered.

\item[Event gaps]
If the annotation implies a contact the event list lacks, write the approach and transport in full but do not emit the missing open/close call.

\item[Contact semantics]
\texttt{grasp}/\texttt{pick} close; \texttt{place}/\texttt{release} open; \texttt{regrasp} opens then closes; \texttt{insert} closes then opens; \texttt{handover} closes, closes, then opens. A repeated same-arm cycle in the list is one literal loop.

\item[Perception and motion]
Every pose used by a motion call comes from a perception result earlier in the program; nothing is hard-coded. Absolute single-arm, synchronized two-arm, and small relative motion each have their own primitive.

\end{promptrules}
\end{agentprompt}

\begin{agentprompt}{Per-slice user message}
\begin{promptrules}

\item[Overall task]
\textit{[task instruction]}

\item[Current subtask]
\textit{[human subtask annotation]}

\item[Frames]
\textit{[ordered frames with timestamps]}. On YAM, the frame before each gripper event is captioned with the event index, kind, arm, and time.

\item[Program so far]
\textit{[code written for the preceding subtasks]}

\item[Egocentric directive]
Decide from the frames and annotation where each hand closes on and opens off an object, and produce exactly those contacts with the visible motion around them. A comment-only slice is correct only when the hands neither move nor touch anything.

\item[YAM directive]
Observed gripper events: \textit{[ordered list, repeated cycles collapsed to a count]}. Reproduce this sequence exactly, with the visible motion around it; a brief blip with nothing in hand gets the minimal call.

\item[Output]
Continue the program with this subtask's code only.

\end{promptrules}
\end{agentprompt}

Only the directive for the current dataset is included.

\begin{agentprompt}{Annotation under the frozen library}
\begin{promptrules}

\item[Same prompt, frozen vocabulary]
The system prompt above is used unchanged, with the skill block fixed to the frozen library. Inherited skills are called as plain functions with \texttt{robot} as the first argument.

\item[No new vocabulary]
The annotator may not invent helper names; a maneuver no verb or skill expresses is written from primitives. Extending the library is the coordinator's decision.

\item[Verified reuse]
A per-task cache of subtask code is consulted before each call. A variant is reused only when the embodiment's evidence verifies it: on YAM, the contact events its code implies match the observed events (mirrored forms accepted, motion arguments re-derived); on egocentric data, its hand-motion profile is within a fixed distance of the new slice, and loop templates are never reused. A miss triggers the ordinary call, whose output is written back.

\end{promptrules}
\end{agentprompt}

\begin{agentprompt}{Coordinator protocol (Stage 1, skill discovery)}
\begin{promptrules}

\item[Role]
A coding agent under a fixed written protocol that dispatches rounds, audits their output, and owns the skill library. It never writes program code and never edits the annotator, prompts, API, or verification code.

\item[Round lifecycle]
Dispatched, gated (all episodes annotated and scored), audited, done (admissions applied and logged). The library is frozen from dispatch to done, and the next round starts only then.

\item[Audit]
Read the round summary, then per-task results, then the stored trace of each questionable slice (frames seen, code written, verification outcome). Classify each as detector limit, annotator misread, abstraction-level break, vocabulary gap, or segmentation artifact. Only vocabulary gaps feed skill proposals.

\item[Skill admission]
A candidate is a straight-line composed function over the API, distilled from observed call sequences. It is admitted if generic (episode-specific values lifted to parameters), recurrent across at least two tasks, at the right level (a maneuver, not a task recipe), and not a duplicate. Its contact contract is derived by expansion; on YAM the evidence slices must also reconcile against observed events. Admitted skills enter the annotator prompt from the next round.

\end{promptrules}
\end{agentprompt}

\section{Limitations and Future Work}
\label{app:limitations_future_work}

\textbf{Program Grounding and Skill Coverage.}
The code stream learns from automatically annotated demonstrations and a curated skill library, making its planning quality dependent on annotation accuracy and skill coverage. Generated programs can select an appropriate skill yet contain incorrect arguments, as illustrated in Appendix~\ref{app:pretrained_code_quality}; syntactic validity therefore does not ensure correct execution. Future work could use execution feedback to improve program grounding and discover or refine reusable skills beyond the initial library.

\textbf{Predictive Robustness and Hybrid Execution.}
Structured planning improves WAM performance but does not eliminate sensitivity to visual distribution shifts. Hybrid execution improves robustness under RoboTwin randomization, yet reduces clean-setting performance (Appendix~\ref{app:robotwin_hybrid_details}). This trade-off reflects both the reliability of the underlying primitives and the limitation of selecting code based on validity without estimating its advantage over the action chunk. Learning to arbitrate using execution reliability and expected task progress could better exploit the complementary strengths of the two modes.

\textbf{Toward Continual Agentic Learning.}
Our real-world hybrid evaluation is limited to YAM platform due to the lack of code-as-policy systems for dexterous hand, and the current system does not learn online from its own interactions. The unified design nevertheless offers a basis for coupling program generation and world prediction with execution feedback and model updates. Future work could explore iterative plan revision, skill-library expansion, and autonomous learning through sustained environmental interaction, with broader evaluation across tasks and embodiments.

\end{document}